\documentclass[a4paper,fleqn]{article}
\usepackage{csl}
\usepackage[numbers,sort&compress]{natbib}

\usepackage[american]{babel}
\usepackage{microtype}
\usepackage{mathtools}
\usepackage{amssymb}
\usepackage{bm}
\usepackage{siunitx}
\usepackage{textcomp}
\usepackage[table,dvipsnames]{xcolor}
\usepackage{array}
\usepackage{tabularx}
\usepackage{tabularray}
\UseTblrLibrary{booktabs}
\usepackage{multirow}
\usepackage{makecell}
\usepackage{float}
\usepackage{tikz}
\usetikzlibrary{arrows.meta,positioning,shapes.geometric,calc}
\usepackage{pgfplots}
\usepgfplotslibrary{groupplots,colormaps}
\pgfplotsset{compat=1.18}
\usepackage{enumitem}
\usepackage{placeins}
\usepackage{nameauth}
\usepackage[nameinlink,capitalise,noabbrev]{cleveref}
\usepackage{adjustbox}
\usepackage{xurl}
\usepackage{float}

\begin{document}
\let\WriteBookmarks\relax
\def\floatpagepagefraction{1}
\def\textpagefraction{.001}

\title{Benchmarking Machine Learning Uncertainty Quantification Methodologies for Predicting Turbine Gas Temperature Degradation}

\cslauthor{Jostein Barry-Straume,\\Changmin Son, Adrian Sandu, \\Gavan Burke, Rekha Sundararajan, \\Andrew Rimell, James G. Steinrock}

\cslyear{26}
\cslreportnumber{2}
\cslemail{jostein@vt.edu}

\csltitlepage{}

\begin{abstract}
Effective prognostics and health management of modern engines relies on accurate turbine gas temperature predictions and robust uncertainty quantification to ensure reliability and safety. This paper investigates five major approaches for constructing prediction intervals---namely the Delta method, Bayesian Monte Carlo Dropout, Bootstrap method, Lower--Upper Bound Estimation, and Mean--Variance Estimation---as a means of capturing the uncertainty in neural network predictions of turbine gas temperature. Each approach is implemented within a unified experimental framework that employs cross-validation for hyperparameter selection, repeated train--test splits for performance robustness, and multiple metrics to evaluate both the accuracy and tightness of the intervals. In particular, Coverage Probability, Normalized Mean Prediction Interval Width, and the Coverage Width-based Criterion are measured to comprehensively assess each method's reliability and sharpness. Experiments conducted on a representative turbine gas temperature dataset reveal distinct trade-offs among the five methods in terms of interval coverage, width, and stability. These findings provide a practical guide for selecting and tuning prediction interval methods in engine health management and prognostics, ensuring both interpretability and precision in real-world applications.
\end{abstract}

\section{Introduction}

Engine Health Management (EHM) has become a cornerstone of modern aerospace operations, aiming to maximize engine availability, optimize maintenance schedules, and mitigate in-flight risks through timely detection and prognosis of component degradation \cite{kobayashi2005}. Within this framework, accurately estimating the Remaining Useful Life (RUL) of critical engine components is paramount, as it allows operators to anticipate when parts may fail and to schedule inspections or replacements in a cost-effective and safe manner \cite{saxena2008phm}. A key indicator in these prognostic efforts is the turbine gas temperature (TGT). As gas-path components degrade, reductions in compressor and turbine efficiency often require higher fuel flow to deliver the same thrust, which manifests as an increase in exhaust gas temperature. Sustained elevated TGT also accelerates hot-section creep, oxidation, and fatigue, thereby shortening RUL. Consequently, TGT trends strongly correlate with performance decline and mechanical wear in turbofan gas turbines \cite{hanachi2018performance}. Modeling and predicting TGT is therefore a critical step in health monitoring and a valuable input to downstream prognostic models \cite{baptista2018}.

In parallel, recent research has underscored the importance of uncertainty quantification (UQ) to capture how model estimates may vary due to data noise, sensor errors, or inherent stochastic processes \cite{khosravi2011}. By providing a rigorous measure of predictive reliability, UQ enables practitioners to identify how confidently particular RUL or TGT predictions should be interpreted, which in turn can guide maintenance decisions and risk analyses. Despite considerable advancements in machine learning--based prognostics, many of these methods primarily focus on point predictions, leaving users with limited insight into the variability and reliability of model outputs. Quantifying this uncertainty provides an additional layer of safety and robustness, especially in mission-critical applications where underestimating the risk of engine failure can lead to catastrophic outcomes.

Most studies that explore RUL prediction for gas turbine engines validate their proposed methods using NASA's publicly available C-MAPSS dataset, a suite of simulated run-to-failure trajectories \cite{saxena2008nasa}. While the simulated setting offers valuable insights into data-driven prognostic modeling, its relatively controlled nature can be misleading; real-world engines generate noisier, incomplete, and more heterogeneous signals. As a result, models trained and benchmarked exclusively on well-curated simulation data often struggle to maintain their performance once confronted with operational measurements that deviate from idealized conditions \cite{coble2010}. This discrepancy highlights an important gap in the literature: the need for robust and comprehensive approaches that can tackle the genuine complexities found in real-world engine health data.

In light of these challenges, the present work provides a comprehensive benchmarking of state-of-the-art uncertainty quantification (UQ) methodologies for \emph{turbine gas temperature (TGT) prediction}. The primary objective is to construct prediction intervals that accompany neural-network point forecasts, allowing practitioners to assess when predicted temperatures approach safety margins and how much uncertainty remains in different operating regimes. Although these uncertainty estimates can ultimately be propagated into Remaining Useful Life (RUL) estimators, this paper focuses on the TGT regression task and on comparing PI construction methods under a unified experimental protocol. Interval quality is evaluated using the coverage and width metrics in \cref{rv:sec:pi_metrics}---including PICP (\cref{rv:eq:picp}), NMPIW (\cref{rv:eq:nmpiw}), and CWC (\cref{rv:eqn:cwc})---alongside pointwise error statistics.

\section{Engine Health Management}

Engine Health Management (EHM)---often discussed within the broader Prognostics and Health Management (PHM) literature---encompasses a set of methodologies and technologies designed to monitor the current condition of an engine, predict its future performance, and manage maintenance actions accordingly \cite{saxena2008phm, Si2011}. Advances in sensors, data acquisition systems, and machine learning techniques have made real-time engine health monitoring not only feasible but also highly effective in reducing downtime, preventing catastrophic failures, and optimizing overall lifecycle costs \cite{Heimes2008, Jardine2006}.

Modern engines, whether in aircraft, automotive, or industrial applications, operate under increasingly stringent performance requirements and environmental constraints \cite{randall2021vibration}. Failures or unexpected breakdowns of these critical components can lead to severe economic and safety consequences. Effective engine health management is thus vital for several reasons: \textbf{(1) Safety:} Proactive detection of potential engine faults and anomalies reduces the risk of catastrophic failures, safeguarding lives and assets \cite{Jardine2006}. \textbf{(2) Reliability:} By continuously assessing engine condition, maintenance can be scheduled as needed, avoiding unplanned downtimes and ensuring consistent performance \cite{Heimes2008}. \textbf{(3) Cost-Effectiveness:} Condition-based maintenance, a core aspect of EHM, optimizes the usage of maintenance resources, spare parts, and workforce, leading to reduced operational costs \cite{Si2011}. \textbf{(4) Extended Lifespan:} Proper monitoring and timely interventions help to extend the useful life of engine components, thereby maximizing the return on investment \cite{saxena2008phm}.

\subsection{Remaining Useful Lifespan}

One of the key concepts in engine health prognostics is Remaining Useful Life (RUL). RUL is defined as the length of time from the current moment until the engine (or a component of the engine) no longer meets its desired functional and safety thresholds \cite{Si2011, liu2022novel}. Mathematically, it can be expressed as:

\begin{equation}
\label{rv:eq:rul-definition}
\mathrm{RUL} = T_{\mathrm{failure}} - T_{\mathrm{current}}.
\end{equation}

where $T_{\text{failure}}$ is the time (or operating cycle) at which the engine is predicted to fail or to operate outside acceptable limits, and $T_{\text{current}}$ is the present time (or operating cycle).

RUL is important because accurate RUL estimation underpins predictive and proactive maintenance strategies, ensuring repairs or part replacements occur right before a failure mode becomes critical \cite{Heimes2008}. Moreover, maintenance teams can better allocate resources, reduce the inventory of spare parts, and plan interventions well in advance \cite{Jardine2006}. Additionally, knowing how much useful life remains allows operators to balance operational demands (e.g., higher thrust requirements in aircraft engines) with maintenance schedules, preventing over-stressing components and prolonging overall system health \cite{liu2022novel}.

\subsection{Turbine Gas Temperature}

Turbine Gas Temperature (TGT) is a critical parameter in gas turbine engines, representing the temperature at the low pressure turbine (LPT) inlet. The TGT directly reflects the operating conditions within the hot section of the engine and can be used to assess engine performance, thermal stress, and component integrity \cite{Boyce2011}.

TGT prediction is important because accurate TGT measurements and predictions allow operators to monitor key engine parameters, detect abnormal conditions, and diagnose potential faults early \cite{jiang2022onboard}. A rise in TGT might indicate a reduction in thermal efficiency or the onset of component degradation. Furthermore, elevated TGT directly contributes to the thermal stress on key components (e.g., blades, nozzles), accelerating fatigue and oxidation \cite{han2012gas}. Predicting TGT helps in gauging the risk of overheating and preventing damage to these critical parts. Turbine engines often rely on TGT or a function thereof in engine control loops to ensure safe and efficient operation \cite{RollsRoyce2015}. Real-time or near-real-time TGT estimates allow for timely adjustments to fuel flow, compressor settings, or other operational parameters.

\subsubsection{Selection of Predictor Variables}

Given its impact on engine degradation, TGT is a critical predictor of remaining engine life and can therefore be used as a surrogate health indicator. With this in mind, the neural network frameworks discussed in this section will predict TGT as the target variable. A full list of predictor and target variables is shown in \cref{rv:tab:EHM-variables}. For a detailed report on the process of selecting predictor variables, please refer to \cite{jung2023sensitivity,jung2024impact}.

\begin{table}
\caption{Predictor and Target Variables from Engine Health Monitoring Data\label{rv:tab:EHM-variables}}
\centering{%
\small
\begin{tabular}{l l}
\toprule
Variable & Description \\
\midrule
N1 & Low pressure spool speed [\%] \\
N2 & Intermediate pressure spool speed [\%] \\
N3 & High pressure spool speed [\%] \\
P25 & Intermediate pressure compressor delivery pressure [psi] \\ % $(\frac{\lb}{\inch^2})$
P3 & High pressure compressor delivery pressure [psi] \\
TGT & Turbine gas temperature [$^\circ$C] \\
\bottomrule
\end{tabular}
}%
\end{table}

\subsection{Relationship between RUL and TGT}

Fluctuations in TGT, especially persistent increases under normal operating conditions, can be symptomatic of deterioration in the combustion or turbine stage, such as fouling, erosion, or seal leakage \cite{Si2011}. By correlating TGT trends with historical maintenance data, one can derive a measure of how quickly the engine is approaching a point where it can no longer meet performance thresholds. Many RUL models consider temperature-based thresholds at which an engine can no longer operate safely. When a predicted TGT surpasses that threshold or indicates a component's end-of-life condition, RUL calculations can trigger repair or replacement scheduling \cite{khosravi2011}. Higher TGT accelerates creep and fatigue failure mechanisms in turbine blades and other hot-section parts. Thus, TGT predictions feed into the degradation models that compute RUL by quantifying how much ``damage'' (in the form of creep or fatigue) accumulates over time \cite{Boyce2011}.

\subsection{Uncertainty Quantification, RUL, and TGT}

Real-world engine data are noisy, subject to sensor drift, and influenced by changing ambient conditions, fuel quality, and usage patterns \cite{Vachtsevanos2006}. By applying uncertainty quantification techniques, TGT forecasts incorporate these uncertainties into their predicted values \cite{khosravi2011}. Instead of relying on a single-point TGT estimate, UQ methods yield prediction intervals that explicitly show the range of likely values. This is critical for maintenance planning: if the upper bound of TGT is nearing a critical limit, a prudent operator might schedule inspections or part replacements sooner \cite{Jardine2006}. Accurate TGT predictions with well-calibrated uncertainty bounds help distinguish true degradation patterns from normal measurement fluctuations, thus minimizing both unnecessary interventions (false alarms) and unrecognized early failures (missed detections) \cite{Si2011}. RUL estimation intrinsically involves future unknowns. By propagating the uncertainty in TGT predictions into the degradation or reliability models, operators gain a probabilistic understanding of when the engine is likely to fail or breach performance limits. This can significantly enhance risk management strategies and maintenance optimization \cite{saxena2008phm, khosravi2011}.

\subsection{Improving RUL Estimation}

Predicting TGT with robust uncertainty estimates sets the stage for more accurate and reliable RUL models. Hybrid prognostic models can fuse physics of failure (e.g., thermal stress, creep) with data-driven TGT predictions, ensuring that uncertainties in measurement and model assumptions are accounted for \cite{Heimes2008}. Instead of relying on fixed intervals, maintenance managers can adapt schedules dynamically based on the real-time probabilistic trajectory of TGT. When the forecast TGT intervals indicate higher variability or an upward trend, proactive maintenance can be initiated \cite{Jardine2006}. Comprehensive TGT forecasts with uncertainty bounds help avoid premature overhaul (which raises costs) and unplanned downtime (which disrupts operations), ultimately improving the cost-effectiveness and safety of engine operations \cite{RollsRoyce2015}. While these motivations highlight why uncertainty-aware TGT forecasting is valuable, integrating the proposed TGT prediction intervals into a full probabilistic RUL estimation pipeline is beyond the scope of this paper. Future work will focus on propagating the prediction intervals through established data-driven and physics-informed prognostic models (e.g., \cite{Heimes2008,Si2011}) to quantify the impact of temperature uncertainty on end-of-life predictions and maintenance decision-making.

\section{Uncertainty Quantification}

Uncertainty quantification is defined as ``the science of identifying, quantifying, and reducing uncertainties associated with models, numerical algorithms, experiments, and predicted outcomes'' \cite{smith2013uncertainty}. Combining the domains of computing, engineering, and statistics, uncertainty quantification is a framework for shedding light on the degree to which a given model is wrong, useful, and how confident one can be in its predictions \cite{warner2023introduction}.

\subsection{Uncertainty Sources: Aleatoric versus Epistemic}

There are two main sources of uncertainty. The first source is aleatoric uncertainty, also referred to as \emph{data uncertainty}, which is the noise or variation of the feature data that can be characterized as a distribution over a given number of samples \cite{roy2011comprehensive}. The cause of aleatoric uncertainty can be due to the data acquisition process, such as instrument errors, frequency of data recording, and data interpolation \cite{gal2016uncertainty, he2023survey}.
As a stochastic characteristic of the data, it is often considered to be an irreducible uncertainty unless better measurement instruments or experimental controls reduce the observed variability \cite{DerKiureghian2009}. In the context of this work, aleatoric uncertainty is represented by the sensor noise in temperature readings for turbine gas temperature (TGT) or random fluctuations in operating conditions (e.g., ambient temperature, fuel composition).

The second source of uncertainty is called epistemic uncertainty. It is also known as \emph{model uncertainty}, and involves a lack of knowledge on the modeler's part leading to an imperfect training process \cite{roy2011comprehensive, he2023survey}. From the lens of machine learning, this epistemic uncertainty comprises the following sources: (1.) Model architecture (2.) Model parameters (3.) Distributions of training and test datasets. In contrast to aleatoric uncertainty, epistemic uncertainty is often reducible through improved modeling techniques, additional data, or refined domain knowledge. In the context of this work, epistemic uncertainty is represented by possible limited understanding of the exact degradation mechanisms for turbine blades, or a mismatch between real-world operating conditions and those assumed in the training dataset.

In practical prognostics---particularly for critical assets like gas turbines---differentiating these two types of uncertainty is essential. Robust prediction intervals that capture both aleatoric and epistemic uncertainties can lead to more reliable health monitoring and better estimates of remaining useful life (RUL) \cite{khosravi2011}. Moreover, when prognostic algorithms explicitly quantify these uncertainties, maintenance decisions can be driven not only by mean predictions but also by the confidence in those predictions \cite{DerKiureghian2009}. Reducing epistemic uncertainty (e.g., gathering more engine data or improving model fidelity) can lead to narrower prediction intervals and more accurate forecasts, ultimately enhancing the safety and cost-effectiveness of engine health management practices.

The aerospace industry is very risk-averse as it operates in high-stake applications, wherein serious consequences can be realized if a learning algorithm makes an unexpected or overly confident prediction in a real-world situation \cite{he2023survey}. To maximize the utility of learning algorithms' predictive power, while simultaneously minimizing data uncertainty and model uncertainty, uncertainty quantification methods for neural networks are therefore emphasized.

\subsection{Prediction Intervals: Confidence versus Prediction}

Accurate forecasting of engine-related parameters---especially turbine gas temperature (TGT)---requires not only a point estimate but also a quantification of the uncertainty around that estimate. Two commonly used statistical tools for expressing such uncertainty are prediction intervals (PIs) and confidence intervals (CIs). Although both describe uncertainty around parameter estimates, they serve different purposes and are tailored to different types of questions \cite{Freedman2005, khosravi2011}.

A confidence interval provides an estimated range for an underlying (often population-level) parameter, such as a mean or a regression coefficient. It answers questions like, ``What is the likely range for the true mean TGT across many samples?'' A 95\% CI for the mean TGT implies that, if one repeatedly sampled and constructed such intervals, 95\% of these intervals would contain the true (unknown) mean. CIs are typically used for aggregate-level inference, such as validating whether a certain parameter (e.g., average TGT under specific conditions) stays within a desired range.

A prediction interval focuses on individual future observations rather than population parameters. It quantifies the range in which an actual future value---such as the next TGT reading---will likely fall. A 95\% PI indicates that the interval construction is calibrated to contain future observed TGT values at the stated rate under the model assumptions. PIs are particularly useful for pointwise or instance-level predictions, where the goal is to capture the likely spread of an upcoming data point.

In engineering and prognostic applications, the most common choice for the significance level $\alpha$ is 0.05, corresponding to a 95\% coverage probability (i.e., $1-\alpha = 0.95$). This is generally referred to as the ``industry standard'' for uncertainty quantification because it represents a balance between confidence in the model and interval width \cite{DerKiureghian2009}. Other coverage probabilities (e.g., 90\%, 99\%) may be chosen based on factors like safety margins, criticality of the system, or cost implications.

The use of prediction intervals for TGT prediction is more appropriate than that of confidence intervals. When operators forecast TGT at a specific future time or under a set of conditions, they need an interval that captures the likely observed temperature. Confidence intervals would only provide bounds on the mean TGT across repeated sampling, which is less practical for real-time operational decisions. A TGT that spikes above predicted bounds can signal immediate maintenance needs or operational adjustments. Because PIs reflect the total observed variability (aleatoric + model uncertainties), they are better suited to capturing the risk of near-future out-of-bound TGT readings \cite{khosravi2011}.

Inversely, the use of confidence intervals for RUL prediction is more appropriate than that of prediction intervals. RUL forecasts typically hinge on the underlying degradation parameters of the engine (e.g., wear rate, damage progression). A confidence interval on these parameters helps determine whether the system degradation is faster or slower than expected on average. Maintenance strategies based on aggregated degradation rates often rely on whether the estimated mean time-to-failure or mean usage limit remains within safe bounds. Hence, a CI on the mean lifetime or average rate of degradation is typically more relevant. Long-horizon decisions (e.g., part replacements, overhaul schedules) often depend on population-level parameters and reliability metrics. In this setting, a confidence interval on the mean or median RUL is more informative than intervals that capture single future instances of the system's behavior \cite{Freedman2005}.

By combining prediction intervals for short-term, pointwise TGT forecasts with confidence intervals for long-term degradation modeling, engine health management can simultaneously ensure operational safety (spotting near-term temperature excursions) and plan effective maintenance strategies (capturing average degradation trends). As a result, operators gain a multi-faceted view of both the immediate and future health of the engine, aligning short-term responses with long-term cost-effectiveness. This, however, is beyond the scope of the presented work.

\subsection{Evaluation Metrics}\label{rv:sec:pi_metrics}

Prediction interval metrics provide a quantitative way to assess both the reliability (coverage) and the sharpness (width) of predictive uncertainty estimates.

The metrics detailed below are widely used in regression and forecasting tasks to measure both how many observed data points are captured by predictive intervals (coverage) and how large or tight these intervals are (width) \cite{chatfield2000time,  gneiting2007strictly}. The ideal method balances achieving high coverage with minimal interval width, which is precisely what the Coverage Width Criterion aims to formalize \cite{khosravi2011}. 

\subsubsection{Prediction Interval Coverage Probability}

The \emph{Prediction Interval Coverage Probability} (PICP) measures how frequently the true values lie within the predicted interval bounds. Let \(y_i\) denote the true value for the \(i\)-th data point, and let \(\widehat{L}_i\) and \(\widehat{U}_i\) be the corresponding lower and upper predicted interval bounds, respectively. For a dataset of \(N\) observations,
\begin{equation}
\mathrm{PICP} \;=\; \frac{1}{N} \sum_{i=1}^{N} \mathbf{1}\bigl\{y_i \in [\widehat{L}_i,\, \widehat{U}_i]\bigr\},
\label{rv:eq:picp}
\end{equation}
where \(\mathbf{1}\{\cdot\}\) is the indicator function, which equals 1 if \(y_i\) is within the interval \([\widehat{L}_i, \widehat{U}_i]\), and 0 otherwise. A coverage near the nominal confidence (e.g., 95\%) is generally desirable.

\subsubsection{Mean Prediction Interval Width}

The \emph{Mean Prediction Interval Width} (MPIW) focuses on the average width (sharpness) of the prediction intervals. For each data point \(i\), the interval width is \(\widehat{U}_i - \widehat{L}_i\). The MPIW over \(N\) data points is
\begin{equation}
\mathrm{MPIW} \;=\; \frac{1}{N} \sum_{i=1}^{N} \Bigl(\widehat{U}_i \;-\; \widehat{L}_i\Bigr).
\label{rv:eq:mpiw}
\end{equation}
Smaller MPIW values indicate narrower intervals, provided the coverage remains acceptable.

\subsubsection{Normalized Mean Prediction Interval Width}

The \emph{Normalized Mean Prediction Interval Width} (NMPIW) rescales the mean width by the range of the observed data. Let 
\(\max_i(y_i)\) and \(\min_i(y_i)\) be the maximum and minimum observed values of the target variable. Then
\begin{equation}
\mathrm{NMPIW} \;=\; \frac{\mathrm{MPIW}}{\max_i(y_i) \;-\; \min_i(y_i)}.
\label{rv:eq:nmpiw}
\end{equation}
By normalizing, a value of NMPIW closer to 0 indicates narrower intervals \emph{relative to the data's overall range}.

\subsubsection{Coverage Width Criterion}

Proposed by Khosravi et~al., the \emph{Coverage Width Criterion} (CWC) balances both coverage and sharpness in a single measure \cite{khosravi2011}. Let 
\(\mathrm{PICP}\) be the coverage; \(P^*\) be the target coverage (e.g., \(0.95\)); \(\mathrm{MPIW}\) be the mean width; and let \(\eta\) be a positive hyperparameter. One commonly used form is:
\begin{equation}
\mathrm{CWC} \;=\; 
\begin{cases}
\mathrm{MPIW}, 
& \text{if } \mathrm{PICP} \;\geq\; P^*, \\[6pt]
\mathrm{MPIW} \times \Bigl(1 \;+\; \exp\bigl[-\eta\bigl(\mathrm{PICP}-P^*\bigr)\bigr]\Bigr), 
& \text{otherwise.}
\end{cases}
\label{rv:eqn:cwc}
\end{equation}
When coverage \(\mathrm{PICP}\) is below the target \(P^*\), the exponential penalty term increases the criterion, penalizing intervals that fail to meet the coverage goal. A lower CWC is thus better, implying narrower intervals (small \(\mathrm{MPIW}\)) that still meet or exceed the desired coverage.

%=====================================================================
\section{Uncertainty Quantification with Machine Learning}
\label{rv:sec:uqml-literature-survey}
%=====================================================================

A taxonomy of existing neural network uncertainty quantification (UQ) methods is presented in \cite{he2023survey, quan2019survey, kabir2018neural,khosravi2011, papadopoulos2001confidence, ding2003backpropagation}. Other survey papers offer viewpoints from the perspective of Bayesian statistics or neural network architecture, whereas in \cite{he2023survey}, He and Jiang detail UQ methods for neural networks with regard to the uncertainty source. In \citep{quan2019survey}, Quan et~al. explore UQ methodologies with applications to wind power and smart grids. In \citep{kabir2018neural}, Kabir et~al. overview how UQ methodologies have been modified over time. In \citep{khosravi2011}, Khosravi et~al. comprehensively evaluate the performance of UQ methodologies over a variety of datasets. In \citep{dybowski2001confidence}, Dybowski et~al. review a number of methods used for estimating uncertainty with feed-forward neural networks, including delta, bootstrap, and Bayesian methodologies.

\subsection{Distribution-Free Calibration and Conformal Prediction}
\label{rv:sec:conformal-literature-survey}

A major recent thread in VVUQ-oriented machine learning is \emph{distribution-free} (finite-sample) uncertainty quantification, where prediction intervals are guaranteed to achieve a target \emph{marginal} coverage under mild assumptions such as exchangeability. Conformal prediction provides a generic wrapper around any base predictor and outputs intervals with user-specified coverage without assuming Gaussianity or a correctly specified likelihood \citep{vovk2005alrw, shafer2008tutorial}. The key strength is the explicit coverage guarantee; key limitations are that the guarantee is marginal (not conditional) and can fail under distribution shift.

For heteroscedastic regression, \emph{conformalized quantile regression} (CQR) combines conformal calibration with conditional-quantile models to adapt interval widths to the local noise level while still preserving finite-sample coverage \citep{romano2019cqr}. Related stability-based constructions such as the jackknife+ provide distribution-free predictive inference with rigorous coverage properties \citep{barber2021jackknifeplus}.

A complementary line of work focuses on calibration of probabilistic regressors that output predictive distributions or variances \citep{kuleshov2018calibrated}. Finally, quantile regression \citep{koenker1978quantile} and deep-learning variants such as simultaneous quantile regression \citep{tagasovska2019singlemodel} offer a likelihood-free route to heteroscedastic prediction intervals, and can be paired with conformal calibration when finite-sample coverage is required.

\subsection{Bayesian Method for Quantifying Uncertainty}
\label{rv:sec:bayesian-literature-survey}

In \citep{mackay1991evidence}, MacKay introduces their novel Bayesian learning framework using a Gaussian approximation for the posterior distribution of network weights, and single value estimates for the mean and variance.

In \citep{ungar1996estimating}, Ungar et~al. explore both the delta method and the Bayesian method, comparing the two methods' performance to provide accurate prediction intervals for two different synthetic problems. The particular Bayesian method is using the hybrid Monte Carlo algorithm combined with the Metropolis algorithm with sampling based on dynamical simulation. The authors weigh the trade-off between computational cost and accuracy improvement across the methods, noting that the added computational cost may not justify the improved accuracy.

In \citep{quan2014incorporating}, Quan et~al. use the particle-swarm-optimization-based Lower Upper Bound Estimation (PSO-based LUBE) method from \cite{quan2014particle} to create prediction intervals of 19 different nominal coverage $(1-\alpha)$ levels ranging from 5\%-95\% confidence. Subsequently, these prediction interval levels are decomposed into quantiles, which form the basis for 40 discrete points within an empirical cumulative distribution function (ECDF). These discrete points are fitted to the ECDF curve, which is then used in a Monte Carlo simulation to generate hourly wind power forecast scenarios. The problem is then formulated as stochastic security-constrained unit commitment (SCUC), which is optimized via a genetic algorithm.

In \citep{kasiviswanathan2016comparison}, Kasiviswanathan et~al. conduct a comparative analysis on hydrologic modeling between bootstrap, Bayesian, and a LUBE method that the authors refer to as Prediction Interval (PI). A Markov chain Monte Carlo algorithm is used as the Bayesian method. Bootstrap produced the highest root mean square error and highest mean biased error, whereas the performance of the Bayesian and Prediction Interval methods were comparable. The authors further note that Bootstrap and Bayesian approaches underestimated the target values, whereas the PI method overestimated target values. With regard to target uncertainty estimation, bootstrap had high variability and the other two methods had a lower variability.

In \cite{kendall2017uncertainties}, Kendall et~al. put forth a Bayesian deep learning framework which combines input-dependent aleatoric uncertainty with epistemic uncertainty. A strength of this approach is the framework's ability to predict both a pointwise estimate and its corresponding variance. The variance of the prediction is used as a metric of model uncertainty. Success is evaluated by minimization of the Gaussian negative log-likelihood loss. A limitation of this approach is the requirement of prior knowledge of the dataset's distribution, as not all datasets are, or can be assumed to be, Gaussian.

%=====================================================================

\subsection{Bootstrap Method for Quantifying Uncertainty}
\label{rv:sec:bootstrap-literature-survey}

In \citep{heskes1996practical}, Heskes proposes a bootstrap method to construct confidence and prediction intervals. In essence, bagged ensemble bootstrap pair sampling is conducted on training data to improve the estimation of the target distribution. Then, bootstrap residual sampling is conducted on the validation dataset to optimize the interval widths. Success is demonstrated via a synthetic dataset, showing that the presented framework can extrapolate and interpolate with a limited amount of data.

In \citep{tibshirani1996comparison}, Tibshirani compares the performances of the bootstrap pairs approach, the bootstrap residual approach, the delta method, and the sandwich estimator. The author found that the bootstrap methods provided the most accurate estimates of variance (uncertainty). Moreover, the bootstrap pairs approach was found to be slightly more robust than the bootstrap residual method in the case of over-fitted models.

In \citep{carney1999confidence}, Carney et~al. present a novel technique to use bootstrap to estimate confidence and prediction intervals for neural network ensembles. The technique is based on the bootstrap pairs sampling method, but the novelty arises in the fact that the authors subset the ensemble (i.e. 200 networks) into smaller ensembles (i.e. 8 groups of 25 networks) to create a set of variances of distributions, leading to more accurate variance measures of the true Gaussian distribution. In other words, to construct Gaussian confidence intervals the authors use bootstrap pairs sampling for bagged ensembles. Likewise, to construct prediction intervals they do the same thing except use bootstrap residual sampling instead of pairs sampling. 

In \citep{zio2006study}, Zio conducts a parametric analysis between bootstrap and delta methodologies for a synthetic sine wave problem. Compared to the delta method, the bootstrap method had a larger bias, smaller variance (uncertainty), and a smaller root mean square error (RMSE) for target predictions. Afterward, the paper includes an accompanying case study on predicting feedwater flow rate, with uncertainty, in a boiling water reactor using the bootstrap method.

In \citep{giordano2007forecasting}, Giordano et~al. introduce a neural network sieve (NN-Sieve) bootstrap framework, which constructs nonparametric prediction intervals for nonlinear time series data. Success is demonstrated by comparing NN-Sieve to an Auto Regressive Sieve bootstrap framework in a Monte Carlo simulation.

In \citep{kumar2012bootstrap}, Kumar et~al. present a procedure for determining prediction intervals for outputs of non-parametric regression models using bootstrap methods. The bootstrap-based prediction intervals are validated on simulations, and employed in an anomaly detection problem on aviation data. Leveraging the bootstrap prediction intervals, their Conditional Anomaly Detection (CAD) algorithm tests for anomalies with respect to conditional distributions, whereas other CAD algorithms test with respect to the joint or marginal distribution. The authors' CAD algorithm is used to detect airplanes that are over-consuming fuel within the Flight Operational Quality Assurance (FOQA) dataset.

In \citep{khosravi2014constructing}, Khosravi et~al. present a novel prediction-interval-based cost function, together with a modified bootstrap technique, that optimizes prediction intervals. An ensemble of networks ($\text{NN}_{B}$) is used to provide an average pointwise prediction, and the corresponding average pointwise variance. Then, a separate network ($\text{NN}_{\epsilon}$) is used to estimate the variance of the errors (or residuals). A two-stage training phase is used to optimize $\text{NN}_{B}$ with maximum-likelihood estimation and $\text{NN}_{\epsilon}$ with the Coverage Width Criterion (CWC) cost function. The authors note that their optimization algorithm improves the quality of prediction intervals by 28\%, leading to narrower intervals that maintain high coverage probability. 

In \citep{errouissi2015bootstrap}, Errouissi et~al. apply a bootstrap method to create prediction intervals for wind speed forecasting. A numerical weather prediction regression model is created using past pointwise forecast predictions. A parametric approach to creating prediction intervals is presented using the residuals of past wind speed forecasts (bootstrap residual sampling). Additionally, a non-parametric approach to creating prediction intervals is presented using bootstrap pair sampling. Success is measured using the following evaluation metrics on real wind farm data: prediction interval coverage probability (PICP), and prediction interval normalized average width (PINAW).

%=====================================================================

\subsection{Delta Method for Quantifying Uncertainty}
\label{rv:sec:delta-literature-survey}

In \citep{hwang1997prediction}, Hwang et~al. construct asymptotically valid prediction intervals using an artificial neural network, and then show how to use said prediction intervals to select the number of nodes in a network. This is one of the earlier papers to employ the delta methodology with regard to neural networks.

In \citep{de1998prediction}, de Vieaux et~al. use weight decay as an alternative training approach to estimating prediction intervals with neural networks, showing how it may be more effective than early-stopping methods. This is a foundational paper for the delta methodology implemented into neural network frameworks.

In \citep{ho2001neural}, Ho et~al. leverage neural networks with asymptotically valid prediction intervals to tackle nonlinear regression tasks relating to soldering and electronic manufacturing. Furthermore, the authors combine confidence bounds with process control to better inform solder paste deposition performance and decision processes.

In \citep{khosravi2009improving}, Khosravi et~al. optimize delta method neural networks via a genetic algorithm. This optimization improves the quality of the created prediction intervals by finding the optimal structure of the network (i.e. network size and hyper-parameters). By doing so, the authors seek to reduce the width of prediction intervals (so they are practically more usable) without compromising their coverage probability (if the target ground truth lies within the prediction interval).

In \citep{lu2009prediction}, Lu et~al. investigate the delta method and bootstrap method for constructing prediction intervals with their Nonlinear AutoRegressive with eXternal input (NNARX) model. The NNARX model is employed to predict indoor temperature and relative humidity for a complicated, difficult-to-simulate physical test house.

In \citep{wu2016probabilistic}, Wu et~al. present a Long Short-Term Memory framework that provides pointwise forecasting for wind power generation. Moreover, the authors employ cluster analysis via k-means algorithm to divide the wind power forecasts by wind-speed fluctuation into five clusters. The error residuals of these clusters are then used to create empirical probability density functions (PDFs). The model selects the appropriate PDF based on the pointwise prediction and error, and then uses that PDF to generate a probabilistic forecast to quantify the uncertainty.

%=====================================================================

\paragraph{Residual-based Gaussian prediction intervals as a baseline}
In addition to delta, bootstrap, and Bayesian strategies that aim to represent model uncertainty explicitly,
a common \emph{engineering baseline} is to assume the remaining regression error is i.i.d.\ and
approximately Gaussian with constant variance estimated from held-out residuals.  This yields a
closed-form symmetric PI of the form \(\hat{y}\pm z_{1-\alpha/2}\hat{\sigma}_{\text{res}}\)
(cf.\ \cref{rv:eq:sigma_res,rv:eq:gauss_PI}).  The approach is transparent and computationally trivial, but
it does not quantify epistemic uncertainty and can be miscalibrated under heteroscedastic noise or
distribution shift \citep{Freedman2005, chatfield2000time}.

\subsection{Lower Upper Bound Estimation for Quantifying Uncertainty}
\label{rv:sec:lube-literature-survey}

In \citep{khosravi2010lower}, Khosravi et~al. introduce their novel Lower Upper Bound Estimation (LUBE) methodology to directly output prediction interval bounds as outputs of a neural network. Simulated annealing is applied to minimize the prediction interval-based cost function (interval width and coverage probability). The authors demonstrate success via ten benchmark regression case studies, and compare the performance against Bayesian, bootstrap, and delta methods.

In \citep{quan2013short}, Quan et~al. aim to forecast short-term load and wind power with quantifiable uncertainty using the neural network-based LUBE method. The network is optimized using particle swarm optimization. The novelty of the work is constraining the problem as a single-objective optimization task. Success is demonstrated via forecasting on electrical demand validation data.

In \citep{wan2013direct, wan2013optimal}, Wan et~al. put forth a novel framework that outputs prediction intervals of wind power generation without prior knowledge of forecasting errors (residuals). Their framework incorporates Extreme Learning Machine (ELM) and particle swarm optimization to directly optimize coverage probability and sharpness. The authors demonstrate success of their methodology via a case study using real-world wind farm data.

In \citep{quan2014particle}, Quan et~al. build upon \citep{khosravi2010lower} by introducing their particle-swarm-optimization-based (PSO) LUBE method with a novel cost function to comprehensively evaluate prediction intervals based on their width and coverage probability. Specifically, the authors replace the normalized mean prediction interval width (NMPIW) term with the normalized root-mean-square width (PINRW) term in the coverage-width criterion (CWC) calculation \cref{rv:eqn:cwc}. The incorporation of PSO into the LUBE method is used to optimize the objective function by finding the network weights that produce the optimal CWC evaluation metrics.

In \citep{zhang2014advanced}, Zhang et~al. use ensemble empirical mode decomposition (EEMD) to decompose wind power time series data into three independent intrinsic mode function (IMF) groups: a cycle component, a noise component, and a trend component. Afterward, the authors use an extreme learning method (ELM) to provide pointwise predictions for the well-behaved cycle and trend components, and use the same ELM to construct prediction intervals for the unstable noisy component. A quantum particle swarm optimization algorithm is used for optimizing the prediction intervals with regard to the cost function. The pointwise predictions and prediction intervals are combined to form overall optimal prediction intervals for the time series.

In \citep{hosen2014prediction}, Hosen et~al. apply a LUBE method to model and capture the nonlinear dynamics of a polymerization reactor. The authors note the inability of traditional neural networks to forecast in the face of data disturbances or perturbation, and consequently leverage a LUBE method to quantify uncertainties. Simulated annealing is used for optimization of network parameters with respect to the CWC cost function found in \citep{khosravi2010prediction}. The novelty of the presented framework is the network's ability to provide a prediction interval of the reactor temperature for one step ahead in time given a lookback window of three steps. The model has one hidden layer and is not a Long Short-Term Memory network.

%=====================================================================
\subsection{Mean Variance Estimation for Quantifying Uncertainty}
\label{rv:sec:mve-literature-survey}

In \citep{nix1994estimating}, Nix and Weigend introduces a novel feedforward neural network that outputs estimates of the mean and variance of a probability distribution of the target as a function of input data. Gaussian negative log-likelihood is used to minimize the cost function. Success is demonstrated via normalized mean squared error results for a synthetic problem where a true input-dependent noise level is known.

In \citep{khosravi2014optimized}, Khosravi et~al. apply the MVE method to wind power data for renewable-energy forecasting. Moreover, the authors optimize prediction intervals generated from an MVE network using the CWC cost function from \cref{rv:eqn:cwc}. Additionally, the prediction intervals of five confidence levels ranging from 50--90\% are compared between using traditional pointwise error minimization backpropagation techniques versus the interval optimization algorithm from \cref{rv:eqn:cwc}.

In \cite{pearce2018high}, Pearce et~al. provide an ensemble-based neural network framework for generating prediction intervals with the purpose of quantifying uncertainty in regression tasks. A strength of this approach is that it does not require an assumption of data distribution. Moreover, it optimizes the prediction interval to be as narrow as possible while still capturing the ground truth. \cite{pearce2018high} evaluates success by comparing prediction-interval quality metrics on ten regression datasets against approaches similar to \cite{kendall2017uncertainties}.

%=====================================================================
\subsection{Challenges and Future Directions}

Despite significant advancements, several challenges remain in the integration of machine learning with uncertainty quantification. Key issues include the computational cost of training ensemble models, demonstrating success on non-simulated datasets, and an overemphasis on optimizing prediction interval bounds without incorporating real-world risk management lessons. In other words, novel frameworks are developed with and demonstrated on neatly created simulated data. In turn, those frameworks are optimized to produce the smallest prediction intervals possible. However, in engineering applications, stakeholders may not even want or need optimized intervals, especially when dealing with high-risk worst-case scenarios. The work presented in this paper offers a comprehensive comparative evaluation of five main uncertainty quantification methods using machine learning, and applies them in a novel way to proprietary real-world data. In doing so, this work can serve as a practical guide for engineers seeking guidance on which uncertainty quantification methodology best suits their needs.

%=====================================================================
\section{Frameworks for Uncertainty Quantification Methodologies}
\label{rv:sec:uq-framework-methodology}
%=====================================================================

%=====================================================================
\subsection{Problem Setup}
\label{rv:sec:setup}
%=====================================================================

Let 
\(
  \mathcal{D}
  =\bigl\{(\mathbf{x}_{i},y_{i})\bigr\}_{i=1}^{n}
  \subset\mathbb{R}^{d}\!\times\!\mathbb{R}
\)
be an i.i.d.\ sample from an unknown distribution.
A neural regressor
\(
  f_{\boldsymbol{\theta}}
  :\mathbb{R}^{d}\!\to\!\mathbb{R},
  \;\boldsymbol{\theta}\in\mathbb{R}^{p},
\)
is trained on~$\mathcal{D}$, yielding weights
\(
  \hat{\boldsymbol{\theta}}
  =\arg\min_{\boldsymbol{\theta}}
     \frac1n\sum_{i=1}^{n}
     \ell\!\bigl(y_{i},f_{\boldsymbol{\theta}}(\mathbf{x}_{i})\bigr).
\)

For a new input \(\mathbf{x}_{*}\), the deterministic point forecast is denoted by
\(
  \hat{y}_{*}=f_{\hat{\boldsymbol{\theta}}}(\mathbf{x}_{*})
\),
and the goal is to construct a two-sided \((1-\alpha)\)-level prediction interval (PI)

\begin{equation}
\label{rv:eq:pi-definition}
  \mathcal{I}_{1-\alpha}(\mathbf{x}_{*})
  =\bigl[L_{1-\alpha}(\mathbf{x}_{*}),
         U_{1-\alpha}(\mathbf{x}_{*})\bigr]
  \quad\text{s.t.}\quad
  \Pr\!\bigl\{y_{*}\in\mathcal{I}_{1-\alpha}(\mathbf{x}_{*})\bigr\}
  =1-\alpha.
\end{equation}

that accounts for 

Aleatoric uncertainty ($\sigma^{2}_{\text{al}}$) denotes irreducible noise in the data. Epistemic uncertainty ($\sigma^{2}_{\text{ep}}$) denotes uncertainty in the learned parameters $\boldsymbol{\theta}$.

All methods below deliver
\(
  \hat{\mu}_{*}\equiv\hat{y}_{*}
\)
and an estimate
\(
  \hat{\sigma}^{2}_{*}
  =\hat{\sigma}^{2}_{\text{ep}}(\mathbf{x}_{*})
   +\hat{\sigma}^{2}_{\text{al}},
\)
from which the Gaussian-approximate interval

\begin{equation}
  \boxed{\;
    \mathcal{I}_{1-\alpha}(\mathbf{x}_{*})
    =\hat{\mu}_{*}\;\pm\;
     z_{1-\alpha/2}\,\hat{\sigma}_{*}
  \;}
  \label{rv:eq:gauss_PI}
\end{equation}

is obtained unless otherwise specified.
Here \(z_{1-\alpha/2}\) is the standard--normal quantile.

\paragraph{Optional distribution-free calibration.}
Several of the methods in this paper construct prediction intervals by assuming approximate normality of the predictive error (Eq.~\eqref{rv:eq:gauss_PI}) or by directly optimizing interval bounds. In safety-critical settings, it is often desirable to guarantee coverage empirically. A practical approach is to apply conformal calibration on top of any base predictor \citep{vovk2005alrw, shafer2008tutorial, romano2019cqr}. For split conformal regression, fit a base model on a training set, evaluate calibration residuals $r_i=|y_i-\hat{y}_i|$ on a held-out calibration set of size $n_{\mathrm{cal}}$, and set
\begin{equation}
\label{rv:eq:conformal-quantile}
  q_{1-\alpha} = \mathrm{Quantile}_{\lceil (n_{\mathrm{cal}}+1)(1-\alpha)\rceil/n_{\mathrm{cal}}}\bigl(\{r_i\}\bigr).
\end{equation}
Then the conformal interval is
\begin{equation}
\label{rv:eq:conformal-interval}
  \mathcal{I}_{1-\alpha}^{\mathrm{conf}}(\mathbf{x}_*)=
  [\hat{y}(\mathbf{x}_*)-q_{1-\alpha},\;\hat{y}(\mathbf{x}_*)+q_{1-\alpha}].
\end{equation}
This wrapper provides finite-sample marginal coverage under exchangeability, and can be paired with heteroscedastic quantile models via CQR to reduce conservatism \citep{romano2019cqr}.

\paragraph{Marginal vs.\ conditional coverage.} Split conformal regression provides a finite-sample \emph{marginal} coverage guarantee under exchangeability; it does not provide a conditional coverage guarantee and coverage can be non-uniform across covariates.

\paragraph{Aleatoric vs.\ epistemic.} Calibrates the base predictor's total error and does not explicitly separate aleatoric and epistemic contributions; any decomposition comes from the underlying base model.

\paragraph{Key approximations.} Requires a held-out calibration set and (approximate) exchangeability between calibration and test examples; temporal dependence, fleet/engine clustering, or distribution shift can invalidate the guarantee. Interval width can become conservative when calibration residuals are heavy-tailed or when the base model is mis-specified.

\subsection{Multilayer Perceptron (MLP) Architecture}
\label{rv:ssec:mlp}

Throughout this study the regression map  
\(f_{\boldsymbol{\theta}}:\mathbb{R}^{d}\!\to\!\mathbb{R}\)  
is instantiated as a fully connected \emph{multilayer perceptron (MLP)}
with $L\!-\!1$ hidden layers.  Let  

\begin{equation}
\label{rv:eq:mlp-layer-widths}
  m_{0}=d,\qquad
  m_{1},\dots,m_{L-1}\in\mathbb{N},
  \qquad
  m_{L}=1.
\end{equation}

denote the layer widths.  For an input
\(\mathbf{x}\in\mathbb{R}^{d}\) the forward pass is

\begin{align}
  \mathbf{h}^{(0)} &= \mathbf{x}, \label{rv:eq:mlp-forward-input}\\
  \mathbf{h}^{(\ell)} &=
     \sigma\!\bigl(
       \mathbf{W}^{(\ell)}\mathbf{h}^{(\ell-1)}
       +\mathbf{b}^{(\ell)}
     \bigr),
     \quad \ell=1,\dots,L-1, \label{rv:eq:mlp-forward-hidden}\\
  f_{\boldsymbol{\theta}}(\mathbf{x}) &=
     \mathbf{W}^{(L)}\mathbf{h}^{(L-1)}
     +\mathbf{b}^{(L)}, \label{rv:eq:mlp-forward-output}
\end{align}

where  \(\mathbf{W}^{(\ell)}\!\in\!\mathbb{R}^{m_{\ell}\times m_{\ell-1}}\)  
  and \(\mathbf{b}^{(\ell)}\!\in\!\mathbb{R}^{m_{\ell}}\)  
  are the weights and biases of layer~\(\ell\);  
\(\sigma(\cdot)\) is a fixed, element--wise nonlinearity  
  (ReLU unless stated otherwise).

The full parameter vector is  
\(
  \boldsymbol{\theta}
  =\bigl\{\mathbf{W}^{(\ell)},\mathbf{b}^{(\ell)}\bigr\}_{\ell=1}^{L}
  \in\mathbb{R}^{p},
\)
with  
\(p=\sum_{\ell=1}^{L} m_{\ell}(m_{\ell-1}+1)\).

\subsection{Mean--Squared Error for Pointwise Prediction Accuracy}
\label{rv:ssec:mse_pointwise}

Because the primary goal is to obtain accurate pointwise forecasts
(i.e., to make each individual prediction $f_{\boldsymbol{\theta}}(\mathbf
x_i)$ as close as possible to its corresponding target $y_i$), the methodology trains the
MLP by minimizing the \emph{mean--squared error} (MSE)

\begin{equation}
  \mathcal{L}_{\text{MSE}}(\boldsymbol{\theta})
  =
  \frac{1}{n}\sum_{i=1}^{n}
     \bigl(y_{i}-f_{\boldsymbol{\theta}}(\mathbf{x}_{i})\bigr)^{2}
  \;+\;
  \lambda\,\|\boldsymbol{\theta}\|_{2}^{2},
  \label{rv:eq:mse_point}
\end{equation}

where the first term penalizes the \textbf{squared deviation at each data
point}, and the $\ell_{2}$ regularizer ($\lambda\!\ge\!0$) guards against
overfitting.

\paragraph{Pointwise interpretation}
Equation~\eqref{rv:eq:mse_point} treats every observation
$(\mathbf{x}_{i},y_{i})$ independently and assigns \emph{equal weight} to
its squared error.  Minimizing this loss therefore drives
$f_{\boldsymbol{\theta}}(\mathbf{x}_{i})$ toward $y_{i}$ on a \emph{sample-by-sample
basis}, producing a model that excels at per-point accuracy rather than at
aggregate metrics such as trend fidelity or distributional fit.

\paragraph{Optimization}
Stochastic gradient descent (SGD) or Adam updates the weights:

\begin{equation}
\label{rv:eq:mse-optimizer-update}
  \boldsymbol{\theta}_{t+1}
  =
  \boldsymbol{\theta}_{t}
  -\eta_{t}\,
   \nabla_{\!\boldsymbol{\theta}}
   \mathcal{L}_{\text{MSE}}(\boldsymbol{\theta}_{t}).
\end{equation}

with learning rate $\eta_{t}$.  The converged solution

\begin{equation}
\label{rv:eq:mse-minimizer}
  \hat{\boldsymbol{\theta}}
  =\arg\min_{\boldsymbol{\theta}}
      \mathcal{L}_{\text{MSE}}(\boldsymbol{\theta}).
\end{equation}

yields residuals

\begin{equation}
\label{rv:eq:residual-definition}
  \varepsilon_{i}=y_{i}-f_{\hat{\boldsymbol{\theta}}}(\mathbf{x}_{i}),
\quad i=1,\dots,n.
\end{equation}

whose empirical variance

\begin{equation}
\label{rv:eq:residual-variance}
  \hat{\sigma}^{2}_{\text{res}}
  =\frac{1}{n}\sum_{i=1}^{n}\varepsilon_{i}^{2}.
\end{equation}

serves as the pointwise aleatoric noise estimate used later in
Sections \ref{rv:ssec:method_bootstrap}--\ref{rv:ssec:method_delta}.  Local
gradients

\begin{equation}
\label{rv:eq:delta-jacobian}
  \mathbf{J}_{g}(\hat{\boldsymbol{\theta}})
  =
  \nabla_{\!\boldsymbol{\theta}}
  f_{\boldsymbol{\theta}}(\mathbf{x}_{*})^{\!\top}
  \Bigl|_{\hat{\boldsymbol{\theta}}}.
\end{equation}

anchor the delta-method variance in
Section \ref{rv:ssec:method_delta}.  Thus, the MSE objective explicitly
conditions the weights on \emph{point-level accuracy}, providing the
foundation for all uncertainty estimates built on top of the trained
network.

%=====================================================================

%=====================================================================
\subsection{Gaussian Negative-Log-Likelihood Loss for Heteroscedastic Regression}
\label{rv:sec:gnll}
%=====================================================================

\subsubsection{Probabilistic Assumption}

To model \textbf{input-dependent (heteroscedastic) noise}, the formulation treats the
response as conditionally Gaussian,

\begin{equation}
  y \;\bigl|\;\mathbf{x}\;\sim\;
  \mathcal{N}\!\bigl(
      \mu_{\boldsymbol{\theta}}(\mathbf{x}),\;
      \sigma^{2}_{\boldsymbol{\theta}}(\mathbf{x})
  \bigr),
  \qquad
  \sigma^{2}_{\boldsymbol{\theta}}(\mathbf{x})>0,
  \label{rv:eq:gnll_model}
\end{equation}

where a single network with weights $\boldsymbol{\theta}$ produces \emph{two}
outputs for every input~$\mathbf{x}$:

\begin{equation}
\label{rv:eq:gnll-network-outputs}
  \mu_{\boldsymbol{\theta}}(\mathbf{x})        \in\mathbb{R},
  \qquad
  s_{\boldsymbol{\theta}}(\mathbf{x})
    := \log\sigma^{2}_{\boldsymbol{\theta}}(\mathbf{x})
    \in\mathbb{R}.
\end{equation}

A \texttt{softplus} or \(\exp\) is applied to guarantee
\(\sigma^{2}_{\boldsymbol{\theta}}(\mathbf{x})>0\).

\subsubsection{Derivation of the Loss}

The negative log-likelihood of a single observation
\((\mathbf{x}_{i},y_{i})\) under
\eqref{rv:eq:gnll_model} is

\begin{equation}
\label{rv:eq:gnll-observation-loss}
  \ell_{i}(\boldsymbol{\theta})
  = -\log
      p\!\bigl(y_{i}\mid\mathbf{x}_{i};\boldsymbol{\theta}\bigr)
  = \frac{1}{2}
      \Bigl[
         s_{\boldsymbol{\theta}}(\mathbf{x}_{i})
         + \frac{
             \bigl(y_{i}-\mu_{\boldsymbol{\theta}}(\mathbf{x}_{i})\bigr)^{2}}
             {e^{\,s_{\boldsymbol{\theta}}(\mathbf{x}_{i})}}
      \Bigr].
\end{equation}

Averaging over the data and adding weight decay gives the empirical
\emph{Gaussian negative-log-likelihood} (GNLL) objective

\begin{equation}
  \boxed{
    \mathcal{L}_{\text{GNLL}}(\boldsymbol{\theta})
    =
    \frac{1}{n}\sum_{i=1}^{n}
        \ell_{i}(\boldsymbol{\theta})
    + \lambda\,\|\boldsymbol{\theta}\|_{2}^{2}
  }
  \label{rv:eq:gnll_loss}
\end{equation}

with $\lambda\!\ge\!0$ controlling $\ell_{2}$ regularization.

\begin{table}
\centering
\caption{Components of the Gaussian NLL loss and their functional roles.}
\begin{tabular}{@{}p{0.43\linewidth}p{0.49\linewidth}@{}}
\toprule
\textbf{Term} & \textbf{Role} \\
\midrule
$\,s_{\boldsymbol{\theta}}(\mathbf{x}_{i})$ &
Penalizes overly \emph{large} predicted variances; acts as a complexity term that discourages inflation of $\sigma^{2}_{\boldsymbol{\theta}}(\mathbf{x}_{i})$. \\[4pt]

$\displaystyle
\frac{\bigl(y_{i}-\mu_{\boldsymbol{\theta}}(\mathbf{x}_{i})\bigr)^{2}}
     {e^{\,s_{\boldsymbol{\theta}}(\mathbf{x}_{i})}}$ &
Weights the squared residual by the inverse variance---
large errors (relative to the current $\sigma^{2}$) encourage the model to increase $\sigma^{2}_{\boldsymbol{\theta}}(\mathbf{x}_{i})$ where noise is high. \\
\bottomrule
\end{tabular}
\end{table}

Together, the two terms in the GNLL loss function push the network to output \emph{tight} variances in
well-behaved regions while inflating the variance where residuals
persist, thereby adapting to heteroscedastic noise patterns.

\subsection{Prediction and Uncertainty Decomposition}

After optimization, the formulation yields
\begin{equation}
\label{rv:eq:gnll-predictive-components}
  \hat{\mu}_{*} = \mu_{\hat{\boldsymbol{\theta}}}(\mathbf{x}_{*}),
  \quad
  \hat{\sigma}^{2}_{\text{al}}(\mathbf{x}_{*})
    = e^{\,s_{\hat{\boldsymbol{\theta}}}(\mathbf{x}_{*})}.
\end{equation}
which directly furnish the aleatoric component of the total predictive
variance.  When combined with an epistemic estimate
(e.g., deep ensembles or MC-Dropout) they slot into the generic
interval formula~\eqref{rv:eq:gauss_PI} to yield fully quantified
prediction intervals.

\paragraph{Contrast with MSE}
While the MSE loss of Section~\ref{rv:ssec:mse_pointwise} focuses solely on
pointwise accuracy, the GNLL simultaneously learns
\emph{which} inputs are noisy and by \emph{how much}, providing richer
uncertainty information essential for downstream risk-sensitive
decisions.

%=====================================================================
\begin{table*}
  \centering
  \caption{Summary of the uncertainty quantification (UQ) methods evaluated in this work, the notion of coverage they target, and their dominant assumptions.  Unless otherwise stated, empirical coverage metrics (e.g., PICP) estimate \emph{marginal} coverage on the held-out evaluation distribution; conditional coverage is only approximate and depends on model adequacy.}
  \label{rv:tab:uq_method_summary}
  \renewcommand{\arraystretch}{1.15}
  \scriptsize
  \setlength{\tabcolsep}{3pt}%
  \begin{adjustbox}{max width=\textwidth}
  \begin{tabular}{p{0.135\textwidth} p{0.175\textwidth} p{0.18\textwidth} p{0.42\textwidth}}
    \toprule
    \textbf{Method} & \textbf{Coverage notion} & \textbf{Uncertainty captured} & \textbf{Key approximations / assumptions} \\
    \midrule
    Bootstrap ensemble (fixed-data) &
    Empirical marginal coverage; conditional coverage only approximate &
    Primarily epistemic via model variability; aleatoric via an explicit residual/MVE term if included &
    Treats stochastic training (random initialization, mini-batch order) as generating a set of plausible predictors; ensemble members are not independent in a strict Bayesian sense; predictive Gaussian approximation if intervals use Eq.~\eqref{rv:eq:gauss_PI}. \\
    \addlinespace
    MC-Dropout (MC-DO) &
    Empirical marginal coverage; approximate Bayesian predictive inference &
    Epistemic via dropout weight sampling; aleatoric only if a noise term (e.g., MVE head or residual variance) is included &
    Interprets dropout as a variational approximation to a Bayesian neural network; assumes a fixed dropout rate approximates posterior uncertainty; Gaussian approximation when aggregating sampled predictions. \\
    \addlinespace
    Gaussian residual (GRM) &
    Approximate \emph{conditional} interval under a homoscedastic residual model; marginal coverage empirical &
    Aleatoric / residual uncertainty only (via $\hat{\sigma}^2_{\text{res}}$); epistemic absent unless combined with ensembles/MC-DO &
    Assumes conditionally independent residuals with (approximately) constant variance estimated on a calibration/validation set and often approximate Gaussianity; no distribution-free guarantee; sensitive to heteroscedasticity and distribution shift. \\
    \addlinespace
    Mean--Variance Estimation (MVE; Gaussian NLL) &
    Targets approximate \emph{conditional} coverage under the assumed likelihood model; marginal coverage depends on calibration &
    Aleatoric (possibly heteroscedastic) directly via $\hat{\sigma}^2_{\text{al}}(\mathbf{x})$; epistemic absent unless combined with ensembles/MC-DO &
    Assumes $Y\mid\mathbf{x}\sim\mathcal{N}(\mu(\mathbf{x}),\sigma^2(\mathbf{x}))$ and conditionally independent residuals; mis-specification or distribution shift can degrade coverage. \\
    \addlinespace
    LUBE (direct interval learning) &
    No explicit probabilistic guarantee; aims for empirical marginal coverage through a surrogate objective &
    Mixed: interval width implicitly absorbs both noise and model error; not decomposed into aleatoric/epistemic &
    Optimizes a non-proper surrogate (e.g., width-plus-penalty/CWC); training can be sensitive to hyperparameters; coverage assessed empirically rather than guaranteed. \\
    \addlinespace
    Delta method (first-order propagation) &
    Approximate \emph{conditional} interval under local linearization; marginal coverage empirical &
    Propagates input (and/or parameter) uncertainty through $\nabla f$; typically combined with an aleatoric residual term &
    First-order Taylor approximation; requires differentiability and a specified covariance (e.g., $\Sigma_{\mathbf{x}}$); often assumes approximate Gaussianity of propagated errors; ignores higher-order terms and strong nonlinearity. \\
    \addlinespace
    Conformal calibration (post-hoc wrapper) &
    Finite-sample \emph{marginal} coverage guarantee $\ge 1-\alpha$ under exchangeability; not a conditional guarantee &
    Calibrates the base method's intervals/quantiles to achieve marginal coverage; does not distinguish aleatoric vs epistemic by itself &
    Requires a held-out calibration set and (approximate) exchangeability between calibration and test; for time-series/fleet data often requires blocked/engine-level splits; may widen intervals when base predictions are miscalibrated. \\
    \bottomrule
  \end{tabular}
  \end{adjustbox}
\end{table*}

%=====================================================================
\subsection{Lower Upper Bound Estimation}
%\subsection{Width--Plus--Penalty Objective for Direct Interval Learning}
\label{rv:sec:width_penalty}
%=====================================================================

\subsubsection{Loss Definition}

In the Lower--Upper Bound Estimation (LUBE) framework the network
outputs two real-valued heads
\(
  L_{\boldsymbol{\phi}}(\mathbf{x})
\)
and
\(
  U_{\boldsymbol{\phi}}(\mathbf{x})
\).
Given a mini-batch
\(\mathcal{B}=\{(\mathbf{x}_{j},y_{j})\}_{j=1}^{m}\),
the \textbf{width-plus-penalty} loss is

\begin{equation}
  \boxed{
  \begin{aligned}
    \mathcal{L}_{\text{WP}}(\boldsymbol{\phi})
    &=
    \underbrace{
      \frac{1}{m}\sum_{j=1}^{m}
        \bigl[
          U_{\boldsymbol{\phi}}(\mathbf{x}_{j})
          -L_{\boldsymbol{\phi}}(\mathbf{x}_{j})
        \bigr]
    }_{\text{mean interval width }W}\\
    \;&+\;
    \lambda_{\text{LUBE}}\,
    \underbrace{
      \frac{1}{m}\sum_{j=1}^{m}
        \Bigl[
          \bigl(L_{\boldsymbol{\phi}}(\mathbf{x}_{j})-y_{j}\bigr)_{+}
          +\bigl(y_{j}-U_{\boldsymbol{\phi}}(\mathbf{x}_{j})\bigr)_{+}
        \Bigr]
    }_{\text{violation penalty }P}
    \end{aligned}
  }
  \label{rv:eq:width_penalty}
\end{equation}

with \((u)_{+}=\max\{0,u\}\) and tuning constant
\(\lambda_{\text{LUBE}}>0\).

\begin{table}
\centering
\begin{tabular}{@{}p{0.42\linewidth}p{0.50\linewidth}@{}}
\toprule
\textbf{Component} & \textbf{Effect on Training} \\
\midrule
\(\displaystyle
  W = \frac{1}{m}\sum_{j}[U(\mathbf{x}_{j})-L(\mathbf{x}_{j})]
\) &
Pushes the model to \emph{narrow} the prediction interval, improving
sharpness. \\[6pt]

\(\displaystyle
  P = \frac{1}{m}\sum_{j}
      \bigl[(L(\mathbf{x}_{j})-y_{j})_{+}+(y_{j}-U(\mathbf{x}_{j}))_{+}\bigr]
\) &
Accumulates a non-zero value \emph{only} when the true target lies
outside the current interval, encouraging the bounds to expand until
coverage is achieved. \\[6pt]

\(\lambda_{\text{LUBE}}\) (trade-off weight) &
Balances calibration against sharpness:
large \(\lambda_{\text{LUBE}}\) favors coverage (wider intervals),
small \(\lambda_{\text{LUBE}}\) favors tightness (risking under-coverage). \\
\bottomrule
\end{tabular}
\end{table}

\subsubsection{Optimization Dynamics}

During early epochs, the penalty term dominates because many points are uncovered. Therefore, intervals expand rapidly.
In contrast, during later epochs the violations vanish, the width term governs and the optimizer tightens bounds where data allow.
The loss is differentiable almost everywhere (thanks to
  ReLU\,/\,\((\cdot)_{+}\)), so standard optimizers (Adam, SGD) apply.

\subsubsection{Resulting Prediction Interval}

After convergence,
\(
  \mathcal{I}_{\text{LUBE}}(\mathbf{x}_{*})
  =
  \bigl[
    L_{\boldsymbol{\phi}^{\star}}(\mathbf{x}_{*}),\;
    U_{\boldsymbol{\phi}^{\star}}(\mathbf{x}_{*})
  \bigr]
\)
is produced in \emph{one} forward pass---no additional sampling or
post-processing required.

Compared with approaches that propagate variance through the model,
Eq.~\eqref{rv:eq:width_penalty} directly targets the interval
as the primary output, making it distribution-free and well suited when
noise characteristics are unknown or heavy-tailed.

%=====================================================================
\subsection{Bootstrap Ensemble (Fixed--Data Model Bootstrap)}
\label{rv:ssec:method_bootstrap}
%=====================================================================

Train \(B\) replicas of the model on \emph{identical} data but with independent random seeds. This approach is often referred to as a \emph{deep ensemble}; the term ``fixed-data bootstrap'' is used to emphasize that the resampling occurs through optimization stochasticity rather than through resampling the dataset:

\(
  \widehat{\boldsymbol{\theta}}^{(b)},\;b=1,\dots,B.
\)
The ensemble moments are

\begin{equation}
\label{rv:eq:bootstrap-moments}
  \hat{\mu}_{*}
  =\frac1B\sum_{b=1}^{B}
     f_{\widehat{\boldsymbol{\theta}}^{(b)}}(\mathbf{x}_{*}),
  \qquad
  \hat{\sigma}^{2}_{\text{ep}}(\mathbf{x}_{*})
  =\frac1B\sum_{b=1}^{B}
     f_{\widehat{\boldsymbol{\theta}}^{(b)}}(\mathbf{x}_{*})^{2}
   -\hat{\mu}_{*}^{2}.
\end{equation}

A separate residual MSE on a validation set supplies
\(\hat{\sigma}^{2}_{\text{al}}\).
Insert these quantities into~\eqref{rv:eq:gauss_PI}.
This \textit{model bootstrap} scales well across hardware and has been
shown to yield calibrated PIs for deep nets
\citep{Lakshminarayanan2017DeepEns}.

\paragraph{Marginal vs.\ conditional coverage.} Intervals target an \emph{approximate conditional} PI under a Gaussian approximation to the ensemble predictive distribution (Eq.~\eqref{rv:eq:gauss_PI}); reported PICP is an \emph{empirical marginal} coverage estimate on the evaluation distribution.

\paragraph{Aleatoric vs.\ epistemic.} Captures \emph{epistemic} uncertainty through variability across ensemble members via $\hat{\sigma}^{2}_{\text{ep}}(\mathbf{x}_{*})$; \emph{aleatoric} uncertainty is represented only through an explicit noise term (e.g., a residual/MVE estimate of $\hat{\sigma}^{2}_{\text{al}}$).

\paragraph{Key approximations.} Treats stochastic training (random initialization, mini-batch order, etc.) as generating a set of plausible predictors; ensemble members are not independent draws from a strict Bayesian posterior. When Eq.~\eqref{rv:eq:gauss_PI} is used, the predictive distribution is summarized by its first two moments and an approximate Gaussian error model; calibration can degrade under correlated ensemble errors, heteroscedastic residuals, or distribution shift.

%---------------------------------------------------------------------
\subsection{Monte-Carlo Dropout (MC-DO)}
\label{rv:ssec:method_mcdo}

Keep dropout active at test time and perform
\(T\) stochastic forward passes:

\begin{equation}
\label{rv:eq:mcdo-forward-passes}
  \hat{y}_{*}^{(t)}
  =f_{\boldsymbol{W}^{(t)}}(\mathbf{x}_{*}),
  \qquad t=1,\dots,T.
\end{equation}

where \(\boldsymbol{W}^{(t)}\) embodies a fresh set of dropout masks.
Compute

\begin{equation}
\label{rv:eq:mcdo-moments}
  \hat{\mu}_{*}
  =\tfrac1T\sum_{t=1}^{T}\hat{y}_{*}^{(t)},
  \qquad
  \hat{\sigma}^{2}_{\text{ep}}(\mathbf{x}_{*})
  =\tfrac1T\sum_{t=1}^{T}(\hat{y}_{*}^{(t)})^{2}
   -\hat{\mu}_{*}^{2}.
\end{equation}

Estimate \(\hat{\sigma}^{2}_{\text{al}}\) as above and apply
\eqref{rv:eq:gauss_PI}.
MC-DO offers a variational Bayesian approximation with negligible
memory overhead
\citep{GalGhahramani2016}.

\paragraph{Marginal vs.\ conditional coverage.} Intervals target an \emph{approximate conditional} PI under the dropout-induced predictive distribution, summarized by Monte-Carlo moments; reported PICP is an \emph{empirical marginal} coverage estimate on the evaluation distribution.

\paragraph{Aleatoric vs.\ epistemic.} Captures \emph{epistemic} uncertainty through dropout weight sampling via $\hat{\sigma}^{2}_{\text{ep}}(\mathbf{x}_{*})$; \emph{aleatoric} uncertainty is represented only if an explicit noise term (e.g., residual variance or an MVE head) is added.

\paragraph{Key approximations.} Interprets dropout as a variational approximation to a Bayesian neural network, with quality depending on the chosen dropout rate and variational family. Finite Monte-Carlo sampling ($T$) introduces estimator noise, and the Gaussian aggregation in Eq.~\eqref{rv:eq:gauss_PI} assumes approximate normality; miscalibration can occur under strong nonlinearity, heteroscedasticity, or distribution shift.

%---------------------------------------------------------------------

%---------------------------------------------------------------------
\subsection{Gaussian Residual Model (Residual-Only Baseline)}
\label{rv:ssec:method_grm}

A simple and widely used baseline for regression prediction intervals is the
\emph{Gaussian residual model}: fit a deterministic predictor and treat the remaining
error as homoscedastic Gaussian noise with variance estimated from held-out residuals
\citep{Freedman2005, chatfield2000time}.  Let $\mathcal{D}_{\mathrm{cal}}$ denote a
calibration/validation set that is disjoint from the data used to fit
$\hat{\boldsymbol{\theta}}$.  Define residuals
\(
  r_i = y_i - f_{\hat{\boldsymbol{\theta}}}(\mathbf{x}_i)
\)
for $(\mathbf{x}_i,y_i)\in\mathcal{D}_{\mathrm{cal}}$ and estimate

\begin{equation}
  \hat{\sigma}^2_{\text{res}}
  =
  \frac{1}{|\mathcal{D}_{\mathrm{cal}}|-1}
  \sum_{(\mathbf{x}_i,y_i)\in\mathcal{D}_{\mathrm{cal}}}
  \bigl(y_i - f_{\hat{\boldsymbol{\theta}}}(\mathbf{x}_i)\bigr)^2 .
  \label{rv:eq:sigma_res}
\end{equation}

The predictive model is then
\(
  Y_{*}\mid \mathbf{X}_{*}=\mathbf{x}_{*}
  \approx
  \mathcal{N}\!\bigl(f_{\hat{\boldsymbol{\theta}}}(\mathbf{x}_{*}),\hat{\sigma}^2_{\text{res}}\bigr),
\)
so that Eq.~\eqref{rv:eq:gauss_PI} applies with
\(
  \hat{\mu}_{*}=f_{\hat{\boldsymbol{\theta}}}(\mathbf{x}_{*})
\)
and
\(
  \hat{\sigma}_{*}=\hat{\sigma}_{\text{res}}.
\)

\paragraph{Marginal vs.\ conditional coverage.} Intervals target an \emph{approximate conditional} PI under the assumed noise model; reported PICP is an \emph{empirical marginal} coverage estimate on the evaluation distribution.

\paragraph{Aleatoric vs.\ epistemic.} Captures only \emph{aleatoric / residual} uncertainty through $\hat{\sigma}^2_{\text{res}}$; epistemic uncertainty in $\hat{\boldsymbol{\theta}}$ is not represented unless an ensemble/MC-DO component is added.

\paragraph{Key approximations.} Assumes conditionally independent residuals, (approximately) constant variance (unless $\hat{\sigma}^2_{\text{res}}$ is estimated by regime/phase), and approximate Gaussianity; it does not provide a distribution-free guarantee and can be miscalibrated under heteroscedasticity or distribution shift.

\paragraph{Relation to the delta method.}
The Gaussian residual baseline is a special case of first-order propagation when
input/parameter uncertainty terms are neglected (e.g., $\boldsymbol{\Sigma}_{\mathbf{x},*}=\mathbf{0}$),
so that the only variance contribution is $\hat{\sigma}^2_{\text{res}}$.

\subsection{Delta Method (First-Order Uncertainty Propagation)}
\label{rv:ssec:method_delta}

The \emph{delta method} can be used to propagate uncertainty through a deterministic neural network by linearizing the model locally.  In engineering practice, a common use case is uncertainty in the \emph{inputs} (e.g., sensor noise, operating-condition variability) rather than a full Bayesian posterior over parameters.  Let the test input be modeled as a random vector
\(
  \mathbf{X}_{*}\sim(\mathbf{x}_{*},\boldsymbol{\Sigma}_{\mathbf{x},*}),
\)
where $\boldsymbol{\Sigma}_{\mathbf{x},*}$ is an assumed or estimated input covariance matrix (from sensor specifications, repeated measurements, or empirical variability after normalization).  A first-order Taylor expansion gives
\(
  f_{\hat{\boldsymbol{\theta}}}(\mathbf{X}_{*})
  \approx
  f_{\hat{\boldsymbol{\theta}}}(\mathbf{x}_{*})
  + \nabla_{\!\mathbf{x}} f_{\hat{\boldsymbol{\theta}}}(\mathbf{x}_{*})^{\!\top}
    (\mathbf{X}_{*}-\mathbf{x}_{*}),
\)
so that the propagated variance is
\begin{equation}
  \hat{\sigma}^{2}_{\text{in}}(\mathbf{x}_{*})
  =
  \nabla_{\!\mathbf{x}} f_{\hat{\boldsymbol{\theta}}}(\mathbf{x}_{*})^{\!\top}
  \boldsymbol{\Sigma}_{\mathbf{x},*}
  \nabla_{\!\mathbf{x}} f_{\hat{\boldsymbol{\theta}}}(\mathbf{x}_{*}).
  \label{rv:eq:delta_input}
\end{equation}

To account for residual model mismatch and observation noise, an empirical residual variance is added $\hat{\sigma}^{2}_{\text{res}}$ estimated on a validation set (Section~\ref{rv:ssec:mse_pointwise}).  The total predictive variance is then
\(
  \hat{\sigma}^{2}_{*}
  =\hat{\sigma}^{2}_{\text{in}}(\mathbf{x}_{*})+\hat{\sigma}^{2}_{\text{res}}.
\)
Finally, a Gaussian-approximate PI is constructed using Eq.~\eqref{rv:eq:gauss_PI}.
The method is analytic and fast, but hinges on local linearity and on the fidelity of the input-noise model \citep{van2000asymptotic}.

\paragraph{Marginal vs.\ conditional coverage.} Intervals target an \emph{approximate conditional} PI under local linearization and Gaussian error propagation; reported PICP is an \emph{empirical marginal} coverage estimate on the evaluation distribution.

\paragraph{Aleatoric vs.\ epistemic.} Captures uncertainty from \emph{input variability} through $\hat{\sigma}^{2}_{\text{in}}(\mathbf{x}_{*})$ and from \emph{residual/observation noise} through $\hat{\sigma}^{2}_{\text{res}}$; epistemic uncertainty in $\hat{\boldsymbol{\theta}}$ is not represented unless a parameter-covariance model or an ensemble/MC-DO component is incorporated.

\paragraph{Key approximations.} Relies on a first-order Taylor expansion and requires a specified or estimated input covariance $\boldsymbol{\Sigma}_{\mathbf{x},*}$; higher-order terms are ignored and uncertainty can be underestimated for strongly nonlinear mappings. The Gaussian PI of Eq.~\eqref{rv:eq:gauss_PI} further assumes approximate normality of the propagated error; mis-specified $\boldsymbol{\Sigma}_{\mathbf{x},*}$, heteroscedastic residuals, or distribution shift can degrade calibration.

\subsection{Lower-Upper Bound Estimation (LUBE)}
\label{rv:ssec:method_lube}

A network outputs two scalars
\(
  (L_{\boldsymbol{\phi}}(\mathbf{x}),
   U_{\boldsymbol{\phi}}(\mathbf{x})).
\)
Training minimizes

\begin{equation}
\label{rv:eq:lube-summary-loss}
\begin{split}
    \mathcal{L}_{\text{LUBE}}
  &=\!
  \underbrace{
    \mathbb{E}_{\mathcal{B}}
      \bigl[U_{\boldsymbol{\phi}}(\mathbf{x})-
            L_{\boldsymbol{\phi}}(\mathbf{x})\bigr]
  }_{\text{mean width}} \\
  &\quad +\lambda_{\text{LUBE}}\,
   \underbrace{
    \mathbb{E}_{\mathcal{B}}
      \bigl[
        (L_{\boldsymbol{\phi}}(\mathbf{x})-y)_{+}
        +(y-U_{\boldsymbol{\phi}}(\mathbf{x}))_{+}
      \bigr]
  }_{\text{violation penalty}}.
\end{split}
\end{equation}

where \(\mathcal{B}\) denotes the mini-batch and \(\lambda_{\text{LUBE}}>0\) tunes
coverage versus width.
The PI for~\(\mathbf{x}_{*}\) is
\(
  [L_{\boldsymbol{\phi}^{*}}(\mathbf{x}_{*}),
   U_{\boldsymbol{\phi}^{*}}(\mathbf{x}_{*})]
\)
with no further computation
\citep{khosravi2010lower}.

\paragraph{Marginal vs.\ conditional coverage.} Learned bounds aim for \emph{empirical marginal} coverage via the surrogate width--penalty objective; no distribution-free guarantee is provided and conditional coverage can vary across the input space.

\paragraph{Aleatoric vs.\ epistemic.} Interval width implicitly absorbs both \emph{aleatoric} noise and \emph{epistemic} model error; the method does not decompose these components unless paired with a separate epistemic estimator.

\paragraph{Key approximations.} Optimizes a non-proper surrogate whose behavior depends on $\lambda_{\text{LUBE}}$ and on the optimization dynamics; training can be sensitive to hyperparameters and to data imbalance. Coverage is assessed empirically and can degrade under distribution shift or when the loss encourages overly conservative or overly narrow intervals in sparsely sampled regimes.

%---------------------------------------------------------------------
\subsection{Mean-Variance Estimation (Gaussian NLL)}
\label{rv:ssec:method_mve}

The network predicts
\(
  \mu_{\boldsymbol{\theta}}(\mathbf{x}),
  \;
  s_{\boldsymbol{\theta}}(\mathbf{x})
   =\log\sigma^{2}_{\boldsymbol{\theta}}(\mathbf{x}).
\)
The loss is the Gaussian NLL

\begin{equation}
\label{rv:eq:mve-loss}
  \mathcal{L}_{\text{MVE}}
  =\frac1n\sum_{i=1}^{n}
     \tfrac12
     \Bigl[
       s_{\boldsymbol{\theta}}(\mathbf{x}_{i})
       +\frac{(y_{i}-\mu_{\boldsymbol{\theta}}(\mathbf{x}_{i}))^{2}}
             {e^{s_{\boldsymbol{\theta}}(\mathbf{x}_{i})}}
     \Bigr].
\end{equation}

After optimization,

\begin{equation}
\label{rv:eq:mve-predictive-components}
  \hat{\mu}_{*}
  =\mu_{\hat{\boldsymbol{\theta}}}(\mathbf{x}_{*}),
  \qquad
  \hat{\sigma}^{2}_{*}
  =e^{s_{\hat{\boldsymbol{\theta}}}(\mathbf{x}_{*})}.
\end{equation}

and \eqref{rv:eq:gauss_PI} gives the PI.  
The method captures input-dependent aleatoric uncertainty; pairing with
an ensemble or MC-DO yields full UQ
\citep{nix1994estimating}.

\paragraph{Marginal vs.\ conditional coverage.} Intervals target an \emph{approximate conditional} PI under the assumed Gaussian likelihood with learned variance; reported PICP is an \emph{empirical marginal} coverage estimate on the evaluation distribution and depends on the calibration of $\hat{\sigma}^{2}_{*}$.

\paragraph{Aleatoric vs.\ epistemic.} Captures \emph{aleatoric} (possibly heteroscedastic) uncertainty through $\hat{\sigma}^{2}_{*}=e^{s_{\hat{\boldsymbol{\theta}}}(\mathbf{x}_{*})}$; epistemic uncertainty in $\hat{\boldsymbol{\theta}}$ is not represented unless an ensemble/MC-DO component is added.

\paragraph{Key approximations.} Assumes conditionally independent targets with \(Y\mid\mathbf{x}\sim\mathcal{N}(\mu(\mathbf{x}),\sigma^{2}(\mathbf{x}))\); likelihood mis-specification, heavy tails, or distribution shift can lead to miscalibration. The Gaussian PI of Eq.~\eqref{rv:eq:gauss_PI} also relies on approximate normality of standardized residuals and on stable variance predictions outside the training distribution.

%=====================================================================
\section{Predicting Turbine Gas Temperature with Uncertainty}
\label{rv:sec:tgt_application}
%=====================================================================

Throughout this section the response variable is
\begin{equation}
\label{rv:eq:tgt-response-definition}
  y_i \;\equiv\; \mathrm{TGT}_i.
\end{equation}
the \emph{turbine gas temperature} measured (in $^{\circ}$C) at the
exhaust of a gas-turbine engine for operating point
$\mathbf{x}_i$.  The feature vector
$\mathbf{x}_i\!\in\!\mathbb{R}^{d}$ collects engine-health indicators
(pressure ratios, shaft speeds, fuel flow, ambient conditions, etc.)
sampled at the same instant.  Accurate \emph{pointwise} prediction of
TGT and well-calibrated \emph{uncertainty bands} are critical for
thermal-stress monitoring and remaining-useful-life estimation.

In the notation of \cref{rv:sec:setup}, the TGT data form the regression
sample
\begin{equation}
  \mathcal{D}_{\mathrm{TGT}}
  =\bigl\{(\mathbf{x}_{i},y_{i})\bigr\}_{i=1}^{n},
  \qquad
  y_{i} = \mathrm{TGT}_{i},
  \label{rv:eq:tgt_dataset}
\end{equation}
where each $\mathbf{x}_{i}\in\mathbb{R}^{d}$ encodes the engine state at a
given time.  Flights are partitioned into disjoint subsets
$\mathcal{D}_{\text{train}}$, $\mathcal{D}_{\text{val}}$ and
$\mathcal{D}_{\text{test}}$; all model parameters and
hyperparameters are fitted on $\mathcal{D}_{\text{train}}$ and tuned using
$\mathcal{D}_{\text{val}}$, while reported performance and coverage metrics
are evaluated on $\mathcal{D}_{\text{test}}$.

The deterministic regression map $f_{\boldsymbol{\theta}}$ is
instantiated as the multilayer perceptron in \cref{rv:ssec:mlp} and
trained with the mean--squared error objective
$\mathcal{L}_{\text{MSE}}$ of \cref{rv:ssec:mse_pointwise}.  Given a new
operating point $\mathbf{x}_{*}$, each UQ methodology from
\cref{rv:sec:uq-framework-methodology} delivers a predictive mean
$\hat{\mu}_{*}$ together with an estimate of the total predictive
variance,
\begin{equation}
\label{rv:eq:total-predictive-variance}
  \hat{\sigma}_{*}^{2}
  =
  \hat{\sigma}^{2}_{\text{ep}}(\mathbf{x}_{*})
  +\hat{\sigma}^{2}_{\text{al}}(\mathbf{x}_{*}).
\end{equation}
where $\hat{\sigma}^{2}_{\text{ep}}$ and $\hat{\sigma}^{2}_{\text{al}}$
represent epistemic and aleatoric contributions respectively.  These
quantities are inserted into the Gaussian-approximate interval
$\mathcal{I}_{1-\alpha}(\mathbf{x}_{*})$ defined in
\cref{rv:eq:gauss_PI} to obtain pointwise uncertainty bands along an
entire flight profile.  The remainder of this section details how each
generic construction in \cref{rv:sec:uq-framework-methodology} is
specialized to $\mathcal{D}_{\mathrm{TGT}}$.

%---------------------------------------------------------------------
\subsection{Bootstrap Ensemble for TGT}
\label{rv:ssec:tgt_bootstrap}

The fixed--data model bootstrap of \cref{rv:ssec:method_bootstrap}
approximates epistemic uncertainty by training $B$ independently
initialized replicas of $f_{\boldsymbol{\theta}}$ on the same
dataset.  For $\mathcal{D}_{\mathrm{TGT}}$, the corresponding ensemble
\(
  \{f_{\widehat{\boldsymbol{\theta}}^{(b)}}\}_{b=1}^{B}
\)
induces, for each $\mathbf{x}_{*}$, an empirical mean and variance
\(
  \hat{\mu}_{*}
\)
and
\(
  \hat{\sigma}^{2}_{\text{ep}}(\mathbf{x}_{*})
\)
as defined in \cref{rv:ssec:method_bootstrap}.  Combined with an estimate
of the aleatoric variance from residuals, these enter
\cref{rv:eq:gauss_PI} to form prediction intervals.  The concrete TGT
instantiation is:

\begin{enumerate}[leftmargin=2.2em,label=\arabic*.]
  \item \textbf{Model definition.}  
        Use the baseline MLP from \cref{rv:ssec:mlp} with
        ReLU--5--ReLU--3 hidden layers, trained on
        $\mathcal{L}_{\text{MSE}}$ as defined in
        \cref{rv:ssec:mse_pointwise}.  
  \item \textbf{Ensemble creation.}  
        Train $B=10$ replicas, each with independent
        weight initialization and shuffled mini-batches.
  \item \textbf{Prediction for a flight-cycle sample
        $\mathbf{x}_{*}$.}  
        Compute the ensemble mean and variance
        $\hat{\mu}_{*},\hat{\sigma}^{2}_{\text{ep}}(\mathbf{x}_{*})$
        using the expressions in \cref{rv:ssec:method_bootstrap}.  Obtain
        $\hat{\sigma}^{2}_{\text{al}}$ from a validation residual MSE
        restricted to steady-state segments (to avoid transient
        spikes).
  \item \textbf{Interval.}  
        Form the Gaussian interval $\mathcal{I}_{1-\alpha}(\mathbf{x}_{*})$
        via \cref{rv:eq:gauss_PI}.  
        These intervals tighten at low-load conditions (where sensor
        noise is lower) and widen during high-temperature transients
        (where models disagree).
\end{enumerate}

%---------------------------------------------------------------------
\subsection{MC-Dropout for TGT}
\label{rv:ssec:tgt_mcdo}

MC-Dropout specializes the variational Bayesian approximation of
\cref{rv:ssec:method_mcdo} to the TGT setting by interpreting the random
dropout masks as samples from an approximate posterior over network
weights.  For each operating point $\mathbf{x}_{*}$, repeated
stochastic forward passes yield Monte-Carlo estimates of
$\hat{\mu}_{*}$ and $\hat{\sigma}^{2}_{\text{ep}}(\mathbf{x}_{*})$
which again enter \cref{rv:eq:gauss_PI} after adding an aleatoric term.
The resulting procedure is:

\begin{enumerate}[leftmargin=2.2em,label=\arabic*.]
  \item \textbf{Network modification.}  
        Insert dropout ($p=0.2$) after each hidden layer of the
        baseline MLP.
  \item \textbf{Training.}  
        Optimize the MSE loss $\mathcal{L}_{\text{MSE}}$ with dropout
        enabled on $\mathcal{D}_{\text{train}}$.  
  \item \textbf{Inference.}  
        Keep dropout active; draw $T=50$ stochastic passes to obtain
        $\hat{\mu}_{*},\hat{\sigma}^{2}_{\text{ep}}(\mathbf{x}_{*})$
        as in \cref{rv:ssec:method_mcdo}.  
        Estimate $\hat{\sigma}^{2}_{\text{al}}$ from residuals
        segmented by engine regime (idle, climb, cruise) because TGT
        noise is regime-dependent.  Construct
        $\mathcal{I}_{1-\alpha}(\mathbf{x}_{*})$ using
        \cref{rv:eq:gauss_PI}.
  \item \textbf{Result.}  
        MC-Dropout yields smooth uncertainty envelopes along an entire
        flight profile with negligible memory overhead.
\end{enumerate}

%---------------------------------------------------------------------
%---------------------------------------------------------------------
\subsection{Gaussian Residual Baseline (Residual-Only Delta) for TGT}
\label{rv:ssec:tgt_delta}

For the proprietary TGT dataset, a fully specified input covariance
$\boldsymbol{\Sigma}_{\mathbf{x},*}$ is not available for all features after
preprocessing and normalization.  Therefore, the ``Delta'' baseline evaluated in
this work corresponds to the \emph{residual-only} Gaussian model described in
\cref{rv:ssec:method_grm}.  Equivalently, it is the special case of
\cref{rv:ssec:method_delta} obtained by setting
$\boldsymbol{\Sigma}_{\mathbf{x},*}=\mathbf{0}$ in \cref{rv:eq:delta_input}, so that
$\hat{\sigma}^{2}_{\text{in}}(\mathbf{x}_{*})=0$ and the total predictive
variance reduces to the residual term $\hat{\sigma}^{2}_{\text{res}}$. The implementation is as follows:

\begin{enumerate}[leftmargin=2.2em,label=\arabic*.]
  \item \textbf{Point predictor.}
        Train the deterministic MLP regressor
        $f_{\hat{\boldsymbol{\theta}}}$ on $\mathcal{D}_{\text{train}}$
        using the MSE loss of \cref{rv:ssec:mse_pointwise}.
  \item \textbf{Residual-variance estimation.}
        Compute held-out residuals
        $r_i = y_i - f_{\hat{\boldsymbol{\theta}}}(\mathbf{x}_i)$
        on $\mathcal{D}_{\text{val}}$ (or $\mathcal{D}_2$ in the experiment
        split) and estimate $\hat{\sigma}^{2}_{\text{res}}$ via
        \cref{rv:eq:sigma_res}.  In practice, this estimate is restricted to
        quasi-steady segments to reduce contamination from transient model bias.
  \item \textbf{Prediction interval.}
        For each test input $\mathbf{x}_{*}$, set
        $\hat{\mu}_{*}=f_{\hat{\boldsymbol{\theta}}}(\mathbf{x}_{*})$ and
        $\hat{\sigma}_{*}=\hat{\sigma}_{\text{res}}$, then construct
        $\mathcal{I}_{1-\alpha}(\mathbf{x}_{*})$ using \cref{rv:eq:gauss_PI}.
\end{enumerate}

This baseline is computationally trivial and provides a transparent reference
for interval width and empirical coverage.  Its main limitation is that it
captures only \emph{aleatoric / residual} uncertainty; epistemic effects (model
uncertainty across engines and regimes) are not represented unless an additional
ensemble or MC-Dropout component is incorporated.

\paragraph{Optional extension.}
If reliable estimates of $\boldsymbol{\Sigma}_{\mathbf{x},*}$ are available (e.g.,
from sensor specifications mapped through preprocessing), the full delta-method
propagation term in \cref{rv:eq:delta_input} can be added to obtain
$\hat{\sigma}^{2}_{*}=\hat{\sigma}^{2}_{\text{in}}(\mathbf{x}_{*})+\hat{\sigma}^{2}_{\text{res}}$.

\subsection{LUBE for TGT}
\label{rv:ssec:tgt_lube}

Within the Lower--Upper Bound Estimation framework of
\cref{rv:ssec:method_lube,rv:sec:width_penalty}, the network is trained to
output interval bounds
$L_{\boldsymbol{\phi}}(\mathbf{x})$ and $U_{\boldsymbol{\phi}}(\mathbf{x})$
directly and to minimize the width--plus--penalty loss
$\mathcal{L}_{\text{WP}}$ of \cref{rv:eq:width_penalty}.  For TGT this
yields a lightweight architecture that can be deployed on embedded
hardware:

\begin{enumerate}[leftmargin=2.2em,label=\arabic*.]
  \item \textbf{Architecture.}  
        Shared hidden layers of width 8 feed two linear heads; enforce
        $U=L+\operatorname{softplus}(w)$ to guarantee non-negative
        interval width.
  \item \textbf{Loss setup.}  
        Use the width--plus--penalty loss
        $\mathcal{L}_{\text{WP}}$ in \cref{rv:eq:width_penalty} with
        $\lambda_{\text{LUBE}}=75$.  
        Mini-batches are stratified by thrust level so that the
        penalty sees a balanced mix of high- and low-variance
        operating points.
  \item \textbf{Training.}  
        Train for 200 epochs with Adam; monitor empirical coverage on a
        held-out flight to retune $\lambda_{\text{LUBE}}$.
  \item \textbf{Deployment.}  
        At inference the interval
        $\bigl[L(\mathbf{x}_{*}),U(\mathbf{x}_{*})\bigr]$ is produced
        in one forward pass, providing a distribution-free estimate of
        $\mathcal{I}_{1-\alpha}(\mathbf{x}_{*})$ that is well suited
        for embedded engine control units.
\end{enumerate}

%---------------------------------------------------------------------
\subsection{MVE with Gaussian NLL for TGT}
\label{rv:ssec:tgt_mve}

The mean--variance estimation strategy of \cref{rv:ssec:method_mve} uses
the heteroscedastic Gaussian likelihood in \cref{rv:sec:gnll} to learn a
mean function $\mu_{\boldsymbol{\theta}}(\mathbf{x})$ and an
input-dependent variance $\sigma^{2}_{\boldsymbol{\theta}}(\mathbf{x})$
jointly by minimizing the GNLL loss of \cref{rv:eq:gnll_loss}.  In the
TGT context this allows the model to increase aleatoric variance in
regimes with elevated sensor noise or unmodelled dynamics:

\begin{enumerate}[leftmargin=2.2em,label=\arabic*.]
  \item \textbf{Two-head network.}  
        A shared feature extractor feeds a mean head that predicts the
        baseline TGT and a variance head that outputs
        $s(\mathbf{x})=\log\sigma^{2}(\mathbf{x})$.
  \item \textbf{Training via GNLL.}  
        Minimize $\mathcal{L}_{\text{GNLL}}$
        from \cref{rv:eq:gnll_loss}.  
        Dropout ($p=0.1$) on hidden layers helps variance learning by
        exposing the model to diverse residuals.
  \item \textbf{Prediction interval.}  
        The aleatoric variance
        $\hat{\sigma}^{2}_{\text{al}}(\mathbf{x}_{*})$
        often rises steeply during throttle transients, matching
        increased sensor jitter.  Combine this term with an epistemic
        component estimated from a small ($B=5$) deep ensemble to
        obtain the total variance $\hat{\sigma}^{2}_{*}$ and evaluate
        $\mathcal{I}_{1-\alpha}(\mathbf{x}_{*})$ via \cref{rv:eq:gauss_PI}.
  \item \textbf{Interpretation.}  
        Wide intervals during hot climbs flag potential thermal margin
        exceedances, prompting maintenance scheduling.
\end{enumerate}

%---------------------------------------------------------------------
\paragraph{Summary.}
All five techniques fit seamlessly into the TGT-prediction workflow;
they differ in computational cost, calibration effort, and their
ability to separate aleatoric from epistemic uncertainty.  The comparative trade-offs between coverage, interval width, and point-forecast accuracy on the proprietary flight-test set are summarized in \cref{rv:tab:method_comparison} and broken down by flight phase in \cref{rv:tab:pi_comparison,rv:tab:pi_metrics,rv:tab:pi_methods_example}.

%=====================================================================

%=====================================================================
\section{Experiment Setup}
\label{rv:sec:experiment_setup}
%=====================================================================

\begin{figure*}
\centering
\begin{tikzpicture}[
    node distance=10mm and 12mm,
    >=Stealth,
    startstop/.style={ellipse, draw, fill=gray!10, minimum width=22mm, minimum height=7mm, align=center},
    process/.style={rectangle, draw, fill=gray!10, minimum width=0.92\textwidth, minimum height=8mm, align=center},
    method/.style={rectangle, draw, fill=gray!10, minimum width=22mm, minimum height=8mm, align=center}
]

\node (start) [startstop] {Start};

\node (split) [process, below=of start] {Split data into training sets ($\mathcal{D}_{1}$ and $\mathcal{D}_{2}$) and test set ($\mathcal{D}_{\text{test}}$)};

\node (cv) [process, below=of split] {Perform five-fold cross validation to determine the optimal NN structure};

\node (construct) [process, below=of cv] {Construct prediction intervals (PIs) using};

% Method nodes (alphabetical order)
\node (bayes)     [method, below=14mm of construct, xshift=-48mm] {Bayesian};
\node (bootstrap) [method, below=14mm of construct, xshift=-24mm] {Bootstrap};
\node (delta)     [method, below=14mm of construct, xshift=0mm]   {Delta};
\node (lube)      [method, below=14mm of construct, xshift=24mm]  {LUBE};
\node (mve)       [method, below=14mm of construct, xshift=48mm]  {MVE};

% Put metrics below the method row
\node (metrics) [process, below=14mm of delta] {Calculate and record PICP, NMPIW, and CWC};

\node (compare) [process, below=of metrics] {Compare the quality of PIs using obtained results};

\node (end) [startstop, below=of compare] {End};

% Main vertical flow
\draw[->] (start) -- (split);
\draw[->] (split) -- (cv);
\draw[->] (cv) -- (construct);

%-----------------------------------------
% Bracket-style split: construct -> methods
%-----------------------------------------
% (If your preamble does not already have it, add: \usetikzlibrary{calc})
\coordinate (topbus) at ($(delta.north)+(0,3mm)$);

% vertical stem down from construct into the bus
\draw (construct.south) -- (topbus);

% horizontal bus spanning all methods
\draw (topbus -| bayes.north) -- (topbus -| mve.north);

% vertical drops with arrowheads into each method
\foreach \m in {bayes,bootstrap,delta,lube,mve}{
  \draw[->] (topbus -| \m.north) -- (\m.north);
}

%-----------------------------------------
% Bracket-style merge: methods -> metrics
%-----------------------------------------
\coordinate (botbus) at ($(delta.south)+(0,-3mm)$);

% horizontal bus spanning all methods
\draw (botbus -| bayes.south) -- (botbus -| mve.south);

% vertical drops from each method into the bus
\foreach \m in {bayes,bootstrap,delta,lube,mve}{
  \draw (\m.south) -- (botbus -| \m.south);
}

% single arrow from the bus down into metrics
\draw[->] (botbus) -- (metrics.north);

% Finish
\draw[->] (metrics) -- (compare);
\draw[->] (compare) -- (end);

\end{tikzpicture}
\caption{Experimental workflow for training, tuning, and evaluating TGT prediction-interval methods.}
\label{rv:fig:exp_flowchart}
\end{figure*}
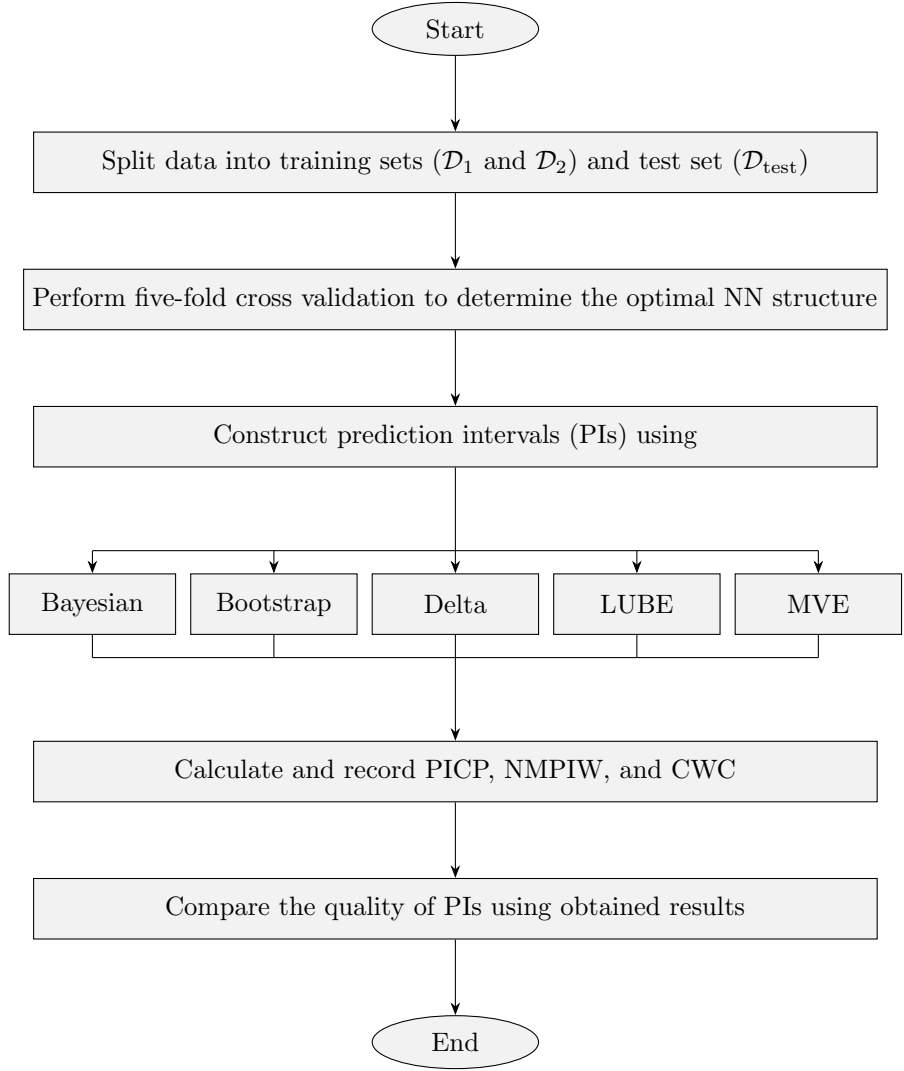

\paragraph{Workflow overview.}
The end-to-end experimental workflow is summarized in \cref{rv:fig:exp_flowchart}. Starting from engine health monitoring records, the dataset is partitioned into (i) a training/validation set and (ii) a held-out test set. The training/validation portion is further divided into two disjoint subsets, $\mathcal{D}_{1}$ and $\mathcal{D}_{2}$, to support both model fitting and method-specific calibration (e.g., residual-variance estimation for Gaussian-based intervals and hyperparameter selection). For each uncertainty quantification approach---Delta (\cref{rv:ssec:tgt_delta}), Bayesian MC-dropout (\cref{rv:ssec:tgt_mcdo}), fixed-data bootstrap ensembles (\cref{rv:ssec:tgt_bootstrap}), Lower--Upper Bound Estimation (LUBE; \cref{rv:ssec:tgt_lube}), and Mean--Variance Estimation (MVE; \cref{rv:ssec:tgt_mve})---the methodology trains the corresponding model(s), generates prediction intervals on the held-out test engines, and evaluates both point accuracy and interval quality using the metrics in \cref{rv:sec:pi_metrics}.

%-----------------------------
% Experimental workflow figure
%-----------------------------

\paragraph{Detailed procedure.}
The individual steps of \cref{rv:fig:exp_flowchart} are implemented as follows.

\begin{enumerate}[leftmargin=2.4em,label=\textbf{Step \arabic*.}]
  \item \textbf{Split data into $\mathcal{D}_{1}$, $\mathcal{D}_{2}$, and $\mathcal{D}_{\text{test}}$.}
  The dataset contains one hundred twenty-two engines used for training/validation and an additional six engines held out for testing. All splits are performed \emph{by engine} (rather than by time step) to avoid leakage between training and evaluation.
  Each engine contains multiple ``segmented'' flight intervals (typically one to six segments) corresponding to scheduled maintenance records, and each segment contains three operational phases: takeoff, climb, and cruise.
  No feature scaling was applied so that all reported errors and interval widths remain in physically interpretable units of $^\circ$C.

  \item \textbf{Five-fold cross validation for NN structure selection.}
  Within the training/validation portion five-fold cross validation is performed to select hyperparameters and NN structure for each UQ method. Candidate configurations use a three-hidden-layer multilayer perceptron with nodes per hidden layer in $\{32,64,128\}$, activation functions in $\{\mathrm{ReLU},\mathrm{GELU}\}$, dropout rates in $\{0,0.02,0.04\}$, training epochs in $\{20,40,60\}$, and learning rates in $\{10^{-2},10^{-3},10^{-4}\}$. The best-performing settings are reported in \cref{rv:tab:best_hyperparameters}.

  \item \textbf{Construct prediction intervals with five UQ methods.}
  After selecting the NN configuration, each UQ methodology is trained and deployed using the TGT specializations described in \cref{rv:ssec:tgt_delta,rv:ssec:tgt_mcdo,rv:ssec:tgt_bootstrap,rv:ssec:tgt_lube,rv:ssec:tgt_mve}. All methods target a fixed nominal confidence level and output lower and upper bounds at each time step of a flight segment.

  \item \textbf{Compute PI and accuracy metrics on $\mathcal{D}_{\text{test}}$.}
  Interval quality is assessed using PICP (\cref{rv:eq:picp}), MPIW (\cref{rv:eq:mpiw}), NMPIW (\cref{rv:eq:nmpiw}), and CWC (\cref{rv:eqn:cwc}), as defined in \cref{rv:sec:pi_metrics}. In addition to PI metrics, point-forecast accuracy is reported using mean absolute error (MAE), and variability is characterized using the standard deviation of absolute error (S.D.\ AE). For methods with variable-width intervals, the standard deviation of prediction-interval width (S.D.\ PIW) is also reported.

  \item \textbf{Compare methods overall and by operating regime.}
  To understand when a given UQ approach is most effective, metrics are aggregated (i) over the full test set, (ii) by test engine, and (iii) by flight phase (takeoff/climb/cruise). These slices reveal trade-offs between sharpness and calibration that may be masked by a single global average.
\end{enumerate}

\section{Results}
\label{rv:sec:results}

Overall, the results reveal a clear trade-off between \emph{calibration} (high coverage) and \emph{sharpness} (narrow intervals). As summarized in \Cref{rv:tab:method_comparison} and visualized in \Cref{rv:fig:overall_picp,rv:fig:overall_mpiw,rv:fig:overall_cwc}, the Bayesian and LUBE methods achieve high empirical coverage but do so with comparatively wide intervals. Bootstrap, in contrast, produces extremely sharp intervals but exhibits substantial under-coverage, which leads to very large CWC values. Among the evaluated approaches, MVE provides the most balanced behavior, combining low MAE with near-nominal coverage and moderate interval width. In the tables/figures below, the method labeled ``Delta'' refers to the
\emph{residual-only} Gaussian baseline described in \cref{rv:ssec:tgt_delta} (i.e., $\boldsymbol{\Sigma}_{\mathbf{x},*}=\mathbf{0}$ so only $\hat{\sigma}^2_{\text{res}}$ contributes to interval width).

A second notable trend is the strong dependence of uncertainty behavior on operating regime. The phase-resolved tables (\Cref{rv:tab:pi_comparison,rv:tab:pi_metrics,rv:tab:pi_methods_example}) and the phase-wise visualizations (\Cref{rv:fig:phase_picp,rv:fig:phase_nmpiw,rv:fig:phase_mpiw}) show that takeoff is the most challenging phase for calibrated coverage (especially for Bayesian and bootstrap). Interval widths, however, increase most dramatically for Bayesian in climb (MPIW 63.53$^\circ$C versus 19.29$^\circ$C in takeoff), while Delta and LUBE remain relatively stable across phases and MVE is widest in takeoff. Cruise is the most benign regime for point accuracy, yet bootstrap still under-covers, underscoring that sharpness alone is not sufficient for risk-aware EHM.

This regime dependence suggests that phase-conditioned UQ or post-hoc calibration may be required for consistent performance across the full flight envelope.

\subsection{Hyperparameter Selection}

\begin{table}
\centering
\caption{Best Hyperparameters for each method.}
\label{rv:tab:best_hyperparameters}
\resizebox{\linewidth}{!}{%
\begin{tabular}{lccccc}
\toprule
Method & Activation Function & Dropout (\%) & Training Epochs & Nodes per Hidden Layer & Learning Rate \\
\midrule
Bayesian & ReLU & 4 & 20 & 128 & 1e-3 \\
Bootstrap & ReLU & 0 & 20 & 128 & 1e-4 \\
Delta    & ReLU & 0 & 20 & 128 & 1e-4 \\
LUBE     & GELU & 0 & 40 & 128 & 1e-4 \\
MVE      & GELU & 0 & 40 & 64  & 1e-4  \\
\bottomrule
\end{tabular}
}
\end{table}

\Cref{rv:tab:best_hyperparameters} lists the hyperparameters selected by five-fold cross validation for each UQ method. The chosen configurations consistently favor relatively small learning rates ($10^{-3}$--$10^{-4}$) and modest regularization (0--4\% dropout), suggesting that aggressive regularization is unnecessary for the available training set size and the three-layer MLP used throughout. ReLU is selected for the Delta and Bayesian (MC-dropout) implementations, while GELU is preferred for LUBE and MVE, which explicitly learn interval bounds or predictive variances and may benefit from smoother activation functions. Across methods, widths of 64--128 hidden units are consistently selected, indicating that representational capacity beyond this range offers limited marginal benefit under the evaluated metric suite.

\subsection{Overall Method Comparison}

\begin{table}
\centering
\resizebox{\linewidth}{!}{%
\begin{tabular}{lccccccc}
\toprule
Method & PICP (\%) & MPIW ($^\circ$C) & S.D.\ of PIW ($^\circ$C) & NMPIW (\%) & CWC & MAE of TGT ($^\circ$C) & S.D.\ AE of TGT ($^\circ$C) \\
\midrule
Bayesian & 96 & 70.60 & 29.21 & 14 & 70.60 & 9.09 & 7.50 \\
Bootstrap & 35 & 6.80 & 1.48 & 3 & 7.79e13 & 6.07 & 4.27 \\
Delta    & 94 & \textbf{32.22} & N/A   & \textbf{6}  & 86.85 & 6.73 & \textbf{5.28} \\
LUBE     & \textbf{98} & 39.67 & 7.75  & 8  & 39.67 & 7.57 & 5.49 \\
MVE      & 96 & 37.14 & \textbf{7.49}  & 7  & \textbf{37.14} & \textbf{6.52} & 5.50 \\
\bottomrule
\end{tabular}
}
\caption{Overall performance comparison of different methods.}
\label{rv:tab:method_comparison}
\end{table}

\Cref{rv:tab:method_comparison} summarizes aggregate performance over the held-out test engines. Bootstrap achieves the lowest MAE (6.07$^\circ$C) and lowest S.D.\ AE (4.27$^\circ$C), but its empirical coverage is only 35\%, indicating that the resulting intervals are far too narrow and systematically miss the ground truth. This severe under-coverage is reflected by an extremely large CWC (7.79e13), which strongly penalizes methods that fail to meet the target coverage. Bayesian MC-dropout attains near-nominal coverage (96\%) but with the widest intervals (MPIW 70.60$^\circ$C) and the largest point error (MAE 9.09$^\circ$C), suggesting a conservative uncertainty estimate. Delta, LUBE, and MVE provide more balanced trade-offs: Delta and MVE yield low MAE (6.73 and 6.52$^\circ$C) with near-nominal coverage (94--96\%) and moderate widths, while LUBE achieves the highest coverage (98\%) at the expense of slightly larger widths. Among the non-bootstrap methods, MVE attains the lowest CWC, indicating the most favorable combined width--coverage behavior under the chosen metric.

\subsection{Phase-Resolved Results}

\begin{table}
\centering
\resizebox{\linewidth}{!}{%
\begin{tabular}{lccccccc}
\toprule
Method & PICP (\%) & MPIW ($^{\circ}$C) & S.D.\ of PIW ($^{\circ}$C) & NMPIW (\%) & CWC & MAE of TGT ($^{\circ}$C) & S.D.\ AE of TGT ($^{\circ}$C) \\
\midrule
Bayesian         & 53  & 19.29 & 6.06 & 5  & 2.06e10  & 7.22 & 4.86 \\
Bootstrap        & 26  & 4.61  & 0.48 & 1  & 5.10e15 & 4.96 & 3.44 \\
Delta            & 100 & 37.40 & N/A  & 11 & 37.40  & 4.14 & 3.50 \\
LUBE& 100 & 41.49 & 2.81 & 12 & 41.49  & 5.09 & 3.59 \\
MVE    & 100 & 64.77 & 2.98 & 18 & 64.77  & 6.35 & 3.89 \\
\bottomrule
\end{tabular}
}
\caption{Performance comparison for takeoff phase.}
\label{rv:tab:pi_comparison}
\end{table}

\Cref{rv:tab:pi_comparison} reports phase-specific results for takeoff. Takeoff is the most challenging regime for calibrated coverage for the Bayesian and bootstrap methods, with PICP values of 53\% and 26\%, respectively. Both methods also produce relatively sharp intervals in this phase (MPIW 19.29$^\circ$C and 4.61$^\circ$C), which explains their extremely large CWC values due to the penalty for under-coverage (\cref{rv:eqn:cwc}). In contrast, Delta, LUBE, and MVE maintain near-perfect coverage (99.78--100\%) during takeoff. Among these, Delta achieves the lowest MAE (4.14$^\circ$C) and the most favorable CWC (37.40), while MVE attains full coverage at the cost of substantially wider intervals (MPIW 64.77$^\circ$C).

\begin{table}
\centering
\resizebox{\linewidth}{!}{%
\begin{tabular}{lccccccc}
\toprule
Method & PICP (\%) & MPIW ($^\circ$C) & S.D.\ of PIW ($^\circ$C) & NMPIW (\%) & CWC & MAE of TGT ($^\circ$C) & S.D.\ AE of TGT ($^\circ$C) \\
\midrule
Bayesian         & 100 & 63.53 & 11.51 & 27 & 63.53  & 8.16 & 5.76 \\
Bootstrap        & 35  & 6.80  & 1.35  & 3  & 7.78e13 & 6.07 & 4.27 \\
Delta            & 99  & 37.94 & N/A   & 16 & 37.94  & 5.40 & 4.16 \\
LUBE& 99  & 43.00 & 1.27  & 18 & 43.00  & 7.70 & 4.41 \\
MVE    & 99  & 44.47 & 0.95  & 19 & 44.47  & 8.85 & 4.82 \\
\bottomrule
\end{tabular}
}
\caption{Performance comparison for climb phase.}
\label{rv:tab:pi_metrics}
\end{table}

\Cref{rv:tab:pi_metrics} shows climb-phase behavior. All non-bootstrap methods achieve near-perfect coverage in climb (98.58--99.78\%), but their sharpness differs markedly. Bayesian is the most conservative (MPIW 63.53$^\circ$C) while Delta attains comparable coverage with substantially narrower intervals (MPIW 37.94$^\circ$C) and the lowest MAE (5.40$^\circ$C), yielding the most favorable climb CWC. LUBE and MVE also maintain high coverage (99.32\% and 98.58\%) with moderate widths, but exhibit higher point errors in this regime. Bootstrap again produces very narrow intervals (MPIW 6.80$^\circ$C) yet under-covers severely (34.86\%), leading to an enormous CWC.

\begin{table}
\centering
\resizebox{\linewidth}{!}{%
\begin{tabular}{lccccccc}
\toprule
Method & PICP (\%) & MPIW ($^\circ$C) & S.D.\ of PIW ($^\circ$C) & NMPIW (\%) & CWC & MAE of TGT ($^\circ$C) & S.D.\ AE of TGT ($^\circ$C) \\
\midrule
Bayesian          & 100 & 46.83 & 9.40 & 18 & 46.83  & 5.23 & 3.98 \\
Bootstrap         &  56 &  8.48 & 1.19 &  3 & 2.50e9  & 3.99 & 2.69 \\
Delta             & 100 & 37.91 & N/A  & 15 & 37.91  & 4.09 & 2.71 \\
LUBE & 100 & 43.26 & 1.57 & 17 & 43.26  & 8.34 & 4.23 \\
MVE     & 100 & 33.45 & 0.44 & 13 & 33.45  & 3.93 & 2.73 \\
\bottomrule
\end{tabular}
}
\caption{Performance comparison for cruise phase.}
\label{rv:tab:pi_methods_example}
\end{table}

\Cref{rv:tab:pi_methods_example} summarizes cruise-phase results. Cruise yields the lowest MAE values for most methods (3.93--5.23$^\circ$C), consistent with a more stable operating regime. Bayesian, Delta, LUBE, and MVE all achieve near-perfect coverage (99.87--99.99\%). MVE provides the sharpest calibrated intervals (MPIW 33.45$^\circ$C) while also achieving the lowest MAE (3.93$^\circ$C), with Delta close behind (MAE 4.09$^\circ$C, MPIW 37.91$^\circ$C). Bootstrap attains competitive MAE (3.99$^\circ$C) but under-covers (55.99\%), resulting in a large CWC.

\subsection{Aggregate Metric Visualizations}

%======================================
% Mean Absolute TGT Error Barchart
%======================================
\begin{figure}[H]
\centering
\resizebox{0.45\textwidth}{!}{
\begin{tikzpicture}
\begin{axis}[
    at={(6,0)},
    anchor=north west,
    width=10cm,
    height=10cm,
    ymin=0,
    ymax=12,
    enlarge x limits=0.4,
    ylabel={Mean Absolute Error ($^\circ$C)},
    symbolic x coords={Bayesian,Bootstrap,Delta,LUBE,MVE},
    xtick={Bayesian,Bootstrap,Delta,LUBE,MVE},
    xticklabel style={align=center, rotate=-45},
    nodes near coords,
    nodes near coords align={vertical},
    colormap/viridis,          % <-- use viridis colormap
    colormap name = viridis
]

\addplot+[ybar, bar width=14pt,
          /utils/exec={\pgfplotscolormapdefinemappedcolor{0}},
          color=mapped color,
          fill=mapped color,
          draw=black,
          every node near coord/.append style={text=black}, 
          mark=-] coordinates {(Bayesian,9.09)};

\addplot+[ybar, bar width=14pt,
          /utils/exec={\pgfplotscolormapdefinemappedcolor{250}},
          color=mapped color,
          fill=mapped color,
          draw=black,
          every node near coord/.append style={text=black}, 
          mark=-] coordinates {(Bootstrap,6.07)};

\addplot+[ybar, bar width=14pt,
          /utils/exec={\pgfplotscolormapdefinemappedcolor{500}},
          color=mapped color,
          fill=mapped color,
          draw=black,
          every node near coord/.append style={text=black}, 
          mark=-] coordinates {(Delta,6.73)};

\addplot+[ybar, bar width=14pt,
          /utils/exec={\pgfplotscolormapdefinemappedcolor{750}},
          color=mapped color,
          fill=mapped color,
          draw=black,
          every node near coord/.append style={text=black}, 
          mark=-] coordinates {(LUBE,7.57)};
          
% sample 5 distinct colors from viridis at positions 0..1000
\addplot+[ybar, bar width=14pt,
          /utils/exec={\pgfplotscolormapdefinemappedcolor{1000}},
          color=mapped color,
          fill=mapped color,
          draw=black,
          every node near coord/.append style={text=black}, 
          %every node near coord/.append style={text=mapped color}, 
          mark=-] coordinates {(MVE,6.52)};

\end{axis}
\end{tikzpicture}
}
\caption{Overall comparison of the mean absolute TGT prediction error for the uncertainty quantification methodologies.}
\label{rv:fig:overall_mae}
\end{figure}
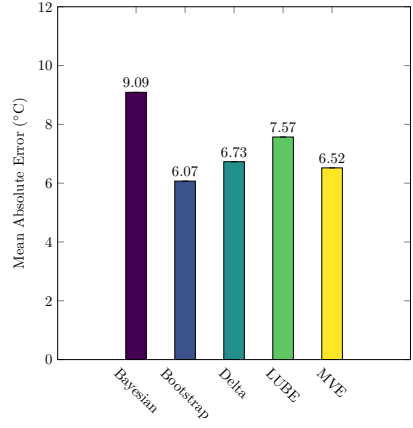

\begin{figure}[H]
\centering
\resizebox{0.45\textwidth}{!}{
\begin{tikzpicture}
\begin{axis}[
    at={(6,0)},
    anchor=north west,
    width=10cm,
    height=10cm,
    ymin=30,
    ymax=105,
    enlarge x limits=0.4,
    ylabel={Prediction Interval Coverage\\Probability (PICP) (\%)},
    ylabel style={align=center},
    symbolic x coords={Bayesian,Bootstrap,Delta,LUBE,MVE},
    xtick={Bayesian,Bootstrap,Delta,LUBE,MVE},
    xticklabel style={align=center, rotate=-45},
    nodes near coords,
    nodes near coords align={vertical},
    colormap/viridis,
    colormap name=viridis
]

% Bayesian
\addplot+[ybar, bar width=14pt,
          /utils/exec={\pgfplotscolormapdefinemappedcolor{0}},
          color=mapped color,
          fill=mapped color,
          draw=black,
          every node near coord/.append style={text=black}, 
          mark=-] coordinates {(Bayesian,96)};

% Bootstrap
\addplot+[ybar, bar width=14pt,
          /utils/exec={\pgfplotscolormapdefinemappedcolor{250}},
          color=mapped color,
          fill=mapped color,
          draw=black,
          every node near coord/.append style={text=black}, 
          mark=-] coordinates {(Bootstrap,35)};

% Delta
\addplot+[ybar, bar width=14pt,
          /utils/exec={\pgfplotscolormapdefinemappedcolor{500}},
          color=mapped color,
          fill=mapped color,
          draw=black,
          every node near coord/.append style={text=black}, 
          mark=-] coordinates {(Delta,94)};

% LUBE
\addplot+[ybar, bar width=14pt,
          /utils/exec={\pgfplotscolormapdefinemappedcolor{750}},
          color=mapped color,
          fill=mapped color,
          draw=black,
          every node near coord/.append style={text=black}, 
          mark=-] coordinates {(LUBE,98)};

% MVE
\addplot+[ybar, bar width=14pt,
          /utils/exec={\pgfplotscolormapdefinemappedcolor{1000}},
          color=mapped color,
          fill=mapped color,
          draw=black,
          every node near coord/.append style={text=black}, 
          mark=-] coordinates {(MVE,96)};

\end{axis}
\end{tikzpicture}
}
\caption{Overall comparison of the prediction interval coverage probability (PICP) for the uncertainty quantification methodologies.}
\label{rv:fig:overall_picp}
\end{figure}
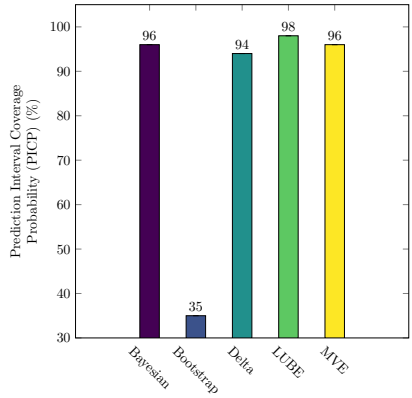

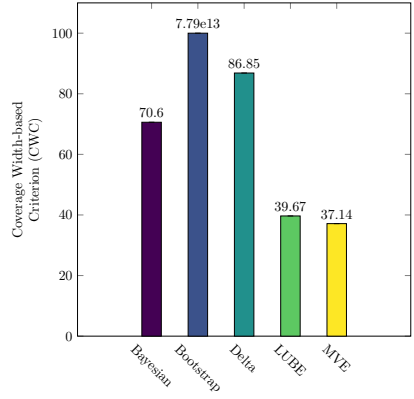
\begin{figure}[H]
\centering
\resizebox{0.45\textwidth}{!}{
\begin{tikzpicture}
\begin{axis}[
    at={(6,0)},
    anchor=north west,
    width=10cm,
    height=10cm,
    ymin=0,
    ymax=110,
    enlarge x limits=0.4,
    ylabel={Coverage Width-based\\Criterion (CWC)},
    ylabel style={align=center},
    symbolic x coords={Bayesian,Bootstrap,Delta,LUBE,MVE},
    xtick={Bayesian,Bootstrap,Delta,LUBE,MVE},
    xticklabel style={align=center, rotate=-45},
    nodes near coords,
    nodes near coords align={vertical},
    colormap/viridis,
    colormap name=viridis
]

% Bayesian
\addplot+[ybar, bar width=14pt,
          /utils/exec={\pgfplotscolormapdefinemappedcolor{0}},
          color=mapped color,
          fill=mapped color,
          draw=black,
          every node near coord/.append style={text=black}, 
          mark=-] coordinates {(Bayesian,70.60)};

% Bootstrap
\addplot+[ybar, bar width=14pt,
          /utils/exec={\pgfplotscolormapdefinemappedcolor{250}},
          color=mapped color,
          fill=mapped color,
          % show true CWC value instead of 100
          nodes near coords={7.79e13},
          draw=black,
          every node near coord/.append style={text=black}, 
          mark=-] coordinates {(Bootstrap,100)};

% Delta
\addplot+[ybar, bar width=14pt,
          /utils/exec={\pgfplotscolormapdefinemappedcolor{500}},
          color=mapped color,
          fill=mapped color,
          draw=black,
          every node near coord/.append style={text=black}, 
          mark=-] coordinates {(Delta,86.85)};

% LUBE
\addplot+[ybar, bar width=14pt,
          /utils/exec={\pgfplotscolormapdefinemappedcolor{750}},
          color=mapped color,
          fill=mapped color,
          draw=black,
          every node near coord/.append style={text=black}, 
          mark=-] coordinates {(LUBE,39.67)};

% MVE
\addplot+[ybar, bar width=14pt,
          /utils/exec={\pgfplotscolormapdefinemappedcolor{1000}},
          color=mapped color,
          fill=mapped color,
          draw=black,
          every node near coord/.append style={text=black}, 
          mark=-] coordinates {(MVE,37.14)};

\end{axis}
\end{tikzpicture}
}
\caption{Overall comparison of the coverage width-based criterion (CWC) for the uncertainty quantification methodologies.}
\label{rv:fig:overall_cwc}
\end{figure}

\Cref{rv:fig:overall_mae} shows the aggregate mean absolute error (MAE) across all test points. Bootstrap yields the smallest MAE, with MVE and Delta close behind, whereas Bayesian exhibits the largest MAE. This highlights that strong uncertainty calibration does not automatically imply strong point accuracy, and also foreshadows that the sharpest point predictor (bootstrap) may still fail to provide well-calibrated intervals when coverage is considered (see \Cref{rv:fig:overall_picp,rv:fig:overall_cwc}).

%======================================
% S.D. Absolute TGT Error Barchart
%======================================
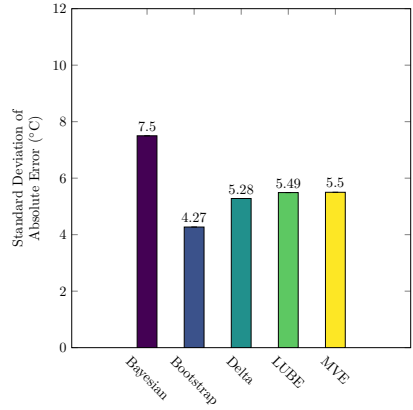
\begin{figure}[H]
\centering
\resizebox{0.45\textwidth}{!}{
\begin{tikzpicture}
\begin{axis}[
    at={(6,0)},
    anchor=north west,
    width=10cm,
    height=10cm,
    ymin=0,
    ymax=12,
    enlarge x limits=0.4,
    ylabel={Standard Deviation of\\Absolute Error ($^\circ$C)},
    ylabel style={align=center},
    symbolic x coords={Bayesian,Bootstrap,Delta,LUBE,MVE},
    xtick={Bayesian,Bootstrap,Delta,LUBE,MVE},
    xticklabel style={align=center, rotate=-45},
    nodes near coords,
    nodes near coords align={vertical},
    colormap/viridis,
    colormap name=viridis
]

% Bayesian
\addplot+[ybar, bar width=14pt,
          /utils/exec={\pgfplotscolormapdefinemappedcolor{0}},
          color=mapped color,
          fill=mapped color,
          draw=black,
          every node near coord/.append style={text=black}, 
          mark=-] coordinates {(Bayesian,7.50)};

% Bootstrap
\addplot+[ybar, bar width=14pt,
          /utils/exec={\pgfplotscolormapdefinemappedcolor{250}},
          color=mapped color,
          fill=mapped color,
          draw=black,
          every node near coord/.append style={text=black}, 
          mark=-] coordinates {(Bootstrap,4.27)};

% Delta
\addplot+[ybar, bar width=14pt,
          /utils/exec={\pgfplotscolormapdefinemappedcolor{500}},
          color=mapped color,
          fill=mapped color,
          draw=black,
          every node near coord/.append style={text=black}, 
          mark=-] coordinates {(Delta,5.28)};

% LUBE
\addplot+[ybar, bar width=14pt,
          /utils/exec={\pgfplotscolormapdefinemappedcolor{750}},
          color=mapped color,
          fill=mapped color,
          draw=black,
          every node near coord/.append style={text=black}, 
          mark=-] coordinates {(LUBE,5.49)};

% MVE
\addplot+[ybar, bar width=14pt,
          /utils/exec={\pgfplotscolormapdefinemappedcolor{1000}},
          color=mapped color,
          fill=mapped color,
          draw=black,
          every node near coord/.append style={text=black}, 
          mark=-] coordinates {(MVE,5.50)};

\end{axis}
\end{tikzpicture}
}
\caption{Overall comparison of the standard deviation of absolute TGT prediction error for the uncertainty quantification methodologies.}
\label{rv:fig:overall_sd_ae}
\end{figure}

\Cref{rv:fig:overall_sd_ae} reports the standard deviation of absolute error (S.D.\ AE) across the test set. Bayesian shows the highest variability in absolute error, consistent with its larger MAE, while bootstrap exhibits the lowest variability. Delta, LUBE, and MVE fall in a narrower band, suggesting comparatively stable point-prediction behavior across a range of engines and operating regimes.

%======================================
% MPIW Barchart
%======================================
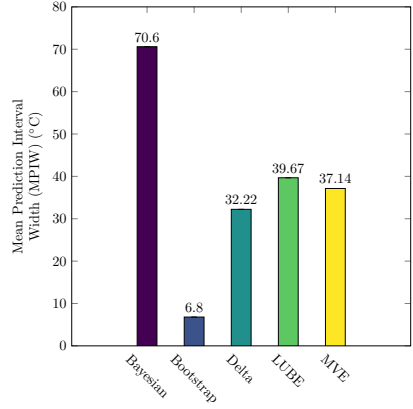
\begin{figure}[H]
\centering
\resizebox{0.45\textwidth}{!}{
\begin{tikzpicture}
\begin{axis}[
    at={(6,0)},
    anchor=north west,
    width=10cm,
    height=10cm,
    ymin=0,
    ymax=80,
    enlarge x limits=0.4,
    ylabel={Mean Prediction Interval\\Width (MPIW) ($^\circ$C)},
    ylabel style={align=center},
    symbolic x coords={Bayesian,Bootstrap,Delta,LUBE,MVE},
    xtick={Bayesian,Bootstrap,Delta,LUBE,MVE},
    xticklabel style={align=center, rotate=-45},
    nodes near coords,
    nodes near coords align={vertical},
    colormap/viridis,
    colormap name=viridis
]

% Bayesian
\addplot+[ybar, bar width=14pt,
          /utils/exec={\pgfplotscolormapdefinemappedcolor{0}},
          color=mapped color,
          fill=mapped color,
          draw=black,
          every node near coord/.append style={text=black}, 
          mark=-] coordinates {(Bayesian,70.60)};

% Bootstrap
\addplot+[ybar, bar width=14pt,
          /utils/exec={\pgfplotscolormapdefinemappedcolor{250}},
          color=mapped color,
          fill=mapped color,
          draw=black,
          every node near coord/.append style={text=black}, 
          mark=-] coordinates {(Bootstrap,6.80)};

% Delta
\addplot+[ybar, bar width=14pt,
          /utils/exec={\pgfplotscolormapdefinemappedcolor{500}},
          color=mapped color,
          fill=mapped color,
          draw=black,
          every node near coord/.append style={text=black}, 
          mark=-] coordinates {(Delta,32.22)};

% LUBE
\addplot+[ybar, bar width=14pt,
          /utils/exec={\pgfplotscolormapdefinemappedcolor{750}},
          color=mapped color,
          fill=mapped color,
          draw=black,
          every node near coord/.append style={text=black}, 
          mark=-] coordinates {(LUBE,39.67)};

% MVE
\addplot+[ybar, bar width=14pt,
          /utils/exec={\pgfplotscolormapdefinemappedcolor{1000}},
          color=mapped color,
          fill=mapped color,
          draw=black,
          every node near coord/.append style={text=black}, 
          mark=-] coordinates {(MVE,37.14)};

\end{axis}
\end{tikzpicture}
}
\caption{Overall comparison of the mean prediction interval width (MPIW) for the uncertainty quantification methodologies.}
\label{rv:fig:overall_mpiw}
\end{figure}

\Cref{rv:fig:overall_mpiw} compares the mean prediction interval width (MPIW). Bayesian produces the widest intervals by a large margin, indicating conservative uncertainty estimates. Bootstrap produces the narrowest intervals, while Delta, LUBE, and MVE occupy an intermediate regime. Notably, LUBE and MVE widen intervals relative to Delta in order to achieve higher empirical coverage (see \Cref{rv:fig:overall_picp}).

%======================================
% NMPIW Barchart
%======================================
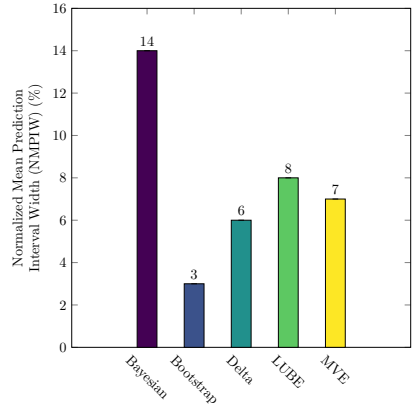
\begin{figure}[H]
\centering
\resizebox{0.45\textwidth}{!}{
\begin{tikzpicture}
\begin{axis}[
    at={(6,0)},
    anchor=north west,
    width=10cm,
    height=10cm,
    ymin=0,
    ymax=16,
    enlarge x limits=0.4,
    ylabel={Normalized Mean Prediction\\Interval Width (NMPIW) (\%)},
    ylabel style={align=center},
    symbolic x coords={Bayesian,Bootstrap,Delta,LUBE,MVE},
    xtick={Bayesian,Bootstrap,Delta,LUBE,MVE},
    xticklabel style={align=center, rotate=-45},
    nodes near coords,
    nodes near coords align={vertical},
    colormap/viridis,
    colormap name=viridis
]

% Bayesian
\addplot+[ybar, bar width=14pt,
          /utils/exec={\pgfplotscolormapdefinemappedcolor{0}},
          color=mapped color,
          fill=mapped color,
          draw=black,
          every node near coord/.append style={text=black}, 
          mark=-] coordinates {(Bayesian,14)};

% Bootstrap
\addplot+[ybar, bar width=14pt,
          /utils/exec={\pgfplotscolormapdefinemappedcolor{250}},
          color=mapped color,
          fill=mapped color,
          draw=black,
          every node near coord/.append style={text=black}, 
          mark=-] coordinates {(Bootstrap,3)};

% Delta
\addplot+[ybar, bar width=14pt,
          /utils/exec={\pgfplotscolormapdefinemappedcolor{500}},
          color=mapped color,
          fill=mapped color,
          draw=black,
          every node near coord/.append style={text=black}, 
          mark=-] coordinates {(Delta,6)};

% LUBE
\addplot+[ybar, bar width=14pt,
          /utils/exec={\pgfplotscolormapdefinemappedcolor{750}},
          color=mapped color,
          fill=mapped color,
          draw=black,
          every node near coord/.append style={text=black}, 
          mark=-] coordinates {(LUBE,8)};

% MVE
\addplot+[ybar, bar width=14pt,
          /utils/exec={\pgfplotscolormapdefinemappedcolor{1000}},
          color=mapped color,
          fill=mapped color,
          draw=black,
          every node near coord/.append style={text=black}, 
          mark=-] coordinates {(MVE,7)};

\end{axis}
\end{tikzpicture}
}
\caption{Overall comparison of the normalized mean prediction interval width (NMPIW) for the uncertainty quantification methodologies.}
\label{rv:fig:overall_nmpiw}
\end{figure}

\Cref{rv:fig:overall_nmpiw} presents the same comparison in normalized form (NMPIW). The ranking mirrors \Cref{rv:fig:overall_mpiw}, confirming that the observed differences reflect true sharpness differences rather than a scaling artifact. Bayesian remains the least sharp, bootstrap the most sharp, and Delta/LUBE/MVE provide intermediate normalized widths.

%======================================
% S.D. of PIW Barchart
%======================================
\begin{figure}[H]
\centering
\resizebox{0.45\textwidth}{!}{
\begin{tikzpicture}
\begin{axis}[
    at={(6,0)},
    anchor=north west,
    width=10cm,
    height=10cm,
    ymin=0,
    ymax=32,
    enlarge x limits=0.4,
    ylabel={Standard Deviation of\\Prediction Interval Width ($^\circ$C)},
    ylabel style={align=center},
    symbolic x coords={Bayesian,Bootstrap,Delta,LUBE,MVE},
    xtick={Bayesian,Bootstrap,Delta,LUBE,MVE},
    xticklabel style={align=center, rotate=-45},
    nodes near coords,
    nodes near coords align={vertical},
    colormap/viridis,
    colormap name=viridis
]

% Bayesian
\addplot+[ybar, bar width=14pt,
          /utils/exec={\pgfplotscolormapdefinemappedcolor{0}},
          color=mapped color,
          fill=mapped color,
          draw=black,
          every node near coord/.append style={text=black}, 
          mark=-] coordinates {(Bayesian,29.21)};

% Bootstrap
\addplot+[ybar, bar width=14pt,
          /utils/exec={\pgfplotscolormapdefinemappedcolor{250}},
          color=mapped color,
          fill=mapped color,
          draw=black,
          every node near coord/.append style={text=black}, 
          mark=-] coordinates {(Bootstrap,1.48)};

% Delta (N/A treated as 0)
\addplot+[ybar, bar width=14pt,
          /utils/exec={\pgfplotscolormapdefinemappedcolor{500}},
          color=mapped color,
          fill=mapped color,
          draw=black,
          every node near coord/.append style={text=black}, 
          mark=-] coordinates {(Delta,0.00)};

% LUBE
\addplot+[ybar, bar width=14pt,
          /utils/exec={\pgfplotscolormapdefinemappedcolor{750}},
          color=mapped color,
          fill=mapped color,
          draw=black,
          every node near coord/.append style={text=black}, 
          mark=-] coordinates {(LUBE,7.75)};

% MVE
\addplot+[ybar, bar width=14pt,
          /utils/exec={\pgfplotscolormapdefinemappedcolor{1000}},
          color=mapped color,
          fill=mapped color,
          draw=black,
          every node near coord/.append style={text=black}, 
          mark=-] coordinates {(MVE,7.49)};

\end{axis}
\end{tikzpicture}
}
\caption{Overall comparison of the standard deviation of prediction interval width for the uncertainty quantification methodologies.}
\label{rv:fig:overall_sd_piw}
\end{figure}
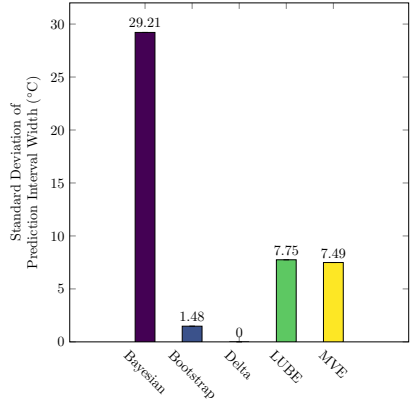

\Cref{rv:fig:overall_sd_piw} quantifies variability in interval width (S.D.\ PIW). Delta has zero variability by construction (constant-width intervals), while Bayesian exhibits the largest width variability, consistent with stochastic forward passes in MC-dropout. LUBE and MVE show moderate variability, reflecting their learned sample-dependent bounds/variances, and bootstrap shows comparatively low variability because ensemble-driven uncertainty is relatively uniform across many samples.

%======================================
% PICP Barchart
%======================================

\Cref{rv:fig:overall_picp} compares prediction-interval coverage (PICP). LUBE achieves the highest coverage (slightly above nominal), while Bayesian and MVE are close to nominal coverage. Delta slightly under-covers, and bootstrap exhibits severe under-coverage. This figure illustrates why coverage-aware criteria such as CWC are needed: sharp intervals that under-cover (bootstrap) are not useful for risk-aware decision-making.

%======================================
% CWC Barchart
%======================================

\Cref{rv:fig:overall_cwc} summarizes the width--coverage trade-off using CWC. Among the calibrated methods, MVE attains the smallest CWC, indicating the most favorable balance between interval width and achieved coverage. Bayesian's CWC is higher due to its wide intervals, and Delta is penalized for under-coverage. Bootstrap is effectively off-scale because under-coverage triggers the exponential penalty term, demonstrating that sharpness without calibration can be strongly disfavored by CWC.

\subsection{Per-Engine Breakdown}

%======================================
% MAE by Method per 6 test engines
%======================================
\begin{figure}[H]
\centering
\resizebox{\textwidth}{!}{
\begin{tikzpicture}
\begin{axis}[
    ybar,
    bar width=9pt,
    ymin=0,
    ymax=11,
    enlarge x limits=0.18,
    width=20cm,
    height=7cm,
    ylabel={Mean Absolute Error ($^\circ$C)},
    xlabel={},
    symbolic x coords={Engine A,Engine B,Engine C,Engine D,Engine E,Engine F},
    xtick=data,
    xticklabel style={align=center},
    legend style={
        at={(0.5,-0.15)},
        anchor=north,
        legend columns=-1
    },
    colormap/viridis,
    colormap name=viridis
]

% Bayesian
\addplot+[
          /utils/exec={\pgfplotscolormapdefinemappedcolor{0}},
          color=mapped color,
          fill=mapped color,
          draw=black,
          ] coordinates {
    (Engine A,9.3)
    (Engine B,8.7)
    (Engine C,7.5)
    (Engine D,9.0)
    (Engine E,9.1)
    (Engine F,10.3)
};

% Bootstrap
\addplot+[
          /utils/exec={\pgfplotscolormapdefinemappedcolor{250}},
          color=mapped color,
          fill=mapped color,
          draw=black,
          ] coordinates {
    (Engine A,9.3)
    (Engine B,8.7)
    (Engine C,7.5)
    (Engine D,9.0)
    (Engine E,9.1)
    (Engine F,10.3)
};

% Delta
\addplot+[
          /utils/exec={\pgfplotscolormapdefinemappedcolor{500}},
          color=mapped color,
          fill=mapped color,
          draw=black,
          ] coordinates {
    (Engine A,5.3)
    (Engine B,7.8)
    (Engine C,4.2)
    (Engine D,9.7)
    (Engine E,6.7)
    (Engine F,7.2)
};

% Lower Upper Bound Estimation
\addplot+[
          /utils/exec={\pgfplotscolormapdefinemappedcolor{750}},
          color=mapped color,
          fill=mapped color,
          draw=black,
          ] coordinates {
    (Engine A,5.6)
    (Engine B,10.0)
    (Engine C,5.2)
    (Engine D,10.3)
    (Engine E,7.5)
    (Engine F,7.5)
};

% Mean Variance Estimation
\addplot+[
          /utils/exec={\pgfplotscolormapdefinemappedcolor{1000}},
          color=mapped color,
          fill=mapped color,
          draw=black,
          ] coordinates {
    (Engine A,5.9)
    (Engine B,6.5)
    (Engine C,4.2)
    (Engine D,7.9)
    (Engine E,6.8)
    (Engine F,7.8)
};

\legend{
  Bayesian,
  Bootstrap,
  Delta,
  LUBE,
  MVE
}

\end{axis}
\end{tikzpicture}
}
\caption{Mean absolute TGT prediction error (MAE) by uncertainty quantification method for six test engines (Engines A--F).}
\label{rv:fig:engine_mae}
\end{figure}
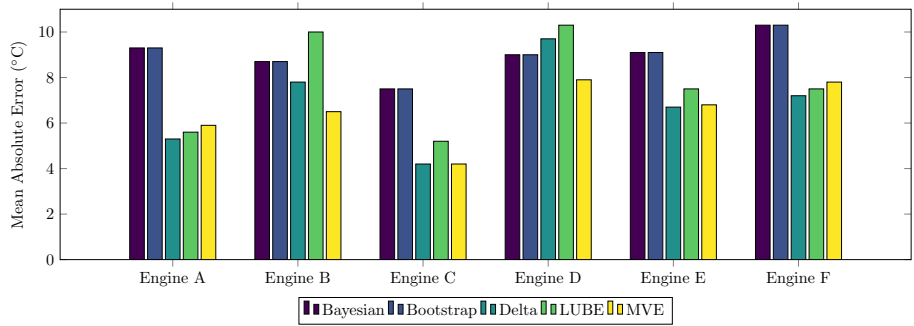

\Cref{rv:fig:engine_mae} breaks MAE down by test engine. All methods experience their largest errors on Engine~D (and, to a lesser extent, Engine~F), suggesting that these engines contain more challenging dynamics or degradation patterns. The remaining engines show tighter clustering of MAE across methods. This spread indicates that method ranking can shift with engine-specific operating history, motivating per-engine diagnostics rather than relying solely on global averages.

%======================================
% Std. Dev. of AE by Method per 6 test engines
%======================================
\begin{figure}[H]
\centering
\resizebox{\textwidth}{!}{
\begin{tikzpicture}
\begin{axis}[
    ybar,
    bar width=9pt,
    ymin=0,
    ymax=9,
    enlarge x limits=0.18,
    width=20cm,
    height=7cm,
    ylabel={Standard Deviation of\\Absolute Error ($^\circ$C)},
    ylabel style={align=center},
    xlabel={},
    symbolic x coords={Engine A,Engine B,Engine C,Engine D,Engine E,Engine F},
    xtick=data,
    xticklabel style={align=center},
    legend style={
        at={(0.5,-0.15)},
        anchor=north,
        legend columns=-1
    },
    colormap/viridis,
    colormap name=viridis
]

% Bayesian
\addplot+[
          /utils/exec={\pgfplotscolormapdefinemappedcolor{0}},
          color=mapped color,
          fill=mapped color,
          draw=black,
          ] coordinates {
    (Engine A,7.8)
    (Engine B,6.6)
    (Engine C,5.5)
    (Engine D,7.7)
    (Engine E,8.0)
    (Engine F,8.3)
};

% Bootstrap
\addplot+[
          /utils/exec={\pgfplotscolormapdefinemappedcolor{250}},
          color=mapped color,
          fill=mapped color,
          draw=black,
          ] coordinates {
    (Engine A,7.8)
    (Engine B,6.6)
    (Engine C,5.5)
    (Engine D,7.7)
    (Engine E,8.0)
    (Engine F,8.3)
};

% Delta
\addplot+[
          /utils/exec={\pgfplotscolormapdefinemappedcolor{500}},
          color=mapped color,
          fill=mapped color,
          draw=black,
          ] coordinates {
    (Engine A,3.7)
    (Engine B,4.6)
    (Engine C,3.2)
    (Engine D,6.8)
    (Engine E,5.7)
    (Engine F,5.3)
};

% Lower Upper Bound Estimation
\addplot+[
          /utils/exec={\pgfplotscolormapdefinemappedcolor{750}},
          color=mapped color,
          fill=mapped color,
          draw=black,
          ] coordinates {
    (Engine A,3.7)
    (Engine B,5.6)
    (Engine C,3.8)
    (Engine D,7.1)
    (Engine E,5.4)
    (Engine F,4.9)
};

% Mean Variance Estimation
\addplot+[
          /utils/exec={\pgfplotscolormapdefinemappedcolor{1000}},
          color=mapped color,
          fill=mapped color,
          draw=black,
          ] coordinates {
    (Engine A,5.1)
    (Engine B,4.5)
    (Engine C,3.5)
    (Engine D,5.6)
    (Engine E,6.4)
    (Engine F,6.1)
};

\legend{
  Bayesian,
  Bootstrap,
  Delta,
  LUBE,
  MVE
}

\end{axis}
\end{tikzpicture}
}
\caption{Standard deviation of TGT prediction error (absolute error) by uncertainty quantification method for six test engines (Engines A--F).}
\label{rv:fig:engine_sd_ae}
\end{figure}
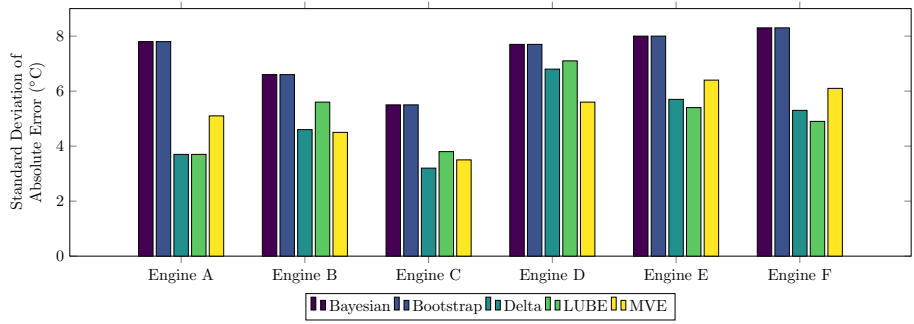

\Cref{rv:fig:engine_sd_ae} shows the variability of absolute error for each test engine. The same engines that exhibit higher MAE (notably Engine~D and Engine~F) also show higher dispersion in error, reflecting increased difficulty in those cases. Across most engines, Bayesian tends to have higher error variability than Delta, LUBE, and MVE, while MVE maintains relatively low dispersion, consistent with its strong overall accuracy.

%======================================
% MPIW by Method per 6 test engines
%======================================
\begin{figure}[H]
\centering
\resizebox{\textwidth}{!}{
\begin{tikzpicture}
\begin{axis}[
    ybar,
    bar width=9pt,
    ymin=20,
    ymax=75,
    enlarge x limits=0.18,
    width=20cm,
    height=7cm,
    ylabel={Mean Prediction Interval\\Width ($^\circ$C)},
    ylabel style={align=center},
    xlabel={},
    symbolic x coords={Engine A,Engine B,Engine C,Engine D,Engine E,Engine F},
    xtick=data,
    xticklabel style={align=center},
    legend style={
        at={(0.5,-0.15)},
        anchor=north,
        legend columns=-1
    },
    colormap/viridis,
    colormap name=viridis
]

% Bayesian
\addplot+[
          /utils/exec={\pgfplotscolormapdefinemappedcolor{0}},
          color=mapped color,
          fill=mapped color,
          draw=black,
          ] coordinates {
    (Engine A,71.5)
    (Engine B,73.5)
    (Engine C,65.4)
    (Engine D,73.9)
    (Engine E,72.8)
    (Engine F,69.1)
};

% Bootstrap
\addplot+[
          /utils/exec={\pgfplotscolormapdefinemappedcolor{250}},
          color=mapped color,
          fill=mapped color,
          draw=black,
          ] coordinates {
    (Engine A,71.5)
    (Engine B,73.5)
    (Engine C,65.4)
    (Engine D,73.9)
    (Engine E,72.8)
    (Engine F,69.1)
};

% Delta
\addplot+[
          /utils/exec={\pgfplotscolormapdefinemappedcolor{500}},
          color=mapped color,
          fill=mapped color,
          draw=black,
          ] coordinates {
    (Engine A,25.0)
    (Engine B,26.2)
    (Engine C,20.3)
    (Engine D,36.1)
    (Engine E,34.6)
    (Engine F,35.2)
};

% Lower Upper Bound Estimation
\addplot+[
          /utils/exec={\pgfplotscolormapdefinemappedcolor{750}},
          color=mapped color,
          fill=mapped color,
          draw=black,
          ] coordinates {
    (Engine A,39.3)
    (Engine B,38.7)
    (Engine C,39.6)
    (Engine D,39.9)
    (Engine E,38.8)
    (Engine F,40.4)
};

% Mean Variance Estimation
\addplot+[
          /utils/exec={\pgfplotscolormapdefinemappedcolor{1000}},
          color=mapped color,
          fill=mapped color,
          draw=black,
          ] coordinates {
    (Engine A,37.6)
    (Engine B,36.8)
    (Engine C,37.1)
    (Engine D,37.8)
    (Engine E,36.9)
    (Engine F,36.4)
};

\legend{
  Bayesian,
  Bootstrap,
  Delta,
  LUBE,
  MVE
}

\end{axis}
\end{tikzpicture}
}
\caption{Mean prediction interval width (MPIW) of TGT prediction intervals by uncertainty quantification method for six test engines (Engines A--F).}
\label{rv:fig:engine_mpiw}
\end{figure}
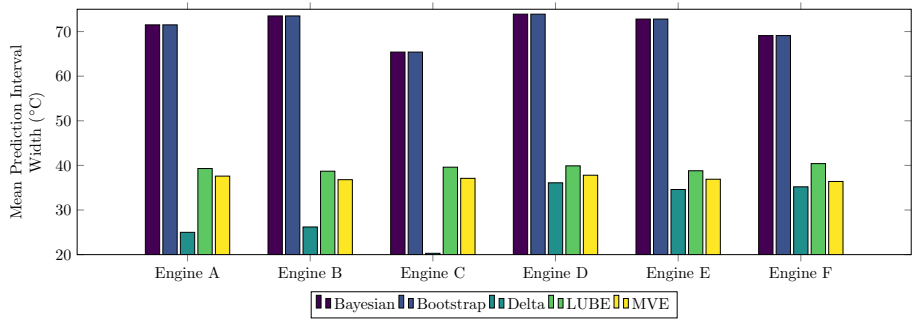

\Cref{rv:fig:engine_mpiw} reports MPIW on a per-engine basis. Interval widths vary across engines, with Engine~B generally producing the widest intervals for the Bayesian method. Delta remains constant across engines by design, while LUBE and MVE adapt their widths modestly from engine to engine, suggesting sensitivity to engine-specific uncertainty levels. This per-engine view complements the aggregate widths shown in \Cref{rv:fig:overall_mpiw}.

%======================================
% Std. Dev. of PI Width by Method per 6 test engines
%======================================
\begin{figure}[H]
\centering
\resizebox{\textwidth}{!}{
\begin{tikzpicture}
\begin{axis}[
    ybar,
    bar width=9pt,
    ymin=0,
    ymax=35,
    enlarge x limits=0.18,
    width=20cm,
    height=7cm,
    ylabel={Standard Deviation of\\Prediction Interval Width ($^\circ$C)},
    ylabel style={align=center},
    xlabel={},
    symbolic x coords={Engine A,Engine B,Engine C,Engine D,Engine E,Engine F},
    xtick=data,
    xticklabel style={align=center},
    legend style={
        at={(0.5,-0.15)},
        anchor=north,
        legend columns=-1
    },
    colormap/viridis,
    colormap name=viridis
]

% Bayesian (orange)
\addplot+[
          /utils/exec={\pgfplotscolormapdefinemappedcolor{0}},
          color=mapped color,
          fill=mapped color,
          draw=black,
          ] coordinates {
    (Engine A,30.5)
    (Engine B,31.5)
    (Engine C,24.8)
    (Engine D,32.3)
    (Engine E,31.8)
    (Engine F,26.0)
};

% Bootstrap
\addplot+[
          /utils/exec={\pgfplotscolormapdefinemappedcolor{250}},
          color=mapped color,
          fill=mapped color,
          draw=black,
          ] coordinates {
    (Engine A,30.5)
    (Engine B,31.5)
    (Engine C,24.8)
    (Engine D,32.3)
    (Engine E,31.8)
    (Engine F,26.0)
};

% Delta
\addplot+[
          /utils/exec={\pgfplotscolormapdefinemappedcolor{500}},
          color=mapped color,
          fill=mapped color,
          draw=black,
          ] coordinates {
    (Engine A,0.3)
    (Engine B,0.4)
    (Engine C,0.2)
    (Engine D,0.3)
    (Engine E,0.3)
    (Engine F,0.4)
};

% Lower Upper Bound Estimation
\addplot+[
          /utils/exec={\pgfplotscolormapdefinemappedcolor{750}},
          color=mapped color,
          fill=mapped color,
          draw=black,
          ] coordinates {
    (Engine A,6.6)
    (Engine B,7.0)
    (Engine C,8.9)
    (Engine D,7.1)
    (Engine E,7.2)
    (Engine F,8.4)
};

% Mean Variance Estimation
\addplot+[
          /utils/exec={\pgfplotscolormapdefinemappedcolor{1000}},
          color=mapped color,
          fill=mapped color,
          draw=black,
          ] coordinates {
    (Engine A,6.9)
    (Engine B,7.1)
    (Engine C,8.3)
    (Engine D,7.0)
    (Engine E,7.2)
    (Engine F,7.8)
};

\legend{
  Bayesian,
  Bootstrap,
  Delta,
  LUBE,
  MVE
}

\end{axis}
\end{tikzpicture}
}
\caption{Standard deviation of TGT prediction interval width by uncertainty quantification method for six test engines (Engines A--F).}
\label{rv:fig:engine_sd_piw}
\end{figure}
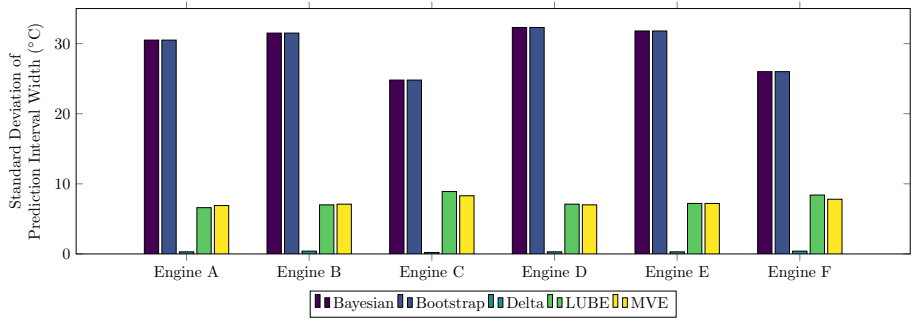

\Cref{rv:fig:engine_sd_piw} highlights within-engine variability of interval widths. Delta again shows zero variability, whereas Bayesian produces very large width variability for every engine, consistent with its stochastic sampling mechanism. LUBE and MVE display moderate variability that differs between engines, indicating that their learned bounds/variances respond to differences in local operating conditions within an engine's data.

%======================================
% NMPIW by Method per 6 test engines
%======================================
\begin{figure}[H]
\centering
\resizebox{\textwidth}{!}{
\begin{tikzpicture}
\begin{axis}[
    ybar,
    bar width=9pt,
    ymin=0,
    ymax=18,
    enlarge x limits=0.18,
    width=20cm,
    height=7cm,
    ylabel={Normalized Mean Prediction\\Interval Width (NMPIW) (\%)},
    ylabel style={align=center},
    xlabel={},
    symbolic x coords={Engine A,Engine B,Engine C,Engine D,Engine E,Engine F},
    xtick=data,
    xticklabel style={align=center},
    legend style={
        at={(0.5,-0.15)},
        anchor=north,
        legend columns=-1
    },
    colormap/viridis,
    colormap name=viridis
]

% Bayesian
\addplot+[
          /utils/exec={\pgfplotscolormapdefinemappedcolor{0}},
          color=mapped color,
          fill=mapped color,
          draw=black,
          ] coordinates {
    (Engine A,16.70)
    (Engine B,15.90)
    (Engine C,15.30)
    (Engine D,15.50)
    (Engine E,17.20)
    (Engine F,15.70)
};

% Bootstrap
\addplot+[
          /utils/exec={\pgfplotscolormapdefinemappedcolor{250}},
          color=mapped color,
          fill=mapped color,
          draw=black,
          ] coordinates {
    (Engine A,16.70)
    (Engine B,15.90)
    (Engine C,15.30)
    (Engine D,15.50)
    (Engine E,17.20)
    (Engine F,15.70)
};

% Delta
\addplot+[
          /utils/exec={\pgfplotscolormapdefinemappedcolor{500}},
          color=mapped color,
          fill=mapped color,
          draw=black,
          ] coordinates {
    (Engine A,5.80)
    (Engine B,5.60)
    (Engine C,4.80)
    (Engine D,7.50)
    (Engine E,8.10)
    (Engine F,7.90)
};

% Lower Upper Bound Estimation
\addplot+[
          /utils/exec={\pgfplotscolormapdefinemappedcolor{750}},
          color=mapped color,
          fill=mapped color,
          draw=black,
          ] coordinates {
    (Engine A,9.20)
    (Engine B,8.40)
    (Engine C,9.40)
    (Engine D,8.50)
    (Engine E,9.30)
    (Engine F,9.10)
};

% Mean Variance Estimation
\addplot+[
          /utils/exec={\pgfplotscolormapdefinemappedcolor{1000}},
          color=mapped color,
          fill=mapped color,
          draw=black,
          ] coordinates {
    (Engine A,8.70)
    (Engine B,8.00)
    (Engine C,8.60)
    (Engine D,7.90)
    (Engine E,8.80)
    (Engine F,8.30)
};

\legend{
  Bayesian,
  Bootstrap,
  Delta,
  LUBE,
  MVE
}

\end{axis}
\end{tikzpicture}
}
\caption{Normalized mean prediction interval width (NMPIW) by uncertainty quantification method for six test engines (Engines A--F).}
\label{rv:fig:engine_nmpiw}
\end{figure}
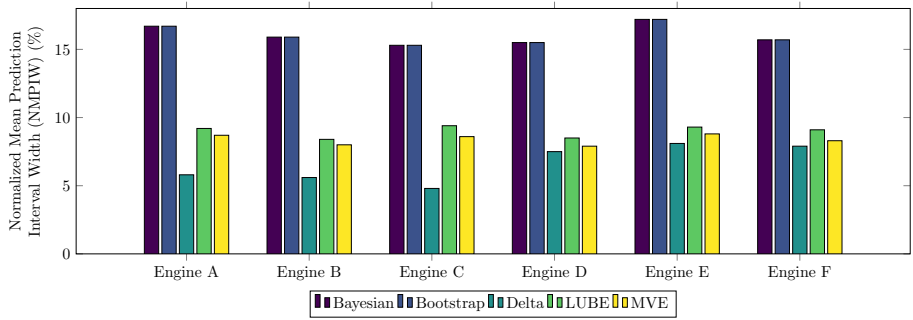

\Cref{rv:fig:engine_nmpiw} presents normalized widths (NMPIW) by engine. The normalized values show that Bayesian remains consistently the least sharp across all engines, while bootstrap is consistently the sharpest. Delta exhibits comparatively stable normalized width from engine to engine, whereas LUBE and MVE occupy an intermediate range and vary moderately with engine, reflecting adaptation to engine-specific uncertainty.

%======================================
% PICP by Method per 6 test engines
%======================================
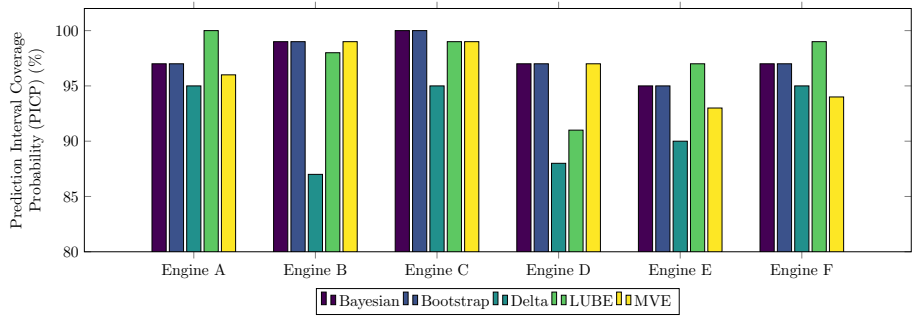
\begin{figure}[H]
\centering
\resizebox{\textwidth}{!}{
\begin{tikzpicture}
\begin{axis}[
    ybar,
    bar width=9pt,
    ymin=80,
    ymax=102,
    enlarge x limits=0.18,
    width=20cm,
    height=7cm,
    ylabel={Prediction Interval Coverage\\Probability (PICP) (\%)},
    ylabel style={align=center},
    xlabel={},
    symbolic x coords={Engine A,Engine B,Engine C,Engine D,Engine E,Engine F},
    xtick=data,
    xticklabel style={align=center},
    legend style={
        at={(0.5,-0.15)},
        anchor=north,
        legend columns=-1
    },
    colormap/viridis,
    colormap name=viridis
]

% Bayesian
\addplot+[
          /utils/exec={\pgfplotscolormapdefinemappedcolor{0}},
          color=mapped color,
          fill=mapped color,
          draw=black,
          ] coordinates {
    (Engine A,97.00)
    (Engine B,99.00)
    (Engine C,100.00)
    (Engine D,97.00)
    (Engine E,95.00)
    (Engine F,97.00)
};

% Bootstrap
\addplot+[
          /utils/exec={\pgfplotscolormapdefinemappedcolor{250}},
          color=mapped color,
          fill=mapped color,
          draw=black,
          ] coordinates {
    (Engine A,97.00)
    (Engine B,99.00)
    (Engine C,100.00)
    (Engine D,97.00)
    (Engine E,95.00)
    (Engine F,97.00)
};

% Delta
\addplot+[
          /utils/exec={\pgfplotscolormapdefinemappedcolor{500}},
          color=mapped color,
          fill=mapped color,
          draw=black,
          ] coordinates {
    (Engine A,95.00)
    (Engine B,87.00)
    (Engine C,95.00)
    (Engine D,88.00)
    (Engine E,90.00)
    (Engine F,95.00)
};

% Lower Upper Bound Estimation
\addplot+[
          /utils/exec={\pgfplotscolormapdefinemappedcolor{750}},
          color=mapped color,
          fill=mapped color,
          draw=black,
          ] coordinates {
    (Engine A,100.00)
    (Engine B,98.00)
    (Engine C,99.00)
    (Engine D,91.00)
    (Engine E,97.00)
    (Engine F,99.00)
};

% Mean Variance Estimation
\addplot+[
          /utils/exec={\pgfplotscolormapdefinemappedcolor{1000}},
          color=mapped color,
          fill=mapped color,
          draw=black,
          ] coordinates {
    (Engine A,96.00)
    (Engine B,99.00)
    (Engine C,99.00)
    (Engine D,97.00)
    (Engine E,93.00)
    (Engine F,94.00)
};

\legend{
  Bayesian,
  Bootstrap,
  Delta,
  LUBE,
  MVE
}

\end{axis}
\end{tikzpicture}
}
\caption{Prediction interval coverage probability (PICP) by uncertainty quantification method for six test engines (Engines A--F).}
\label{rv:fig:engine_picp}
\end{figure}

\Cref{rv:fig:engine_picp} reports coverage by test engine. LUBE and MVE maintain high coverage across all engines, while Delta under-covers on multiple engines (e.g., Engine~B and Engine~E), consistent with the global under-coverage observed in \Cref{rv:fig:overall_picp}. Bayesian coverage remains near nominal for most engines, indicating that its wide intervals translate into robust calibration even when point accuracy is weaker.

\subsection{Per-Phase Breakdown}

%======================================
% MAE by Method per Flight Phase
%======================================
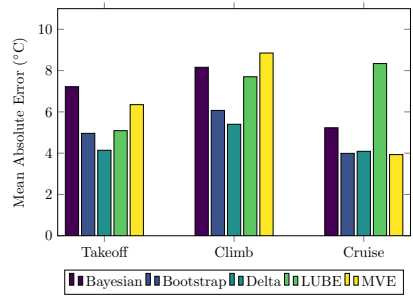
\begin{figure}[H]
\centering
\resizebox{0.45\textwidth}{!}{
\begin{tikzpicture}
\begin{axis}[
    ybar,
    bar width=9pt,
    ymin=0,
    ymax=11,
    enlarge x limits=0.18,
    width=10cm,
    height=7cm,
    ylabel={Mean Absolute Error ($^\circ$C)},
    xlabel={},
    symbolic x coords={Takeoff, Climb, Cruise},
    xtick=data,
    xticklabel style={align=center},
    legend style={
        at={(0.5,-0.15)},
        anchor=north,
        legend columns=-1
    },
    colormap/viridis,
    colormap name=viridis
]

% Bayesian
\addplot+[
          /utils/exec={\pgfplotscolormapdefinemappedcolor{0}},
          color=mapped color,
          fill=mapped color,
          draw=black,
          ] coordinates {
    (Takeoff,7.22)
    (Climb,8.16)
    (Cruise,5.23)
};

% Bootstrap
\addplot+[
          /utils/exec={\pgfplotscolormapdefinemappedcolor{250}},
          color=mapped color,
          fill=mapped color,
          draw=black,
          ] coordinates {
    (Takeoff,4.96)
    (Climb,6.07)
    (Cruise,3.99)
};

% Delta
\addplot+[
          /utils/exec={\pgfplotscolormapdefinemappedcolor{500}},
          color=mapped color,
          fill=mapped color,
          draw=black,
          ] coordinates {
    (Takeoff,4.14)
    (Climb,5.40)
    (Cruise,4.09)
};

% Lower Upper Bound Estimation
\addplot+[
          /utils/exec={\pgfplotscolormapdefinemappedcolor{750}},
          color=mapped color,
          fill=mapped color,
          draw=black,
          ] coordinates {
    (Takeoff,5.09)
    (Climb,7.70)
    (Cruise,8.34)
};

% Mean Variance Estimation
\addplot+[
          /utils/exec={\pgfplotscolormapdefinemappedcolor{1000}},
          color=mapped color,
          fill=mapped color,
          draw=black,
          ] coordinates {
    (Takeoff,6.35)
    (Climb,8.85)
    (Cruise,3.93)
};

\legend{
  Bayesian,
  Bootstrap,
  Delta,
  LUBE,
  MVE
}

\end{axis}
\end{tikzpicture}
}
\caption{Mean absolute TGT prediction error (MAE) by uncertainty quantification method for various flight phases (takeoff, climb, and cruise).}
\label{rv:fig:phase_mae}
\end{figure}

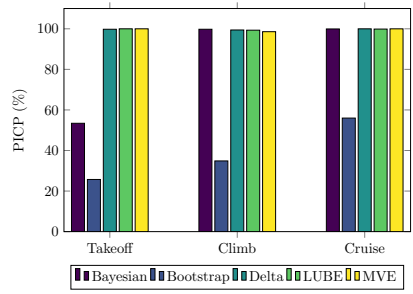
\begin{figure}[H]
\centering
\resizebox{0.45\textwidth}{!}{
\begin{tikzpicture}
\begin{axis}[
    ybar,
    bar width=9pt,
    ymin=0,
    ymax=110,
    enlarge x limits=0.18,
    width=10cm,
    height=7cm,
    ylabel={PICP (\%)},
    xlabel={},
    symbolic x coords={Takeoff, Climb, Cruise},
    xtick=data,
    xticklabel style={align=center},
    legend style={
        at={(0.5,-0.15)},
        anchor=north,
        legend columns=-1
    },
    colormap/viridis,
    colormap name=viridis
]

% Bayesian
\addplot+[
          /utils/exec={\pgfplotscolormapdefinemappedcolor{0}},
          color=mapped color,
          fill=mapped color,
          draw=black,
          ] coordinates {
    (Takeoff,53.42)
    (Climb,99.78)
    (Cruise,99.93)
};

% Bootstrap
\addplot+[
          /utils/exec={\pgfplotscolormapdefinemappedcolor{250}},
          color=mapped color,
          fill=mapped color,
          draw=black,
          ] coordinates {
    (Takeoff,25.72)
    (Climb,34.86)
    (Cruise,55.99)
};

% Delta
\addplot+[
          /utils/exec={\pgfplotscolormapdefinemappedcolor{500}},
          color=mapped color,
          fill=mapped color,
          draw=black,
          ] coordinates {
    (Takeoff,99.78)
    (Climb,99.42)
    (Cruise,99.99)
};

% Lower Upper Bound Estimation
\addplot+[
          /utils/exec={\pgfplotscolormapdefinemappedcolor{750}},
          color=mapped color,
          fill=mapped color,
          draw=black,
          ] coordinates {
    (Takeoff,100)
    (Climb,99.32)
    (Cruise,99.87)
};

% Mean Variance Estimation
\addplot+[
          /utils/exec={\pgfplotscolormapdefinemappedcolor{1000}},
          color=mapped color,
          fill=mapped color,
          draw=black,
          ] coordinates {
    (Takeoff,100)
    (Climb,98.58)
    (Cruise,99.99)
};

\legend{
  Bayesian,
  Bootstrap,
  Delta,
  LUBE,
  MVE
}

\end{axis}
\end{tikzpicture}
}
\caption{Prediction interval coverage probability (PICP) by uncertainty quantification method for various flight phases (takeoff, climb, and cruise).}
\label{rv:fig:phase_picp}
\end{figure}

\Cref{rv:fig:phase_mae} shows MAE stratified by flight phase. Errors are highest in takeoff and climb and decrease into cruise, consistent with cruise being the most stable regime. Delta achieves the lowest MAE in takeoff and climb, while MVE provides the lowest MAE in cruise. Bootstrap is also competitive in MAE across phases, but its uncertainty estimates remain poorly calibrated (see \Cref{rv:fig:phase_picp}), indicating that strong point accuracy does not guarantee reliable prediction intervals.

%======================================
% Std. Dev. of AE by Method per Flight Phase
%======================================
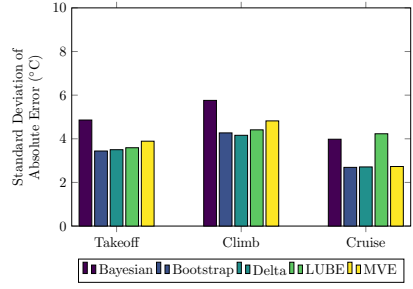
\begin{figure}[H]
\centering
\resizebox{0.45\textwidth}{!}{
\begin{tikzpicture}
\begin{axis}[
    ybar,
    bar width=9pt,
    ymin=0,
    ymax=10,
    enlarge x limits=0.18,
    width=10cm,
    height=7cm,
    ylabel={Standard Deviation of\\Absolute Error ($^\circ$C)},
    ylabel style={align=center},
    xlabel={},
    symbolic x coords={Takeoff, Climb, Cruise},
    xtick=data,
    xticklabel style={align=center},
    legend style={
        at={(0.5,-0.15)},
        anchor=north,
        legend columns=-1
    },
    colormap/viridis,
    colormap name=viridis
]

% Bayesian
\addplot+[
          /utils/exec={\pgfplotscolormapdefinemappedcolor{0}},
          color=mapped color,
          fill=mapped color,
          draw=black,
          ] coordinates {
    (Takeoff,4.86)
    (Climb,5.76)
    (Cruise,3.98)
};

% Bootstrap
\addplot+[
          /utils/exec={\pgfplotscolormapdefinemappedcolor{250}},
          color=mapped color,
          fill=mapped color,
          draw=black,
          ] coordinates {
    (Takeoff,3.44)
    (Climb,4.27)
    (Cruise,2.69)
};

% Delta
\addplot+[
          /utils/exec={\pgfplotscolormapdefinemappedcolor{500}},
          color=mapped color,
          fill=mapped color,
          draw=black,
          ] coordinates {
    (Takeoff,3.50)
    (Climb,4.16)
    (Cruise,2.71)
};

% Lower Upper Bound Estimation
\addplot+[
          /utils/exec={\pgfplotscolormapdefinemappedcolor{750}},
          color=mapped color,
          fill=mapped color,
          draw=black,
          ] coordinates {
    (Takeoff,3.59)
    (Climb,4.41)
    (Cruise,4.23)
};

% Mean Variance Estimation
\addplot+[
          /utils/exec={\pgfplotscolormapdefinemappedcolor{1000}},
          color=mapped color,
          fill=mapped color,
          draw=black,
          ] coordinates {
    (Takeoff,3.89)
    (Climb,4.82)
    (Cruise,2.73)
};

\legend{
  Bayesian,
  Bootstrap,
  Delta,
  LUBE,
  MVE
}

\end{axis}
\end{tikzpicture}
}
\caption{Standard deviation of TGT prediction error (absolute error) by uncertainty quantification method for various flight phases (takeoff, climb, and cruise).}
\label{rv:fig:phase_sd_ae}
\end{figure}

\Cref{rv:fig:phase_sd_ae} reports the phase-wise dispersion of absolute error. Takeoff exhibits the highest error variability, reflecting strong transients and rapid changes in operating conditions. Bayesian and LUBE show elevated variability in climb relative to the other methods, while MVE and Delta maintain more moderate variability across phases. Lower variability is desirable for consistent decision thresholds in EHM applications.

%======================================
% MPIW by Method per Flight Phase
%======================================
\begin{figure}[H]
\centering
\resizebox{0.45\textwidth}{!}{
\begin{tikzpicture}
\begin{axis}[
    ybar,
    bar width=9pt,
    ymin=0,
    ymax=80,
    enlarge x limits=0.18,
    width=10cm,
    height=7cm,
    ylabel={Mean Prediction Interval\\Width ($^\circ$C)},
    ylabel style={align=center},
    xlabel={},
    symbolic x coords={Takeoff, Climb, Cruise},
    xtick=data,
    xticklabel style={align=center},
    legend style={
        at={(0.5,-0.15)},
        anchor=north,
        legend columns=-1
    },
    colormap/viridis,
    colormap name=viridis
]

% Bayesian
\addplot+[
          /utils/exec={\pgfplotscolormapdefinemappedcolor{0}},
          color=mapped color,
          fill=mapped color,
          draw=black,
          ] coordinates {
    (Takeoff,19.29)
    (Climb,63.53)
    (Cruise,46.83)
};

% Bootstrap
\addplot+[
          /utils/exec={\pgfplotscolormapdefinemappedcolor{250}},
          color=mapped color,
          fill=mapped color,
          draw=black,
          ] coordinates {
    (Takeoff,4.61)
    (Climb,6.80)
    (Cruise,8.48)
};

% Delta
\addplot+[
          /utils/exec={\pgfplotscolormapdefinemappedcolor{500}},
          color=mapped color,
          fill=mapped color,
          draw=black,
          ] coordinates {
    (Takeoff,37.40)
    (Climb,37.94)
    (Cruise,37.91)
};

% Lower Upper Bound Estimation
\addplot+[
          /utils/exec={\pgfplotscolormapdefinemappedcolor{750}},
          color=mapped color,
          fill=mapped color,
          draw=black,
          ] coordinates {
    (Takeoff,41.49)
    (Climb,43.00)
    (Cruise,43.26)
};

% Mean Variance Estimation
\addplot+[
          /utils/exec={\pgfplotscolormapdefinemappedcolor{1000}},
          color=mapped color,
          fill=mapped color,
          draw=black,
          ] coordinates {
    (Takeoff,64.77)
    (Climb,44.47)
    (Cruise,33.45)
};

\legend{
  Bayesian,
  Bootstrap,
  Delta,
  LUBE,
  MVE
}

\end{axis}
\end{tikzpicture}
}
\caption{Mean prediction interval width (MPIW) of TGT prediction intervals by uncertainty quantification method for various flight phases (takeoff, climb, and cruise).}
\label{rv:fig:phase_mpiw}
\end{figure}
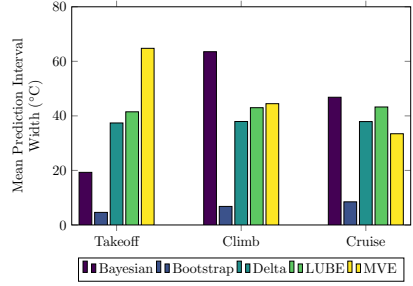

\Cref{rv:fig:phase_mpiw} compares interval widths by phase. Bayesian exhibits the strongest regime sensitivity, with much wider intervals in climb (63.53$^\circ$C) than in takeoff (19.29$^\circ$C) or cruise (46.83$^\circ$C). Delta and LUBE remain approximately constant-width across phases, while MVE is widest in takeoff and narrows substantially into cruise. Bootstrap produces the narrowest intervals in every phase, but these intervals are not calibrated (low PICP; \Cref{rv:fig:phase_picp}).

%======================================
% NMPIW by Method per Flight Phase
%======================================
\begin{figure}[H]
\centering
\resizebox{0.45\textwidth}{!}{
\begin{tikzpicture}
\begin{axis}[
    ybar,
    bar width=9pt,
    ymin=0,
    ymax=30,
    enlarge x limits=0.18,
    width=10cm,
    height=7cm,
    ylabel={NMPIW (\%)},
    xlabel={},
    symbolic x coords={Takeoff, Climb, Cruise},
    xtick=data,
    xticklabel style={align=center},
    legend style={
        at={(0.5,-0.15)},
        anchor=north,
        legend columns=-1
    },
    colormap/viridis,
    colormap name=viridis
]

% Bayesian
\addplot+[
          /utils/exec={\pgfplotscolormapdefinemappedcolor{0}},
          color=mapped color,
          fill=mapped color,
          draw=black,
          ] coordinates {
    (Takeoff,5.48)
    (Climb,26.69)
    (Cruise,18.28)
};

% Bootstrap
\addplot+[
          /utils/exec={\pgfplotscolormapdefinemappedcolor{250}},
          color=mapped color,
          fill=mapped color,
          draw=black,
          ] coordinates {
    (Takeoff,1.31)
    (Climb,2.86)
    (Cruise,3.31)
};

% Delta
\addplot+[
          /utils/exec={\pgfplotscolormapdefinemappedcolor{500}},
          color=mapped color,
          fill=mapped color,
          draw=black,
          ] coordinates {
    (Takeoff,10.63)
    (Climb,15.94)
    (Cruise,14.80)
};

% Lower Upper Bound Estimation
\addplot+[
          /utils/exec={\pgfplotscolormapdefinemappedcolor{750}},
          color=mapped color,
          fill=mapped color,
          draw=black,
          ] coordinates {
    (Takeoff,11.79)
    (Climb,18.07)
    (Cruise,16.89)
};

% Mean Variance Estimation
\addplot+[
          /utils/exec={\pgfplotscolormapdefinemappedcolor{1000}},
          color=mapped color,
          fill=mapped color,
          draw=black,
          ] coordinates {
    (Takeoff,18.40)
    (Climb,18.69)
    (Cruise,13.06)
};

\legend{
  Bayesian,
  Bootstrap,
  Delta,
  LUBE,
  MVE
}

\end{axis}
\end{tikzpicture}
}
\caption{Normalized mean prediction interval width (NMPIW) by uncertainty quantification method for various flight phases (takeoff, climb, and cruise).}
\label{rv:fig:phase_nmpiw}
\end{figure}
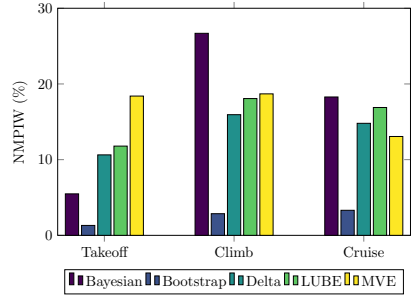

\Cref{rv:fig:phase_nmpiw} presents the normalized width comparison by phase. The same pattern persists: climb has the largest normalized widths for the calibrated methods, and bootstrap remains the sharpest across all phases. The increase in normalized width for LUBE and MVE in climb aligns with their ability to maintain near-nominal coverage, whereas methods that remain sharp in climb risk under-coverage.

%======================================
% PICP by Method per Flight Phase
%======================================

\Cref{rv:fig:phase_picp} shows phase-wise coverage. Delta, LUBE, and MVE maintain near-perfect coverage across all phases (98.58--100\%), whereas Bayesian achieves very high coverage in climb and cruise but drops sharply in takeoff, indicating that MC-dropout alone does not capture all sources of uncertainty in the most transient regime. Bootstrap under-covers in every phase, confirming that additional calibration is required if bootstrap ensembles are to be used for reliable prediction intervals.

%%%%%%%%%%%%%%%%%%%%%%%%%%%%%%%%%%%%%%%%%%%%%%%%%%%%%%%%
\subsection{Takeoff Phase}

This section provides additional bar-chart views that focus exclusively on takeoff. These figures mirror the takeoff row of \Cref{rv:tab:pi_comparison} but make method-to-method contrasts easier to see at a glance.

%===================================================
% Mean Absolute TGT Error Barchart for Takeoff Phase
%===================================================
\begin{figure}[H]
\centering
\resizebox{0.45\textwidth}{!}{
\begin{tikzpicture}
\begin{axis}[
    at={(6,0)},
    anchor=north west,
    width=10cm,
    height=7cm,
    ymin=0,
    ymax=10,
    enlarge x limits=0.4,
    ylabel={Mean Absolute Error ($^\circ$C)},
    symbolic x coords={Bayesian,Bootstrap,Delta,LUBE,MVE},
    xtick={Bayesian,Bootstrap,Delta,LUBE,MVE},
    xticklabel style={align=center, rotate=-45},
    nodes near coords,
    nodes near coords align={vertical},
    colormap/viridis,          % <-- use viridis colormap
    colormap name = viridis
]

\addplot+[ybar, bar width=14pt,
          /utils/exec={\pgfplotscolormapdefinemappedcolor{0}},
          color=mapped color,
          fill=mapped color,
          draw=black,
          every node near coord/.append style={text=black}, 
          mark=-] coordinates {(Bayesian,7.22)};

\addplot+[ybar, bar width=14pt,
          /utils/exec={\pgfplotscolormapdefinemappedcolor{250}},
          color=mapped color,
          fill=mapped color,
          draw=black,
          every node near coord/.append style={text=black}, 
          mark=-] coordinates {(Bootstrap,4.96)};

\addplot+[ybar, bar width=14pt,
          /utils/exec={\pgfplotscolormapdefinemappedcolor{500}},
          color=mapped color,
          fill=mapped color,
          draw=black,
          every node near coord/.append style={text=black}, 
          mark=-] coordinates {(Delta,4.14)};

\addplot+[ybar, bar width=14pt,
          /utils/exec={\pgfplotscolormapdefinemappedcolor{750}},
          color=mapped color,
          fill=mapped color,
          draw=black,
          every node near coord/.append style={text=black}, 
          mark=-] coordinates {(LUBE,5.09)};
          
% sample 5 distinct colors from viridis at positions 0..1000
\addplot+[ybar, bar width=14pt,
          /utils/exec={\pgfplotscolormapdefinemappedcolor{1000}},
          color=mapped color,
          fill=mapped color,
          draw=black,
          every node near coord/.append style={text=black}, 
          %every node near coord/.append style={text=mapped color}, 
          mark=-] coordinates {(MVE,6.35)};

\end{axis}
\end{tikzpicture}
}
\caption{Takeoff Phase. Comparison of the mean absolute TGT prediction error for the uncertainty quantification methodologies.}
\label{rv:fig:app_takeoff_mae}
\end{figure}
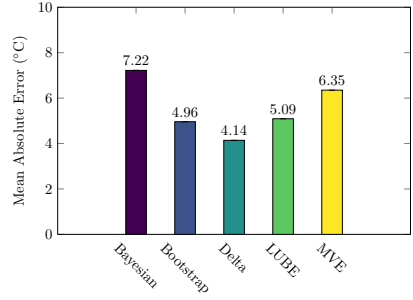

\Cref{rv:fig:app_takeoff_mae} provides a takeoff-only view of point-prediction accuracy. Delta achieves the lowest MAE in takeoff (4.14$^\circ$C), with bootstrap close behind, while Bayesian and MVE exhibit larger errors. This reinforces that point-prediction accuracy varies substantially by method in the most transient regime.

%====================================================
% S.D. Absolute TGT Error Barchart for Takeoff Phase
%====================================================
\begin{figure}[H]
\centering
\resizebox{0.45\textwidth}{!}{
\begin{tikzpicture}
\begin{axis}[
    at={(6,0)},
    anchor=north west,
    width=10cm,
    height=7cm,
    ymin=0,
    ymax=10,
    enlarge x limits=0.4,
    ylabel={Standard Deviation of\\Absolute Error ($^\circ$C)},
    ylabel style={align=center},
    symbolic x coords={Bayesian,Bootstrap,Delta,LUBE,MVE},
    xtick={Bayesian,Bootstrap,Delta,LUBE,MVE},
    xticklabel style={align=center, rotate=-45},
    nodes near coords,
    nodes near coords align={vertical},
    colormap/viridis,
    colormap name=viridis
]

% Bayesian
\addplot+[ybar, bar width=14pt,
          /utils/exec={\pgfplotscolormapdefinemappedcolor{0}},
          color=mapped color,
          fill=mapped color,
          draw=black,
          every node near coord/.append style={text=black}, 
          mark=-] coordinates {(Bayesian,4.86)};

% Bootstrap
\addplot+[ybar, bar width=14pt,
          /utils/exec={\pgfplotscolormapdefinemappedcolor{250}},
          color=mapped color,
          fill=mapped color,
          draw=black,
          every node near coord/.append style={text=black}, 
          mark=-] coordinates {(Bootstrap,3.44)};

% Delta
\addplot+[ybar, bar width=14pt,
          /utils/exec={\pgfplotscolormapdefinemappedcolor{500}},
          color=mapped color,
          fill=mapped color,
          draw=black,
          every node near coord/.append style={text=black}, 
          mark=-] coordinates {(Delta,3.5)};

% LUBE
\addplot+[ybar, bar width=14pt,
          /utils/exec={\pgfplotscolormapdefinemappedcolor{750}},
          color=mapped color,
          fill=mapped color,
          draw=black,
          every node near coord/.append style={text=black}, 
          mark=-] coordinates {(LUBE,3.59)};

% MVE
\addplot+[ybar, bar width=14pt,
          /utils/exec={\pgfplotscolormapdefinemappedcolor{1000}},
          color=mapped color,
          fill=mapped color,
          draw=black,
          every node near coord/.append style={text=black}, 
          mark=-] coordinates {(MVE,3.89)};

\end{axis}
\end{tikzpicture}
}
\caption{Takeoff Phase. Comparison of the standard deviation of absolute TGT prediction error for the uncertainty quantification methodologies.}
\label{rv:fig:app_takeoff_sd_ae}
\end{figure}
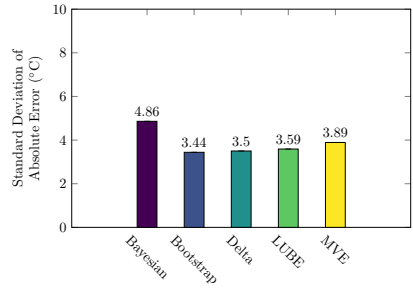

\Cref{rv:fig:app_takeoff_sd_ae} shows that the dispersion of absolute error in takeoff, reflecting rapid changes in operating conditions. Bayesian exhibits the largest S.D.\ AE, indicating less consistent point accuracy during takeoff, while the remaining methods cluster at lower variability.

%======================================
% MPIW Barchart for Takeoff Phase
%======================================
\begin{figure}[H]
\centering
\resizebox{0.45\textwidth}{!}{
\begin{tikzpicture}
\begin{axis}[
    at={(6,0)},
    anchor=north west,
    width=10cm,
    height=7cm,
    ymin=0,
    ymax=80,
    enlarge x limits=0.4,
    ylabel={Mean Prediction Interval\\Width (MPIW) ($^\circ$C)},
    ylabel style={align=center},
    symbolic x coords={Bayesian,Bootstrap,Delta,LUBE,MVE},
    xtick={Bayesian,Bootstrap,Delta,LUBE,MVE},
    xticklabel style={align=center, rotate=-45},
    nodes near coords,
    nodes near coords align={vertical},
    colormap/viridis,
    colormap name=viridis
]

% Bayesian
\addplot+[ybar, bar width=14pt,
          /utils/exec={\pgfplotscolormapdefinemappedcolor{0}},
          color=mapped color,
          fill=mapped color,
          draw=black,
          every node near coord/.append style={text=black}, 
          mark=-] coordinates {(Bayesian,19.29)};

% Bootstrap
\addplot+[ybar, bar width=14pt,
          /utils/exec={\pgfplotscolormapdefinemappedcolor{250}},
          color=mapped color,
          fill=mapped color,
          draw=black,
          every node near coord/.append style={text=black}, 
          mark=-] coordinates {(Bootstrap,4.61)};

% Delta
\addplot+[ybar, bar width=14pt,
          /utils/exec={\pgfplotscolormapdefinemappedcolor{500}},
          color=mapped color,
          fill=mapped color,
          draw=black,
          every node near coord/.append style={text=black}, 
          mark=-] coordinates {(Delta,37.40)};

% LUBE
\addplot+[ybar, bar width=14pt,
          /utils/exec={\pgfplotscolormapdefinemappedcolor{750}},
          color=mapped color,
          fill=mapped color,
          draw=black,
          every node near coord/.append style={text=black}, 
          mark=-] coordinates {(LUBE,41.49)};

% MVE
\addplot+[ybar, bar width=14pt,
          /utils/exec={\pgfplotscolormapdefinemappedcolor{1000}},
          color=mapped color,
          fill=mapped color,
          draw=black,
          every node near coord/.append style={text=black}, 
          mark=-] coordinates {(MVE,64.77)};

\end{axis}
\end{tikzpicture}
}
\caption{Takeoff Phase. Comparison of the mean prediction interval width (MPIW) for the uncertainty quantification methodologies.}
\label{rv:fig:app_takeoff_mpiw}
\end{figure}
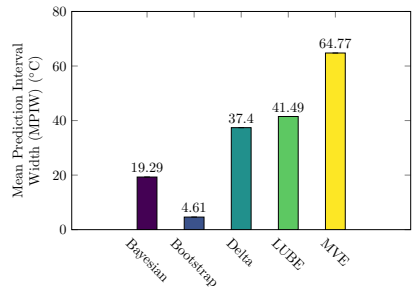

\Cref{rv:fig:app_takeoff_mpiw} compares interval widths in takeoff. Bootstrap produces the narrowest intervals, whereas MVE produces the widest intervals to achieve full coverage. Delta and LUBE occupy an intermediate width regime, providing near-perfect coverage without becoming as conservative as MVE.

%======================================
% NMPIW Barchart for Takeoff Phase
%======================================
\begin{figure}[H]
\centering
\resizebox{0.45\columnwidth}{!}{
\begin{tikzpicture}
\begin{axis}[
    at={(6,0)},
    anchor=north west,
    width=10cm,
    height=7cm,
    ymin=0,
    ymax=30,
    enlarge x limits=0.4,
    ylabel={Normalized Mean Prediction\\Interval Width (NMPIW) (\%)},
    ylabel style={align=center},
    symbolic x coords={Bayesian,Bootstrap,Delta,LUBE,MVE},
    xtick={Bayesian,Bootstrap,Delta,LUBE,MVE},
    xticklabel style={align=center, rotate=-45},
    nodes near coords,
    nodes near coords align={vertical},
    colormap/viridis,
    colormap name=viridis
]

% Bayesian
\addplot+[ybar, bar width=14pt,
          /utils/exec={\pgfplotscolormapdefinemappedcolor{0}},
          color=mapped color,
          fill=mapped color,
          draw=black,
          every node near coord/.append style={text=black}, 
          mark=-] coordinates {(Bayesian,5.48)};

% Bootstrap
\addplot+[ybar, bar width=14pt,
          /utils/exec={\pgfplotscolormapdefinemappedcolor{250}},
          color=mapped color,
          fill=mapped color,
          draw=black,
          every node near coord/.append style={text=black}, 
          mark=-] coordinates {(Bootstrap,1.31)};

% Delta
\addplot+[ybar, bar width=14pt,
          /utils/exec={\pgfplotscolormapdefinemappedcolor{500}},
          color=mapped color,
          fill=mapped color,
          draw=black,
          every node near coord/.append style={text=black}, 
          mark=-] coordinates {(Delta,10.63)};

% LUBE
\addplot+[ybar, bar width=14pt,
          /utils/exec={\pgfplotscolormapdefinemappedcolor{750}},
          color=mapped color,
          fill=mapped color,
          draw=black,
          every node near coord/.append style={text=black}, 
          mark=-] coordinates {(LUBE,11.79)};

% MVE
\addplot+[ybar, bar width=14pt,
          /utils/exec={\pgfplotscolormapdefinemappedcolor{1000}},
          color=mapped color,
          fill=mapped color,
          draw=black,
          every node near coord/.append style={text=black}, 
          mark=-] coordinates {(MVE,18.40)};

\end{axis}
\end{tikzpicture}
}
\caption{Takeoff Phase. Comparison of the normalized mean prediction interval width (NMPIW) for the uncertainty quantification methodologies.}
\label{rv:fig:app_takeoff_nmpiw}
\end{figure}
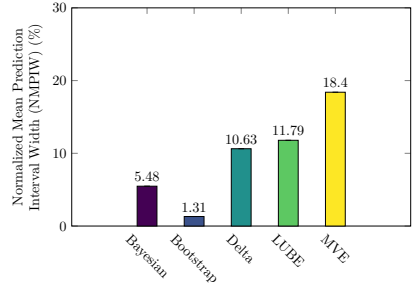

\Cref{rv:fig:app_takeoff_nmpiw} presents the same takeoff width comparison in normalized form. The normalized view confirms that bootstrap remains extremely sharp, and that MVE's conservatism in takeoff is not an artifact of scaling.

%======================================
% PICP Barchart for Takeoff Phase
%======================================
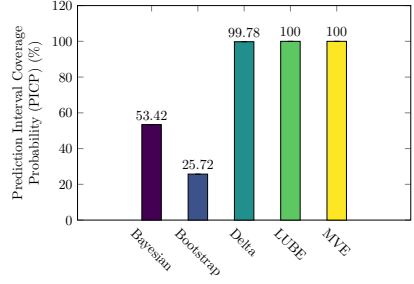
\begin{figure}[H]
\centering
\resizebox{0.45\columnwidth}{!}{
\begin{tikzpicture}
\begin{axis}[
    at={(6,0)},
    anchor=north west,
    width=10cm,
    height=7cm,
    ymin=0,
    ymax=120,
    enlarge x limits=0.4,
    ylabel={Prediction Interval Coverage\\Probability (PICP) (\%)},
    ylabel style={align=center},
    symbolic x coords={Bayesian,Bootstrap,Delta,LUBE,MVE},
    xtick={Bayesian,Bootstrap,Delta,LUBE,MVE},
    xticklabel style={align=center, rotate=-45},
    nodes near coords,
    nodes near coords align={vertical},
    colormap/viridis,
    colormap name=viridis
]

% Bayesian
\addplot+[ybar, bar width=14pt,
          /utils/exec={\pgfplotscolormapdefinemappedcolor{0}},
          color=mapped color,
          fill=mapped color,
          draw=black,
          every node near coord/.append style={text=black}, 
          mark=-] coordinates {(Bayesian,53.42)};

% Bootstrap
\addplot+[ybar, bar width=14pt,
          /utils/exec={\pgfplotscolormapdefinemappedcolor{250}},
          color=mapped color,
          fill=mapped color,
          draw=black,
          every node near coord/.append style={text=black}, 
          mark=-] coordinates {(Bootstrap,25.72)};

% Delta
\addplot+[ybar, bar width=14pt,
          /utils/exec={\pgfplotscolormapdefinemappedcolor{500}},
          color=mapped color,
          fill=mapped color,
          draw=black,
          every node near coord/.append style={text=black}, 
          mark=-] coordinates {(Delta,99.78)};

% LUBE
\addplot+[ybar, bar width=14pt,
          /utils/exec={\pgfplotscolormapdefinemappedcolor{750}},
          color=mapped color,
          fill=mapped color,
          draw=black,
          every node near coord/.append style={text=black}, 
          mark=-] coordinates {(LUBE,100)};

% MVE
\addplot+[ybar, bar width=14pt,
          /utils/exec={\pgfplotscolormapdefinemappedcolor{1000}},
          color=mapped color,
          fill=mapped color,
          draw=black,
          every node near coord/.append style={text=black}, 
          mark=-] coordinates {(MVE,100)};

\end{axis}
\end{tikzpicture}
}
\caption{Takeoff Phase. Comparison of the prediction interval coverage probability (PICP) for the uncertainty quantification methodologies.}
\label{rv:fig:app_takeoff_picp}
\end{figure}

\Cref{rv:fig:app_takeoff_picp} highlights the core calibration challenge in takeoff. LUBE and MVE achieve perfect coverage, Delta achieves near-perfect coverage, whereas Bayesian and bootstrap exhibit severe under-coverage. This illustrates that epistemic-only mechanisms (MC-dropout and bootstrap ensembles) can under-estimate uncertainty in a transient regime unless paired with additional calibration.

%======================================
% CWC Barchart for Takeoff Phase
%======================================
\begin{figure}[H]
\centering
\resizebox{0.45\textwidth}{!}{
\begin{tikzpicture}
\begin{axis}[
    at={(6,0)},
    anchor=north west,
    width=10cm,
    height=7cm,
    ymin=0,
    ymax=160,
    enlarge x limits=0.4,
    ylabel={Coverage Width-based\\Criterion (CWC)},
    ylabel style={align=center},
    symbolic x coords={Bayesian,Bootstrap,Delta,LUBE,MVE},
    xtick={Bayesian,Bootstrap,Delta,LUBE,MVE},
    xticklabel style={align=center, rotate=-45},
    nodes near coords,
    nodes near coords align={vertical},
    colormap/viridis,
    colormap name=viridis
]

% Bayesian
\addplot+[ybar, bar width=14pt,
          /utils/exec={\pgfplotscolormapdefinemappedcolor{0}},
          color=mapped color,
          fill=mapped color,
          draw=black,
          nodes near coords={2.06e10},
          every node near coord/.append style={text=black,
            yshift=1pt, rotate=45, anchor=south west}, 
          mark=-] coordinates {(Bayesian,110)};

% Bootstrap
\addplot+[ybar, bar width=14pt,
          /utils/exec={\pgfplotscolormapdefinemappedcolor{250}},
          color=mapped color,
          fill=mapped color,
          % show true CWC value instead of 100
          nodes near coords={5.10e15},
          draw=black,
          every node near coord/.append style={text=black,
            yshift=1pt, rotate=45, anchor=south west}, 
          mark=-] coordinates {(Bootstrap,110)};

% Delta
\addplot+[ybar, bar width=14pt,
          /utils/exec={\pgfplotscolormapdefinemappedcolor{500}},
          color=mapped color,
          fill=mapped color,
          draw=black,
          every node near coord/.append style={text=black}, 
          mark=-] coordinates {(Delta,37.40)};

% LUBE
\addplot+[ybar, bar width=14pt,
          /utils/exec={\pgfplotscolormapdefinemappedcolor{750}},
          color=mapped color,
          fill=mapped color,
          draw=black,
          every node near coord/.append style={text=black}, 
          mark=-] coordinates {(LUBE,41.49)};

% MVE
\addplot+[ybar, bar width=14pt,
          /utils/exec={\pgfplotscolormapdefinemappedcolor{1000}},
          color=mapped color,
          fill=mapped color,
          draw=black,
          every node near coord/.append style={text=black}, 
          mark=-] coordinates {(MVE,64.77)};

\end{axis}
\end{tikzpicture}
}
\caption{Takeoff Phase. Comparison of the coverage width-based criterion (CWC) for the uncertainty quantification methodologies.}
\label{rv:fig:app_takeoff_cwc}
\end{figure}
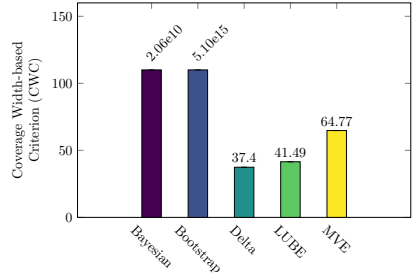

Finally, \Cref{rv:fig:app_takeoff_cwc} combines coverage and width. Delta achieves the most favorable takeoff CWC by meeting coverage with moderate widths, while bootstrap and Bayesian incur extremely large CWC values due to under-coverage penalties. This further supports the conclusion that calibrated intervals, rather than sharp intervals alone, are essential for risk-aware engine health management.
%%%%%%%%%%%%%%%%%%%%%%%%%%%%%%%%%%%%%%%%%%%%%%%%%%%%%%%%

%%%%%%%%%%%%%%%%%%%%%%%%%%%%%%%%%%%%%%%%%%%%%%%%%%%%%%%%
\subsection{Climb Phase}

This section provides additional bar-chart views that focus exclusively on climb. These figures mirror the climb row of \Cref{rv:tab:pi_metrics} but make method-to-method contrasts easier to see at a glance.

%===================================================
% Mean Absolute TGT Error Barchart for Climb Phase
%===================================================
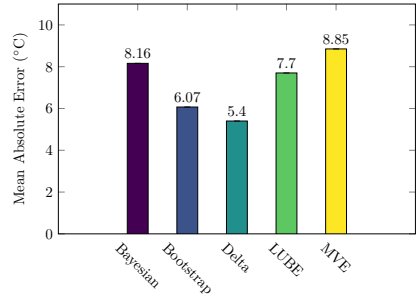
\begin{figure}[H]
        \centering
        \resizebox{0.45\textwidth}{!}{%
        \begin{tikzpicture}
        \begin{axis}[
            at={(6,0)},
            anchor=north west,
            width=10cm,
            height=7cm,
            ymin=0,
            ymax=11,
            enlarge x limits=0.4,
            ylabel={Mean Absolute Error ($^\circ$C)},
            symbolic x coords={Bayesian,Bootstrap,Delta,LUBE,MVE},
            xtick={Bayesian,Bootstrap,Delta,LUBE,MVE},
            xticklabel style={align=center, rotate=-45},
            nodes near coords,
            nodes near coords align={vertical},
            colormap/viridis,
            colormap name=viridis
]

        % Bayesian
\addplot+[ybar, bar width=14pt,
          /utils/exec={\pgfplotscolormapdefinemappedcolor{0}},
          color=mapped color,
          fill=mapped color,
          draw=black,
          every node near coord/.append style={text=black}, 
          mark=-] coordinates {(Bayesian,8.16)};

% Bootstrap
\addplot+[ybar, bar width=14pt,
          /utils/exec={\pgfplotscolormapdefinemappedcolor{250}},
          color=mapped color,
          fill=mapped color,
          draw=black,
          every node near coord/.append style={text=black}, 
          mark=-] coordinates {(Bootstrap,6.07)};

% Delta
\addplot+[ybar, bar width=14pt,
          /utils/exec={\pgfplotscolormapdefinemappedcolor{500}},
          color=mapped color,
          fill=mapped color,
          draw=black,
          every node near coord/.append style={text=black}, 
          mark=-] coordinates {(Delta,5.40)};

% LUBE
\addplot+[ybar, bar width=14pt,
          /utils/exec={\pgfplotscolormapdefinemappedcolor{750}},
          color=mapped color,
          fill=mapped color,
          draw=black,
          every node near coord/.append style={text=black}, 
          mark=-] coordinates {(LUBE,7.70)};

% MVE
\addplot+[ybar, bar width=14pt,
          /utils/exec={\pgfplotscolormapdefinemappedcolor{1000}},
          color=mapped color,
          fill=mapped color,
          draw=black,
          every node near coord/.append style={text=black}, 
          mark=-] coordinates {(MVE,8.85)};

        \end{axis}
        \end{tikzpicture}
        }
        \caption{Climb Phase. Comparison of the mean absolute TGT prediction error for the uncertainty quantification methodologies.}
        \label{rv:fig:app_climb_mae}
        \end{figure}

\Cref{rv:fig:app_climb_mae} highlights point-prediction accuracy in climb. Delta achieves the lowest MAE (5.40$^\circ$C), while Bayesian, LUBE, and MVE exhibit larger errors. Bootstrap remains competitive in MAE but, as shown below, fails to provide calibrated intervals.

%====================================================
% S.D. Absolute TGT Error Barchart for Climb Phase
%====================================================
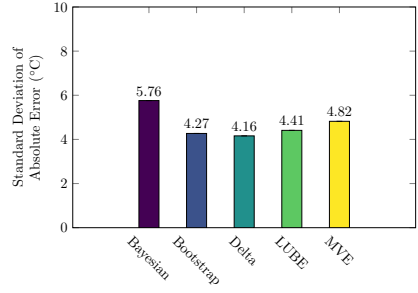
\begin{figure}[H]
        \centering
        \resizebox{0.45\textwidth}{!}{%
        \begin{tikzpicture}
        \begin{axis}[
            at={(6,0)},
            anchor=north west,
            width=10cm,
            height=7cm,
            ymin=0,
            ymax=10,
            enlarge x limits=0.4,
            ylabel={Standard Deviation of\\Absolute Error ($^\circ$C)},
    ylabel style={align=center},
            symbolic x coords={Bayesian,Bootstrap,Delta,LUBE,MVE},
            xtick={Bayesian,Bootstrap,Delta,LUBE,MVE},
            xticklabel style={align=center, rotate=-45},
            nodes near coords,
            nodes near coords align={vertical},
            colormap/viridis,
            colormap name=viridis
]

        % Bayesian
\addplot+[ybar, bar width=14pt,
          /utils/exec={\pgfplotscolormapdefinemappedcolor{0}},
          color=mapped color,
          fill=mapped color,
          draw=black,
          every node near coord/.append style={text=black}, 
          mark=-] coordinates {(Bayesian,5.76)};

% Bootstrap
\addplot+[ybar, bar width=14pt,
          /utils/exec={\pgfplotscolormapdefinemappedcolor{250}},
          color=mapped color,
          fill=mapped color,
          draw=black,
          every node near coord/.append style={text=black}, 
          mark=-] coordinates {(Bootstrap,4.27)};

% Delta
\addplot+[ybar, bar width=14pt,
          /utils/exec={\pgfplotscolormapdefinemappedcolor{500}},
          color=mapped color,
          fill=mapped color,
          draw=black,
          every node near coord/.append style={text=black}, 
          mark=-] coordinates {(Delta,4.16)};

% LUBE
\addplot+[ybar, bar width=14pt,
          /utils/exec={\pgfplotscolormapdefinemappedcolor{750}},
          color=mapped color,
          fill=mapped color,
          draw=black,
          every node near coord/.append style={text=black}, 
          mark=-] coordinates {(LUBE,4.41)};

% MVE
\addplot+[ybar, bar width=14pt,
          /utils/exec={\pgfplotscolormapdefinemappedcolor{1000}},
          color=mapped color,
          fill=mapped color,
          draw=black,
          every node near coord/.append style={text=black}, 
          mark=-] coordinates {(MVE,4.82)};

        \end{axis}
        \end{tikzpicture}
        }
        \caption{Climb Phase. Comparison of the standard deviation of absolute TGT prediction error for the uncertainty quantification methodologies.}
        \label{rv:fig:app_climb_sd_ae}
        \end{figure}

\Cref{rv:fig:app_climb_sd_ae} shows that Bayesian has the largest dispersion of absolute error in climb (5.76$^\circ$C), whereas Delta and bootstrap exhibit lower variability. Lower variability is desirable for stable decision thresholds in EHM applications.

%======================================
% MPIW Barchart for Climb Phase
%======================================
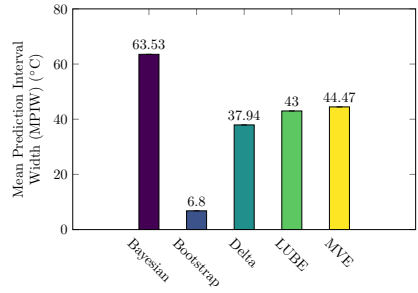
\begin{figure}[H]
        \centering
        \resizebox{0.45\textwidth}{!}{%
        \begin{tikzpicture}
        \begin{axis}[
            at={(6,0)},
            anchor=north west,
            width=10cm,
            height=7cm,
            ymin=0,
            ymax=80,
            enlarge x limits=0.4,
            ylabel={Mean Prediction Interval\\Width (MPIW) ($^\circ$C)},
    ylabel style={align=center},
            symbolic x coords={Bayesian,Bootstrap,Delta,LUBE,MVE},
            xtick={Bayesian,Bootstrap,Delta,LUBE,MVE},
            xticklabel style={align=center, rotate=-45},
            nodes near coords,
            nodes near coords align={vertical},
            colormap/viridis,
            colormap name=viridis
]

        % Bayesian
\addplot+[ybar, bar width=14pt,
          /utils/exec={\pgfplotscolormapdefinemappedcolor{0}},
          color=mapped color,
          fill=mapped color,
          draw=black,
          every node near coord/.append style={text=black}, 
          mark=-] coordinates {(Bayesian,63.53)};

% Bootstrap
\addplot+[ybar, bar width=14pt,
          /utils/exec={\pgfplotscolormapdefinemappedcolor{250}},
          color=mapped color,
          fill=mapped color,
          draw=black,
          every node near coord/.append style={text=black}, 
          mark=-] coordinates {(Bootstrap,6.80)};

% Delta
\addplot+[ybar, bar width=14pt,
          /utils/exec={\pgfplotscolormapdefinemappedcolor{500}},
          color=mapped color,
          fill=mapped color,
          draw=black,
          every node near coord/.append style={text=black}, 
          mark=-] coordinates {(Delta,37.94)};

% LUBE
\addplot+[ybar, bar width=14pt,
          /utils/exec={\pgfplotscolormapdefinemappedcolor{750}},
          color=mapped color,
          fill=mapped color,
          draw=black,
          every node near coord/.append style={text=black}, 
          mark=-] coordinates {(LUBE,43.00)};

% MVE
\addplot+[ybar, bar width=14pt,
          /utils/exec={\pgfplotscolormapdefinemappedcolor{1000}},
          color=mapped color,
          fill=mapped color,
          draw=black,
          every node near coord/.append style={text=black}, 
          mark=-] coordinates {(MVE,44.47)};

        \end{axis}
        \end{tikzpicture}
        }
        \caption{Climb Phase. Comparison of the mean prediction interval width (MPIW) for the uncertainty quantification methodologies.}
        \label{rv:fig:app_climb_mpiw}
        \end{figure}

\Cref{rv:fig:app_climb_mpiw} shows that Bayesian becomes more conservative in climb (MPIW 63.53$^\circ$C). In contrast, Delta and LUBE maintain widths comparable to other phases, and MVE widens moderately to maintain near-perfect coverage.

%======================================
% NMPIW Barchart for Climb Phase
%======================================
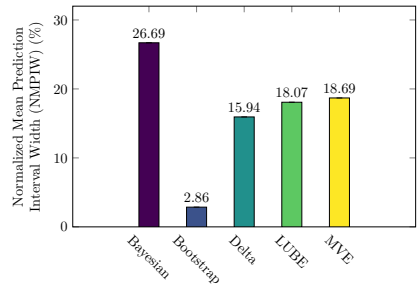
\begin{figure}[H]
        \centering
        \resizebox{0.45\textwidth}{!}{%
        \begin{tikzpicture}
        \begin{axis}[
            at={(6,0)},
            anchor=north west,
            width=10cm,
            height=7cm,
            ymin=0,
            ymax=32,
            enlarge x limits=0.4,
            ylabel={Normalized Mean Prediction\\Interval Width (NMPIW) (\%)},
    ylabel style={align=center},
            symbolic x coords={Bayesian,Bootstrap,Delta,LUBE,MVE},
            xtick={Bayesian,Bootstrap,Delta,LUBE,MVE},
            xticklabel style={align=center, rotate=-45},
            nodes near coords,
            nodes near coords align={vertical},
            colormap/viridis,
            colormap name=viridis
]

        % Bayesian
\addplot+[ybar, bar width=14pt,
          /utils/exec={\pgfplotscolormapdefinemappedcolor{0}},
          color=mapped color,
          fill=mapped color,
          draw=black,
          every node near coord/.append style={text=black}, 
          mark=-] coordinates {(Bayesian,26.69)};

% Bootstrap
\addplot+[ybar, bar width=14pt,
          /utils/exec={\pgfplotscolormapdefinemappedcolor{250}},
          color=mapped color,
          fill=mapped color,
          draw=black,
          every node near coord/.append style={text=black}, 
          mark=-] coordinates {(Bootstrap,2.86)};

% Delta
\addplot+[ybar, bar width=14pt,
          /utils/exec={\pgfplotscolormapdefinemappedcolor{500}},
          color=mapped color,
          fill=mapped color,
          draw=black,
          every node near coord/.append style={text=black}, 
          mark=-] coordinates {(Delta,15.94)};

% LUBE
\addplot+[ybar, bar width=14pt,
          /utils/exec={\pgfplotscolormapdefinemappedcolor{750}},
          color=mapped color,
          fill=mapped color,
          draw=black,
          every node near coord/.append style={text=black}, 
          mark=-] coordinates {(LUBE,18.07)};

% MVE
\addplot+[ybar, bar width=14pt,
          /utils/exec={\pgfplotscolormapdefinemappedcolor{1000}},
          color=mapped color,
          fill=mapped color,
          draw=black,
          every node near coord/.append style={text=black}, 
          mark=-] coordinates {(MVE,18.69)};

        \end{axis}
        \end{tikzpicture}
        }
        \caption{Climb Phase. Comparison of the normalized mean prediction interval width (NMPIW) for the uncertainty quantification methodologies.}
        \label{rv:fig:app_climb_nmpiw}
        \end{figure}

\Cref{rv:fig:app_climb_nmpiw} confirms the same ordering in normalized form, while also making clear that the apparent increase in Bayesian and MVE interval width during climb is not merely a units issue. Even after normalization, the climb regime still demands materially wider calibrated intervals than the other phases.

%======================================
% PICP Barchart for Climb Phase
%======================================
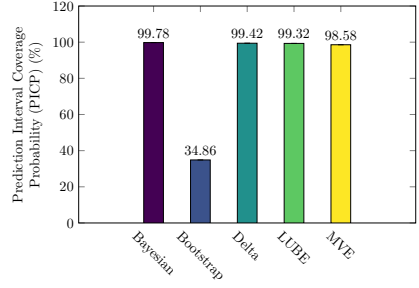
\begin{figure}[H]
        \centering
        \resizebox{0.45\textwidth}{!}{%
        \begin{tikzpicture}
        \begin{axis}[
            at={(6,0)},
            anchor=north west,
            width=10cm,
            height=7cm,
            ymin=0,
            ymax=120,
            enlarge x limits=0.4,
            ylabel={Prediction Interval Coverage\\Probability (PICP) (\%)},
    ylabel style={align=center},
            symbolic x coords={Bayesian,Bootstrap,Delta,LUBE,MVE},
            xtick={Bayesian,Bootstrap,Delta,LUBE,MVE},
            xticklabel style={align=center, rotate=-45},
            nodes near coords,
            nodes near coords align={vertical},
            colormap/viridis,
            colormap name=viridis
]

        % Bayesian
\addplot+[ybar, bar width=14pt,
          /utils/exec={\pgfplotscolormapdefinemappedcolor{0}},
          color=mapped color,
          fill=mapped color,
          draw=black,
          every node near coord/.append style={text=black}, 
          mark=-] coordinates {(Bayesian,99.78)};

% Bootstrap
\addplot+[ybar, bar width=14pt,
          /utils/exec={\pgfplotscolormapdefinemappedcolor{250}},
          color=mapped color,
          fill=mapped color,
          draw=black,
          every node near coord/.append style={text=black}, 
          mark=-] coordinates {(Bootstrap,34.86)};

% Delta
\addplot+[ybar, bar width=14pt,
          /utils/exec={\pgfplotscolormapdefinemappedcolor{500}},
          color=mapped color,
          fill=mapped color,
          draw=black,
          every node near coord/.append style={text=black}, 
          mark=-] coordinates {(Delta,99.42)};

% LUBE
\addplot+[ybar, bar width=14pt,
          /utils/exec={\pgfplotscolormapdefinemappedcolor{750}},
          color=mapped color,
          fill=mapped color,
          draw=black,
          every node near coord/.append style={text=black}, 
          mark=-] coordinates {(LUBE,99.32)};

% MVE
\addplot+[ybar, bar width=14pt,
          /utils/exec={\pgfplotscolormapdefinemappedcolor{1000}},
          color=mapped color,
          fill=mapped color,
          draw=black,
          every node near coord/.append style={text=black}, 
          mark=-] coordinates {(MVE,98.58)};

        \end{axis}
        \end{tikzpicture}
        }
        \caption{Climb Phase. Comparison of the prediction interval coverage probability (PICP) for the uncertainty quantification methodologies.}
        \label{rv:fig:app_climb_picp}
        \end{figure}

\Cref{rv:fig:app_climb_picp} indicates that all non-bootstrap methods achieve near-perfect coverage in climb (98.58--99.78\%), while bootstrap under-covers severely (34.86\%).

%======================================
% CWC Barchart for Climb Phase
%======================================
\begin{figure}[H]
        \centering
        \resizebox{0.45\textwidth}{!}{%
        \begin{tikzpicture}
        \begin{axis}[
            at={(6,0)},
            anchor=north west,
            width=10cm,
            height=7cm,
            ymin=0,
            ymax=130,
            enlarge x limits=0.4,
            ylabel={Coverage Width-based\\Criterion (CWC)},
    ylabel style={align=center},
            symbolic x coords={Bayesian,Bootstrap,Delta,LUBE,MVE},
            xtick={Bayesian,Bootstrap,Delta,LUBE,MVE},
            xticklabel style={align=center, rotate=-45},
            nodes near coords,
            nodes near coords align={vertical},
            colormap/viridis,
            colormap name=viridis
]

        % Bayesian
\addplot+[ybar, bar width=14pt,
          /utils/exec={\pgfplotscolormapdefinemappedcolor{0}},
          color=mapped color,
          fill=mapped color,
          draw=black,
          every node near coord/.append style={text=black}, 
          mark=-] coordinates {(Bayesian,63.53)};

% Bootstrap
\addplot+[ybar, bar width=14pt,
          /utils/exec={\pgfplotscolormapdefinemappedcolor{250}},
          color=mapped color,
          fill=mapped color,
          % show true CWC value instead of 100
          nodes near coords={7.79e13},
          draw=black,
          every node near coord/.append style={text=black}, 
          mark=-] coordinates {(Bootstrap,110)};

% Delta
\addplot+[ybar, bar width=14pt,
          /utils/exec={\pgfplotscolormapdefinemappedcolor{500}},
          color=mapped color,
          fill=mapped color,
          draw=black,
          every node near coord/.append style={text=black}, 
          mark=-] coordinates {(Delta,37.94)};

% LUBE
\addplot+[ybar, bar width=14pt,
          /utils/exec={\pgfplotscolormapdefinemappedcolor{750}},
          color=mapped color,
          fill=mapped color,
          draw=black,
          every node near coord/.append style={text=black}, 
          mark=-] coordinates {(LUBE,43.00)};

% MVE
\addplot+[ybar, bar width=14pt,
          /utils/exec={\pgfplotscolormapdefinemappedcolor{1000}},
          color=mapped color,
          fill=mapped color,
          draw=black,
          every node near coord/.append style={text=black}, 
          mark=-] coordinates {(MVE,44.47)};

        \end{axis}
        \end{tikzpicture}
        }
        \caption{Climb Phase. Comparison of the coverage width-based criterion (CWC) for the uncertainty quantification methodologies.}
        \label{rv:fig:app_climb_cwc}
        \end{figure}
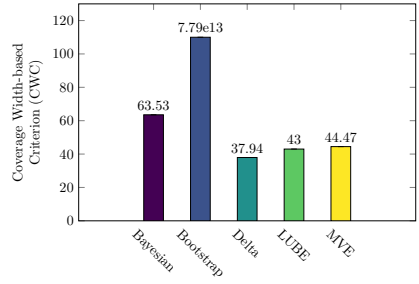

Finally, \Cref{rv:fig:app_climb_cwc} combines coverage and width. Because Delta attains near-perfect coverage with the narrowest calibrated intervals, it yields the lowest CWC in climb, whereas bootstrap incurs an enormous CWC due to under-coverage.

%%%%%%%%%%%%%%%%%%%%%%%%%%%%%%%%%%%%%%%%%%%%%%%%%%%%%%%%

%%%%%%%%%%%%%%%%%%%%%%%%%%%%%%%%%%%%%%%%%%%%%%%%%%%%%%%%%%
\subsection{Cruise Phase}

This section provides additional bar-chart views that focus exclusively on cruise. These figures mirror the cruise row of \Cref{rv:tab:pi_methods_example} but make method-to-method contrasts easier to see at a glance.

%===================================================
% Mean Absolute TGT Error Barchart for Cruise Phase
%===================================================
\begin{figure}[H]
        \centering
        \resizebox{0.45\textwidth}{!}{%
        \begin{tikzpicture}
        \begin{axis}[
            at={(6,0)},
            anchor=north west,
            width=10cm,
            height=7cm,
            ymin=0,
            ymax=10,
            enlarge x limits=0.4,
            ylabel={Mean Absolute Error ($^\circ$C)},
            symbolic x coords={Bayesian,Bootstrap,Delta,LUBE,MVE},
            xtick={Bayesian,Bootstrap,Delta,LUBE,MVE},
            xticklabel style={align=center, rotate=-45},
            nodes near coords,
            nodes near coords align={vertical},
            colormap/viridis,
            colormap name=viridis
]

        % Bayesian
\addplot+[ybar, bar width=14pt,
          /utils/exec={\pgfplotscolormapdefinemappedcolor{0}},
          color=mapped color,
          fill=mapped color,
          draw=black,
          every node near coord/.append style={text=black}, 
          mark=-] coordinates {(Bayesian,5.23)};

% Bootstrap
\addplot+[ybar, bar width=14pt,
          /utils/exec={\pgfplotscolormapdefinemappedcolor{250}},
          color=mapped color,
          fill=mapped color,
          draw=black,
          every node near coord/.append style={text=black}, 
          mark=-] coordinates {(Bootstrap,3.99)};

% Delta
\addplot+[ybar, bar width=14pt,
          /utils/exec={\pgfplotscolormapdefinemappedcolor{500}},
          color=mapped color,
          fill=mapped color,
          draw=black,
          every node near coord/.append style={text=black}, 
          mark=-] coordinates {(Delta,4.09)};

% LUBE
\addplot+[ybar, bar width=14pt,
          /utils/exec={\pgfplotscolormapdefinemappedcolor{750}},
          color=mapped color,
          fill=mapped color,
          draw=black,
          every node near coord/.append style={text=black}, 
          mark=-] coordinates {(LUBE,8.34)};

% MVE
\addplot+[ybar, bar width=14pt,
          /utils/exec={\pgfplotscolormapdefinemappedcolor{1000}},
          color=mapped color,
          fill=mapped color,
          draw=black,
          every node near coord/.append style={text=black}, 
          mark=-] coordinates {(MVE,3.93)};

        \end{axis}
        \end{tikzpicture}
        }
        \caption{Cruise Phase. Comparison of the mean absolute TGT prediction error for the uncertainty quantification methodologies.}
        \label{rv:fig:app_cruise_mae}
        \end{figure}
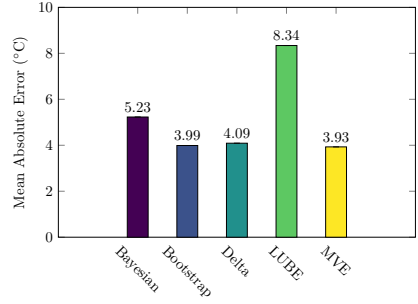

\Cref{rv:fig:app_cruise_mae} shows cruise point accuracy. MVE and bootstrap achieve the lowest MAE (3.93--3.99$^\circ$C), followed closely by Delta, while LUBE exhibits substantially higher error.

%====================================================
% S.D. Absolute TGT Error Barchart for Cruise Phase
%====================================================
\begin{figure}[H]
        \centering
        \resizebox{0.45\textwidth}{!}{%
        \begin{tikzpicture}
        \begin{axis}[
            at={(6,0)},
            anchor=north west,
            width=10cm,
            height=7cm,
            ymin=0,
            ymax=10,
            enlarge x limits=0.4,
            ylabel={Standard Deviation of\\Absolute Error ($^\circ$C)},
    ylabel style={align=center},
            symbolic x coords={Bayesian,Bootstrap,Delta,LUBE,MVE},
            xtick={Bayesian,Bootstrap,Delta,LUBE,MVE},
            xticklabel style={align=center, rotate=-45},
            nodes near coords,
            nodes near coords align={vertical},
            colormap/viridis,
            colormap name=viridis
]

        % Bayesian
\addplot+[ybar, bar width=14pt,
          /utils/exec={\pgfplotscolormapdefinemappedcolor{0}},
          color=mapped color,
          fill=mapped color,
          draw=black,
          every node near coord/.append style={text=black}, 
          mark=-] coordinates {(Bayesian,3.98)};

% Bootstrap
\addplot+[ybar, bar width=14pt,
          /utils/exec={\pgfplotscolormapdefinemappedcolor{250}},
          color=mapped color,
          fill=mapped color,
          draw=black,
          every node near coord/.append style={text=black}, 
          mark=-] coordinates {(Bootstrap,2.69)};

% Delta
\addplot+[ybar, bar width=14pt,
          /utils/exec={\pgfplotscolormapdefinemappedcolor{500}},
          color=mapped color,
          fill=mapped color,
          draw=black,
          every node near coord/.append style={text=black}, 
          mark=-] coordinates {(Delta,2.71)};

% LUBE
\addplot+[ybar, bar width=14pt,
          /utils/exec={\pgfplotscolormapdefinemappedcolor{750}},
          color=mapped color,
          fill=mapped color,
          draw=black,
          every node near coord/.append style={text=black}, 
          mark=-] coordinates {(LUBE,4.23)};

% MVE
\addplot+[ybar, bar width=14pt,
          /utils/exec={\pgfplotscolormapdefinemappedcolor{1000}},
          color=mapped color,
          fill=mapped color,
          draw=black,
          every node near coord/.append style={text=black}, 
          mark=-] coordinates {(MVE,2.73)};

        \end{axis}
        \end{tikzpicture}
        }
        \caption{Cruise Phase. Comparison of the standard deviation of absolute TGT prediction error for the uncertainty quantification methodologies.}
        \label{rv:fig:app_cruise_sd_ae}
        \end{figure}
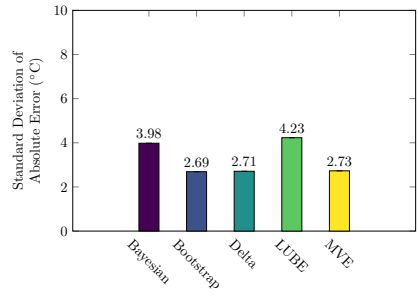

\Cref{rv:fig:app_cruise_sd_ae} indicates that bootstrap, Delta, and MVE have similarly low error dispersion (2.69--2.73$^\circ$C), while Bayesian and LUBE show higher variability.

%======================================
% MPIW Barchart for Cruise Phase
%======================================
\begin{figure}[H]
        \centering
        \resizebox{0.45\textwidth}{!}{%
        \begin{tikzpicture}
        \begin{axis}[
            at={(6,0)},
            anchor=north west,
            width=10cm,
            height=7cm,
            ymin=0,
            ymax=80,
            enlarge x limits=0.4,
            ylabel={Mean Prediction Interval\\Width (MPIW) ($^\circ$C)},
    ylabel style={align=center},
            symbolic x coords={Bayesian,Bootstrap,Delta,LUBE,MVE},
            xtick={Bayesian,Bootstrap,Delta,LUBE,MVE},
            xticklabel style={align=center, rotate=-45},
            nodes near coords,
            nodes near coords align={vertical},
            colormap/viridis,
            colormap name=viridis
]

        % Bayesian
\addplot+[ybar, bar width=14pt,
          /utils/exec={\pgfplotscolormapdefinemappedcolor{0}},
          color=mapped color,
          fill=mapped color,
          draw=black,
          every node near coord/.append style={text=black}, 
          mark=-] coordinates {(Bayesian,46.83)};

% Bootstrap
\addplot+[ybar, bar width=14pt,
          /utils/exec={\pgfplotscolormapdefinemappedcolor{250}},
          color=mapped color,
          fill=mapped color,
          draw=black,
          every node near coord/.append style={text=black}, 
          mark=-] coordinates {(Bootstrap,8.48)};

% Delta
\addplot+[ybar, bar width=14pt,
          /utils/exec={\pgfplotscolormapdefinemappedcolor{500}},
          color=mapped color,
          fill=mapped color,
          draw=black,
          every node near coord/.append style={text=black}, 
          mark=-] coordinates {(Delta,37.91)};

% LUBE
\addplot+[ybar, bar width=14pt,
          /utils/exec={\pgfplotscolormapdefinemappedcolor{750}},
          color=mapped color,
          fill=mapped color,
          draw=black,
          every node near coord/.append style={text=black}, 
          mark=-] coordinates {(LUBE,43.26)};

% MVE
\addplot+[ybar, bar width=14pt,
          /utils/exec={\pgfplotscolormapdefinemappedcolor{1000}},
          color=mapped color,
          fill=mapped color,
          draw=black,
          every node near coord/.append style={text=black}, 
          mark=-] coordinates {(MVE,33.45)};

        \end{axis}
        \end{tikzpicture}
        }
        \caption{Cruise Phase. Comparison of the mean prediction interval width (MPIW) for the uncertainty quantification methodologies.}
        \label{rv:fig:app_cruise_mpiw}
        \end{figure}
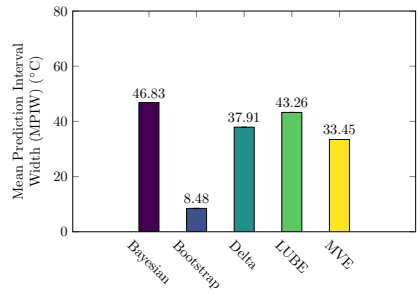

\Cref{rv:fig:app_cruise_mpiw} shows that MVE provides the sharpest calibrated intervals in cruise (33.45$^\circ$C) while maintaining near-perfect coverage, whereas Bayesian and LUBE remain more conservative.

%======================================
% NMPIW Barchart for Cruise Phase
%======================================
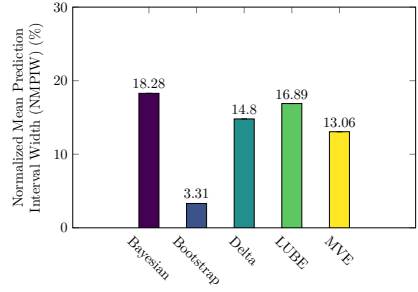
\begin{figure}[H]
        \centering
        \resizebox{0.45\textwidth}{!}{%
        \begin{tikzpicture}
        \begin{axis}[
            at={(6,0)},
            anchor=north west,
            width=10cm,
            height=7cm,
            ymin=0,
            ymax=30,
            enlarge x limits=0.4,
            ylabel={Normalized Mean Prediction\\Interval Width (NMPIW) (\%)},
    ylabel style={align=center},
            symbolic x coords={Bayesian,Bootstrap,Delta,LUBE,MVE},
            xtick={Bayesian,Bootstrap,Delta,LUBE,MVE},
            xticklabel style={align=center, rotate=-45},
            nodes near coords,
            nodes near coords align={vertical},
            colormap/viridis,
            colormap name=viridis
]

        % Bayesian
\addplot+[ybar, bar width=14pt,
          /utils/exec={\pgfplotscolormapdefinemappedcolor{0}},
          color=mapped color,
          fill=mapped color,
          draw=black,
          every node near coord/.append style={text=black}, 
          mark=-] coordinates {(Bayesian,18.28)};

% Bootstrap
\addplot+[ybar, bar width=14pt,
          /utils/exec={\pgfplotscolormapdefinemappedcolor{250}},
          color=mapped color,
          fill=mapped color,
          draw=black,
          every node near coord/.append style={text=black}, 
          mark=-] coordinates {(Bootstrap,3.31)};

% Delta
\addplot+[ybar, bar width=14pt,
          /utils/exec={\pgfplotscolormapdefinemappedcolor{500}},
          color=mapped color,
          fill=mapped color,
          draw=black,
          every node near coord/.append style={text=black}, 
          mark=-] coordinates {(Delta,14.80)};

% LUBE
\addplot+[ybar, bar width=14pt,
          /utils/exec={\pgfplotscolormapdefinemappedcolor{750}},
          color=mapped color,
          fill=mapped color,
          draw=black,
          every node near coord/.append style={text=black}, 
          mark=-] coordinates {(LUBE,16.89)};

% MVE
\addplot+[ybar, bar width=14pt,
          /utils/exec={\pgfplotscolormapdefinemappedcolor{1000}},
          color=mapped color,
          fill=mapped color,
          draw=black,
          every node near coord/.append style={text=black}, 
          mark=-] coordinates {(MVE,13.06)};

        \end{axis}
        \end{tikzpicture}
        }
        \caption{Cruise Phase. Comparison of the normalized mean prediction interval width (NMPIW) for the uncertainty quantification methodologies.}
        \label{rv:fig:app_cruise_nmpiw}
        \end{figure}
\Cref{rv:fig:app_cruise_nmpiw} shows the same trend in normalized terms, reinforcing that MVE remains the sharpest of the calibrated methods even after accounting for scale. This strengthens the conclusion that cruise is the regime where the method achieves its cleanest accuracy--coverage trade-off.

%======================================
% PICP Barchart for Cruise Phase
%======================================
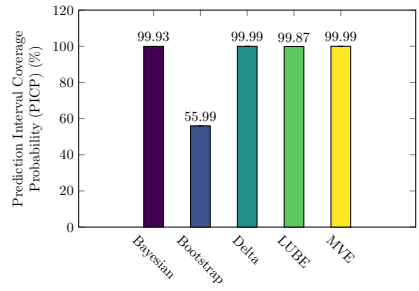
\begin{figure}[H]
        \centering
        \resizebox{0.45\textwidth}{!}{%
        \begin{tikzpicture}
        \begin{axis}[
            at={(6,0)},
            anchor=north west,
            width=10cm,
            height=7cm,
            ymin=0,
            ymax=120,
            enlarge x limits=0.4,
            ylabel={Prediction Interval Coverage\\Probability (PICP) (\%)},
    ylabel style={align=center},
            symbolic x coords={Bayesian,Bootstrap,Delta,LUBE,MVE},
            xtick={Bayesian,Bootstrap,Delta,LUBE,MVE},
            xticklabel style={align=center, rotate=-45},
            nodes near coords,
            nodes near coords align={vertical},
            colormap/viridis,
            colormap name=viridis
]

        % Bayesian
\addplot+[ybar, bar width=14pt,
          /utils/exec={\pgfplotscolormapdefinemappedcolor{0}},
          color=mapped color,
          fill=mapped color,
          draw=black,
          every node near coord/.append style={text=black}, 
          mark=-] coordinates {(Bayesian,99.93)};

% Bootstrap
\addplot+[ybar, bar width=14pt,
          /utils/exec={\pgfplotscolormapdefinemappedcolor{250}},
          color=mapped color,
          fill=mapped color,
          draw=black,
          every node near coord/.append style={text=black}, 
          mark=-] coordinates {(Bootstrap,55.99)};

% Delta
\addplot+[ybar, bar width=14pt,
          /utils/exec={\pgfplotscolormapdefinemappedcolor{500}},
          color=mapped color,
          fill=mapped color,
          draw=black,
          every node near coord/.append style={text=black}, 
          mark=-] coordinates {(Delta,99.99)};

% LUBE
\addplot+[ybar, bar width=14pt,
          /utils/exec={\pgfplotscolormapdefinemappedcolor{750}},
          color=mapped color,
          fill=mapped color,
          draw=black,
          every node near coord/.append style={text=black}, 
          mark=-] coordinates {(LUBE,99.87)};

% MVE
\addplot+[ybar, bar width=14pt,
          /utils/exec={\pgfplotscolormapdefinemappedcolor{1000}},
          color=mapped color,
          fill=mapped color,
          draw=black,
          every node near coord/.append style={text=black}, 
          mark=-] coordinates {(MVE,99.99)};

        \end{axis}
        \end{tikzpicture}
        }
        \caption{Cruise Phase. Comparison of the prediction interval coverage probability (PICP) for the uncertainty quantification methodologies.}
        \label{rv:fig:app_cruise_picp}
        \end{figure}

\Cref{rv:fig:app_cruise_picp} shows that Bayesian, Delta, LUBE, and MVE achieve near-perfect coverage (99.87--99.99\%), while bootstrap under-covers (55.99\%).

%======================================
% CWC Barchart for Cruise Phase
%======================================
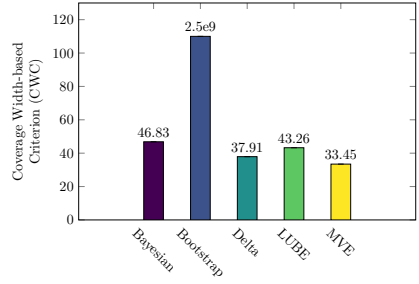
\begin{figure}[H]
        \centering
        \resizebox{0.45\textwidth}{!}{%
        \begin{tikzpicture}
        \begin{axis}[
            at={(6,0)},
            anchor=north west,
            width=10cm,
            height=7cm,
            ymin=0,
            ymax=130,
            enlarge x limits=0.4,
            ylabel={Coverage Width-based\\Criterion (CWC)},
    ylabel style={align=center},
            symbolic x coords={Bayesian,Bootstrap,Delta,LUBE,MVE},
            xtick={Bayesian,Bootstrap,Delta,LUBE,MVE},
            xticklabel style={align=center, rotate=-45},
            nodes near coords,
            nodes near coords align={vertical},
            colormap/viridis,
            colormap name=viridis
]

        % Bayesian
\addplot+[ybar, bar width=14pt,
          /utils/exec={\pgfplotscolormapdefinemappedcolor{0}},
          color=mapped color,
          fill=mapped color,
          draw=black,
          every node near coord/.append style={text=black}, 
          mark=-] coordinates {(Bayesian,46.83)};

% Bootstrap
\addplot+[ybar, bar width=14pt,
          /utils/exec={\pgfplotscolormapdefinemappedcolor{250}},
          color=mapped color,
          fill=mapped color,
          % show true CWC value instead of 100
          nodes near coords={2.5e9},
          draw=black,
          every node near coord/.append style={text=black}, 
          mark=-] coordinates {(Bootstrap,110)};

% Delta
\addplot+[ybar, bar width=14pt,
          /utils/exec={\pgfplotscolormapdefinemappedcolor{500}},
          color=mapped color,
          fill=mapped color,
          draw=black,
          every node near coord/.append style={text=black}, 
          mark=-] coordinates {(Delta,37.91)};

% LUBE
\addplot+[ybar, bar width=14pt,
          /utils/exec={\pgfplotscolormapdefinemappedcolor{750}},
          color=mapped color,
          fill=mapped color,
          draw=black,
          every node near coord/.append style={text=black}, 
          mark=-] coordinates {(LUBE,43.26)};

% MVE
\addplot+[ybar, bar width=14pt,
          /utils/exec={\pgfplotscolormapdefinemappedcolor{1000}},
          color=mapped color,
          fill=mapped color,
          draw=black,
          every node near coord/.append style={text=black}, 
          mark=-] coordinates {(MVE,33.45)};

        \end{axis}
        \end{tikzpicture}
        }
        \caption{Cruise Phase. Comparison of the coverage width-based criterion (CWC) for the uncertainty quantification methodologies.}
        \label{rv:fig:app_cruise_cwc}
        \end{figure}

Finally, \Cref{rv:fig:app_cruise_cwc} reinforces that MVE achieves the most favorable width--coverage trade-off in cruise, while bootstrap again is penalized heavily due to under-coverage.

%%%%%%%%%%%%%%%%%%%%%%%%%%%%%%%%%%%%%%%%%%%%%%%%%%%%%%%%
\section{Conclusion}
\label{rv:sec:conclusion}

This paper benchmarks five uncertainty quantification (UQ) approaches for constructing prediction intervals (PIs) around neural-network forecasts of turbine gas temperature (TGT), a key variable for Engine Health Management (EHM). Following the workflow in \cref{rv:fig:exp_flowchart}, the experiment setup conducts engine-level train/test splitting, selects network hyperparameters using five-fold cross validation, trains each UQ method, and evaluates both point accuracy and interval quality on a held-out set of engines using PICP, MPIW/NMPIW, and CWC (\cref{rv:sec:pi_metrics}).

The novel contributions of this work are threefold: (i) a unified experimental protocol for comparing diverse PI construction methods on the same TGT forecasting task; (ii) a detailed analysis that reports not only aggregate results but also per-engine and per-flight-phase behavior; and (iii) a consistent evaluation suite that makes the accuracy--calibration--sharpness trade-offs explicit for practitioners.

Several key findings emerge from the results. Mean--Variance Estimation (MVE) provides the best overall balance between accuracy, coverage, and interval sharpness: it achieves low MAE while maintaining near-nominal coverage and the most favorable CWC among the calibrated methods (\Cref{rv:tab:method_comparison,rv:fig:overall_cwc}). In contrast, bootstrap ensembles can yield strong point accuracy but poor PI calibration (low PICP), leading to extremely large CWC values and demonstrating that sharpness without coverage is not suitable for risk-aware decision-making (\Cref{rv:fig:overall_picp,rv:fig:overall_cwc}). LUBE achieves high coverage, but often at the expense of wider intervals, while Bayesian MC-dropout tends to be conservative with wide intervals. Finally, the phase-resolved analysis shows strong regime dependence: takeoff is the most challenging phase for calibrated coverage (especially for Bayesian and bootstrap), Bayesian uncertainty grows substantially in climb, and the best method depends on regime---Delta offers the strongest width--coverage trade-off in takeoff and climb, while MVE provides the sharpest calibrated intervals and lowest MAE in cruise (\Cref{rv:tab:pi_comparison,rv:tab:pi_metrics,rv:tab:pi_methods_example,rv:fig:phase_picp,rv:fig:phase_mpiw}).

This study has several limitations. The dataset is proprietary and the held-out evaluation set is small (six engines), so the reported performance may not fully capture the diversity of operational conditions encountered in service. The model class is restricted to a three-hidden-layer MLP and a fixed nominal confidence level, and more expressive sequence architectures, alternative loss formulations, and additional UQ techniques (e.g., conformal prediction) were not explored. Finally, while uncertainty-aware TGT forecasting is a key enabler for prognostics, these PIs were not integrated into an end-to-end probabilistic RUL estimator; that integration is deferred to future work.

Future work will focus on four directions: (i) adding post-hoc calibration methods, such as conformal prediction, to improve coverage under covariate shift; (ii) developing regime-aware or phase-conditioned UQ models for changing operational dynamics; (iii) extending the benchmark to additional architectures and sensors; and (iv) propagating TGT uncertainty through prognostic models to quantify its effect on RUL and maintenance decision-making.

%%%%%%%%%%%%%%%%%%%%%%%%%%%%%%%%%%%%%%%%%%%%%%%%%%%%%%%%%%%%%%%%%%%%%%
\section{Funding}
This research was sponsored by Rolls-Royce and Virginia Tech.

%%%%%%%%%%%%%%%%%%%%%%%%%%%%%%%%%%%%%%%%%%%%%%%%%%%%%%%%%%%%%%%%%%%%%%
\section{Data Availability}

The data supporting the findings of this study were made available from Rolls-Royce plc, but restrictions apply to the availability of these data, which were used under license for the current study and are not publicly available. The point of contact for data requests is changminson@vt.edu.

%%%%%%%%%%%%%%%%%%%%%%%%%%%%%%%%%%%%%%%%%%%%%%%%%%%%%%%%%%%%%%%%%%%%%%
\section{Acknowledgements}

The authors would like to express sincere appreciation to Rolls-Royce for sharing Engine Health Management (EHM) data and technical insights. The authors extend special thanks to Jin-Sol Jung for technical discussion about mechanical engineering aspects pertaining to the EHM data.

%%%%%%%%%%%%%%%%%%%%%%%%%%%%%%%%%%%%%%%%%%%%%%%%%%

\section{Declaration of Competing Interest}
Authors Gavan Burke, Rekha Sundararajan, Andrew Rimell, and Gregory Steinrock were employed by Rolls-Royce. The remaining authors declare that the research was conducted in the absence of any commercial or financial relationships that could be construed as a potential conflict of interest.

%%%%%%%%%%%%%%%%%%%%%%%%%%%%%%%%%%%%%%%%%%%%%%%%%%%%%%%%%%%%%%%%%%%%%%

\section{Declaration of generative AI and AI-assisted technologies in the manuscript preparation process}
During the preparation of this work, the authors used OpenAI's ChatGPT in order to assist with copy editing, literature review, and crafting the formal notation of the presented methodologies and their underlying concepts. After using this tool/service, the authors reviewed and edited the content as needed and take full responsibility for the content of the published article.

\clearpage
\bibliographystyle{elsarticle-num}
\bibliography{references}

\end{document}